\documentclass[lettersize,journal]{IEEEtran}
\usepackage{color,array}

\usepackage{amsmath,amsfonts}
\usepackage{algorithmic}
\usepackage{algorithm}
\usepackage[caption=false,font=normalsize,labelfont=sf,textfont=sf]{subfig}
\usepackage{textcomp}
\usepackage{stfloats}
\usepackage{url}
\usepackage{verbatim}
\usepackage{graphicx}
\usepackage{multirow}
\usepackage{booktabs}
\usepackage[pagebackref,breaklinks,colorlinks]{hyperref}
\usepackage{orcidlink}
\usepackage{ulem}
\usepackage[numbers,sort&compress]{natbib}

\def\BibTeX{{\rm B\kern-.05em{\sc i\kern-.025em b}\kern-.08em
    T\kern-.1667em\lower.7ex\hbox{E}\kern-.125emX}}
\usepackage{balance}

\def\eg{{\it{e.g.}}}
\def\etal{{\it{et al.}}}
\def\ie{{\it{i.e.}}}
\def\etc{{\it{etc.}}}

\newcommand{\tao}[1]{\textcolor{black}{#1}}
\newcommand{\taore}[1]{\textcolor{black}{#1}}
\newcommand{\taoR}[1]{\textcolor{black}{#1}}
\newcommand{\taoq}[1]{\textcolor{black}{\textbf{#1}}}
\newcommand{\hh}[1]{\textcolor{black}{#1}}
\newcommand{\hhh}[1]{\textcolor{black}{#1}}
\newcommand{\lyw}[1]{\textcolor{black}{#1}}
\newcommand{\lywre}[1]{\textcolor{black}{#1}}

\newcommand{\Best}[1]{\textcolor{red}{\textbf{#1}}}
\newcommand{\SecondBest}[1]{\textcolor{cyan}{#1}}
\newcommand{\Better}[1]{\textcolor{cyan}{{#1}}}

\begin{document}

\title{Glass Surface Detection: Leveraging Reflection Dynamics in Flash/No-flash Imagery}

\author{
    Tao Yan\IEEEauthorrefmark{1},~\IEEEmembership{Senior Member,~IEEE},
    Zeyu Wang,
    Hao Huang,
    Yiwei Lu,
    Ke Xu,
    Chunping Ge,
    Yinghui Wang,
    Xiaojun Chang,~\IEEEmembership{Senior Member,~IEEE},
    Rynson W.H. Lau,~\IEEEmembership{Senior Member,~IEEE}
    \thanks{Tao Yan, Zeyu Wang, Yiwei Lu, Hao Huang and Yinghui Wang are with the School of Artificial Intelligence and Computer Science at Jiangnan University, Wuxi, Jiangsu 214122, China.}
    \thanks{Ke Xu and Xiaojun Chang are with University of Science and Technology of China, Hefei 230026, China.}
    \thanks{Chunping Ge is with Weinan Normal University, Weinan 714099, China.}
    \thanks{Rynson W.H. Lau are with the Department of Computer Science, City University of Hong Kong, Hong, Kong.}
    \thanks{*Corresponding authors is Tao Yan (e-mail: yantao.ustc@gmail.com).
    }
}

\markboth{Journal of \LaTeX\ Class Files,~Vol.~18, No.~9, September~2020}%
{How to Use the IEEEtran \LaTeX \ Templates}

\maketitle
  
\begin{abstract}
Glass surfaces are ubiquitous in daily life, typically appearing colorless, transparent, and lacking distinctive features. These characteristics make glass surface detection a challenging computer vision task. Existing glass surface detection methods always rely on boundary cues (\textit{e.g.}, window and door frames) or reflection cues to locate glass surfaces, but they fail to fully exploit the intrinsic properties of the glass itself for accurate localization. 
We observed that in most real-world scenes, the illumination intensity in front of the glass surface differs from that behind it, which results in variations in the reflections visible on the glass surface.
Specifically, when standing on the brighter side of the glass and applying a flash towards the darker side, existing reflections on the glass surface tend to disappear. Conversely, while standing on the darker side and applying a flash towards the brighter side, distinct reflections will appear on the glass surface.
Based on this phenomenon, we propose \textit{NFGlassNet}, a novel method for glass surface detection that leverages the reflection dynamics present in flash/no-flash imagery. Specifically, we propose a Reflection Contrast Mining Module (RCMM) for extracting reflections, and a Reflection Guided Attention Module (RGAM) for fusing features from reflection and glass surface for accurate glass surface detection. For learning our network, we also construct a dataset consisting of $\sim$3.3$K$ no-flash and flash image pairs captured from various scenes with corresponding ground truth annotations. Extensive experiments demonstrate that our method outperforms the state-of-the-art methods. Our code, model, and dataset will be available upon acceptance of the manuscript.
\end{abstract}


\section{Introduction}
\IEEEPARstart{G}{lass} surfaces, including glass walls, \tao{glass doors,} \tao{f}rench windows, glass tabletops, and glass railings are prevalent in daily life. 
\tao{Glass} surfaces lack distinctive external features and \tao{their appearances mainly depend on surroundings}, which poses challenges for \tao{related} computer vision tasks such as autonomous driving and 3D reconstruction. 
\tao{In the past few years, several single visible (RGB) image-based glass surface detection methods}~\cite{mei2020don,lin2021rich,GEM2025,he2021enhanced,yu2022progressive,qi2024glass,Yan2025ghosting} have been proposed \tao{and achieved good performance in the daytime}. 
However, most of these methods heavily rely on boundary \tao{cues}~\cite{mei2020don,he2021enhanced,yu2022progressive}, \tao{reflection or ghosting cues~\cite{lin2021rich,Yan2025ghosting}}, and \tao{moreover,} limited cues \tao{embedding in} a single RGB image \tao{may} result in detection failures in challenging scenes. 
\tao{Recently, several multi-modal based methods~\cite{mei2022polarization,DGSDNet,yan2024nrglassnet,huo2023glass, panoglass} are able to} identified glass surfaces with \tao{different types of sensors}. \tao{However,} these methods usually require extra expensive sensors \tao{in addition to regular cameras}, and \tao{multi-modal image} alignment \tao{and fusion}.

\newcommand{\newsubwidth}{0.156}
\begin{figure}
	\renewcommand{\tabcolsep}{0.8pt}
	\renewcommand\arraystretch{0.6}
        \begin{center}
            \begin{tabular}{cccccc}
                \includegraphics[width=\newsubwidth\linewidth]{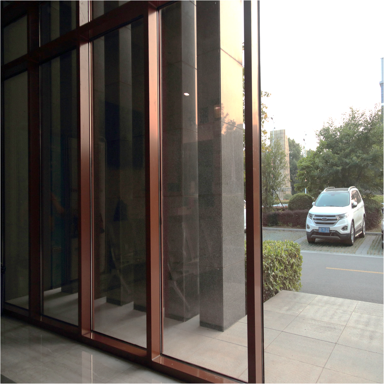}&
                \includegraphics[width=\newsubwidth\linewidth]{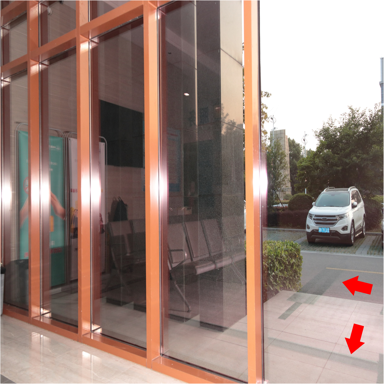}&
                \includegraphics[width=\newsubwidth\linewidth]{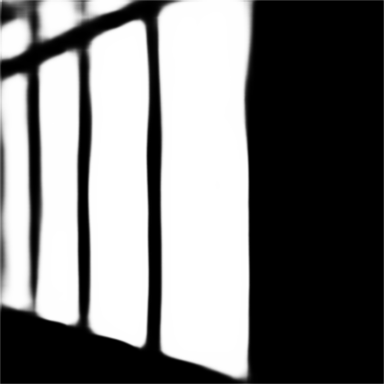}&
                \includegraphics[width=\newsubwidth\linewidth]{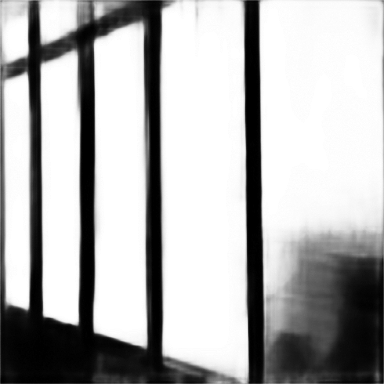}&
                \includegraphics[width=\newsubwidth\linewidth]{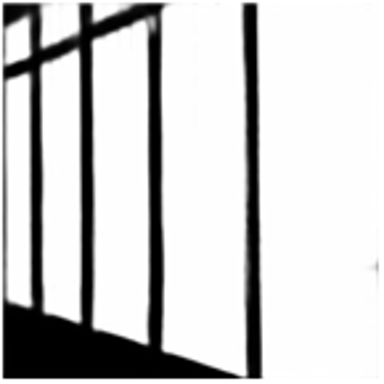}&
                \includegraphics[width=\newsubwidth\linewidth]{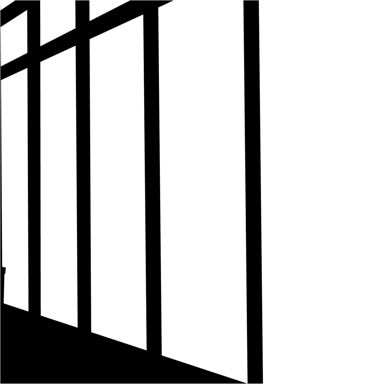}\\ 
                
                \includegraphics[width=\newsubwidth\linewidth]{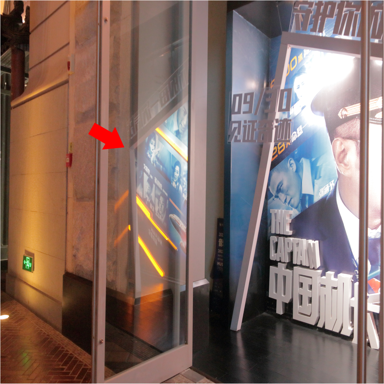}&
                \includegraphics[width=\newsubwidth\linewidth]{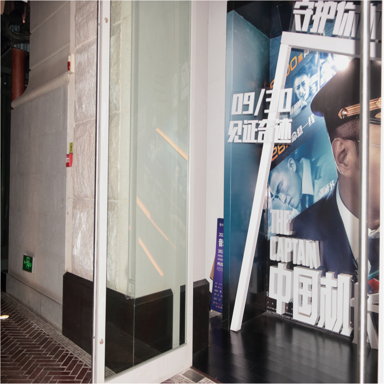}&
                \includegraphics[width=\newsubwidth\linewidth]{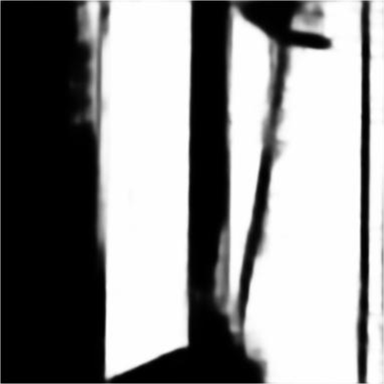}&
                \includegraphics[width=\newsubwidth\linewidth]{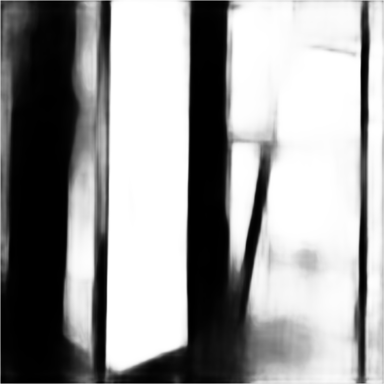}&
                \includegraphics[width=\newsubwidth\linewidth]{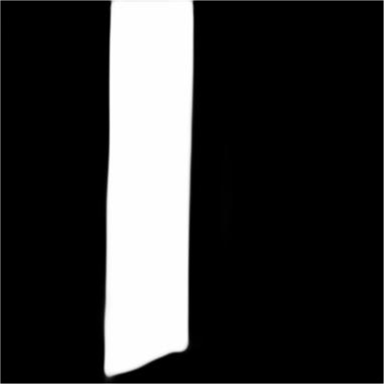}&
                \includegraphics[width=\newsubwidth\linewidth]{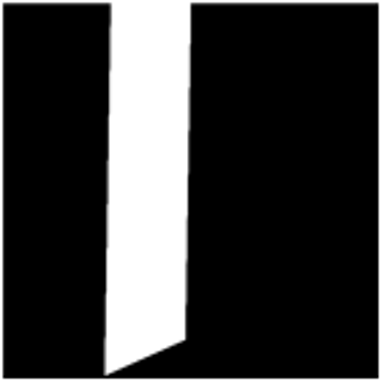}\\ 
                
                \includegraphics[width=\newsubwidth\linewidth]{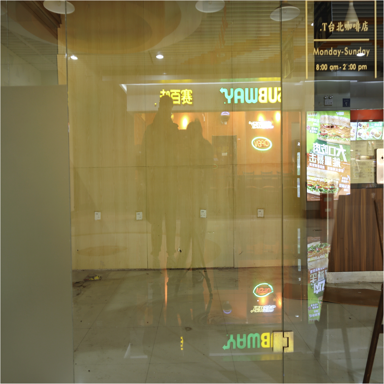}&
                \includegraphics[width=\newsubwidth\linewidth]{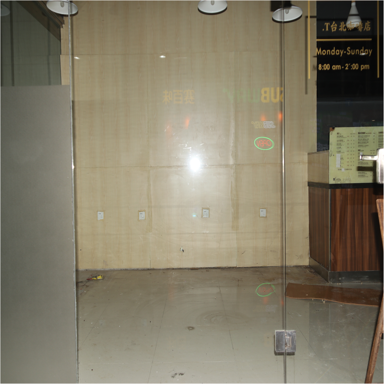}&
                \includegraphics[width=\newsubwidth\linewidth]{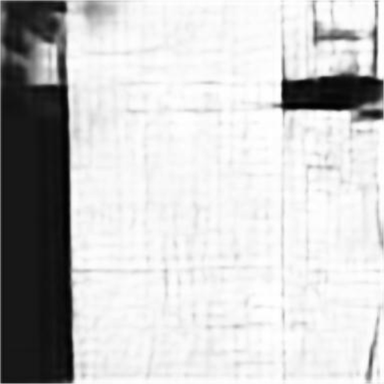}&
                \includegraphics[width=\newsubwidth\linewidth]{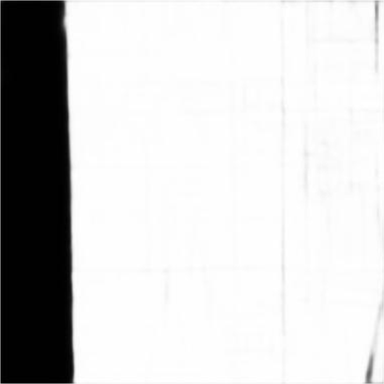}&
                \includegraphics[width=\newsubwidth\linewidth]{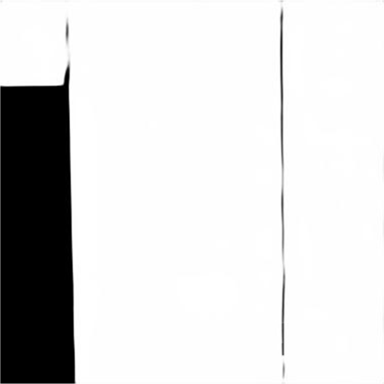}&
                \includegraphics[width=\newsubwidth\linewidth]{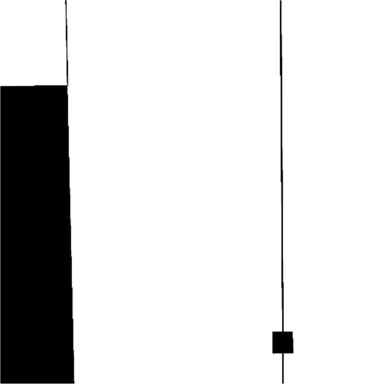}\\

                \includegraphics[width=\newsubwidth\linewidth]{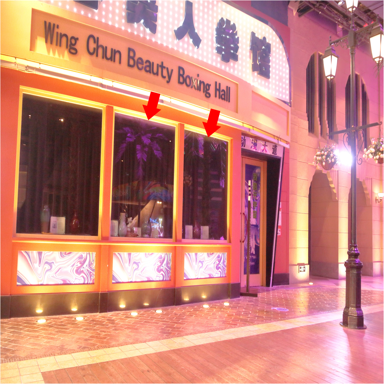}&
                \includegraphics[width=\newsubwidth\linewidth]{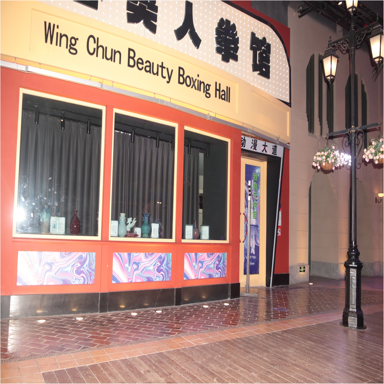}&
                \includegraphics[width=\newsubwidth\linewidth]{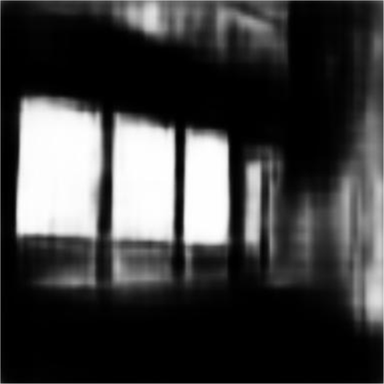}&
                \includegraphics[width=\newsubwidth\linewidth]{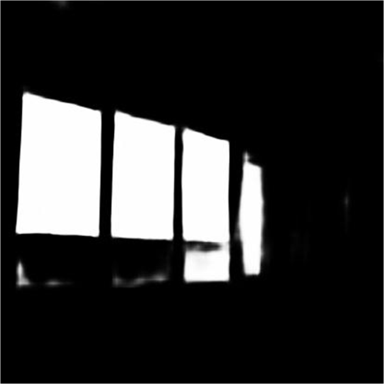}&
                \includegraphics[width=\newsubwidth\linewidth]{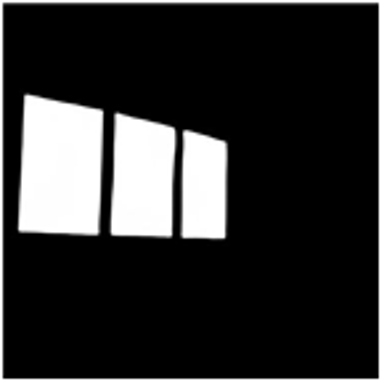}&
                \includegraphics[width=\newsubwidth\linewidth]{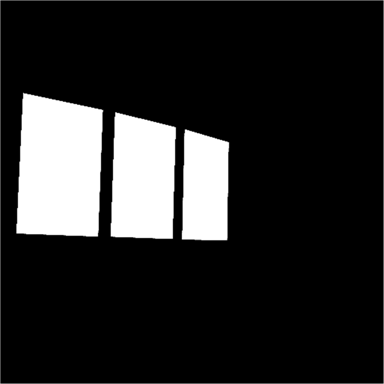}\\

                \fontsize{7.0pt}{\baselineskip}\selectfont{No-flash Image}&
                \fontsize{7.0pt}{\baselineskip}\selectfont{Flash Image}&
                \fontsize{7.0pt}{\baselineskip}\selectfont{GSDNet~\cite{lin2021rich}}&
                \fontsize{7.0pt}{\baselineskip}\selectfont{RGB-T~\cite{huo2023glass}}&
                \fontsize{7.0pt}{\baselineskip}\selectfont{Ours}&
                \fontsize{7.0pt}{\baselineskip}\selectfont{GT}\\
                
            \end{tabular}
        \end{center}
 	\vspace{-0.01\textwidth}
        \caption{Comparison of our \textit{NFGlassNet} with the state-of-the-art glass surface detection methods. \taore{For each scene from left to right: the first and second columns are no-flash image and flash image, respectively, and the rest columns are results produced by the competing methods and our method.} GSDNet~\cite{lin2021rich} tends to under-detect glass when there is no reflection on the glass surface ($1st$ scene, no-flash image). Both GSDNet~\cite{lin2021rich} and RGB-T~\cite{huo2023glass} are prone to over-detection when there are numerous glass-like frames ($2nd$ scene), as they heavily rely on \tao{the} boundary cues. \tao{Our method can effectively leverage the appearance and} disappearance of reflections in the \tao{same} scene, which enables more accurate identification of glass regions. The \textcolor{red}{red arrows} indicate the locations of reflections, and zooming in on the images can provide a better visual effect.}
        \label{fig:instrcution_result_show}
\end{figure}

We \tao{can observe a valuable phenomenon that} \tao{the illumination intensity of the space in front of a glass surface (e.g., glass wall, door, or window) is often different from that of the space behind in daily life.
Typically, when we take photographs from the brighter side of a glass surface towards the other darker side}, \tao{there are always obvious reflections on the glass surface in the captured image}. 
However, while a flash \tao{accompanying the camera toward the glass surface} is applied, the reflections tend to \tao{be weakened} and even disappear, \tao{since the dark space behind the glass is also illuminated, as shown in Fig.~\ref{fig:instrcution_result_show}.} 
Conversely, when we photograph from the darker side of the glass, there \taore{usually will not appear very obvious reflections} on the glass surface, \tao{but applying a flash while shooting would} result in \tao{more prominent} reflections on the glass surface.
\begin{figure}[t]
  \centering
    \begin{minipage}[b]{\linewidth}
      \subfloat[Light Side \textit{w/o} Flash]{
        \label{fig:light_side_no_flash}
        \includegraphics[width=0.47\linewidth]{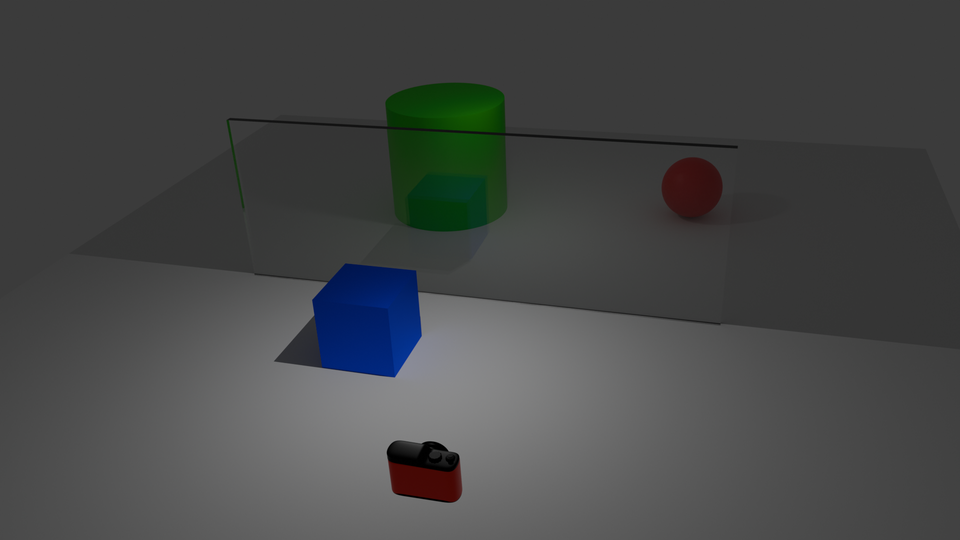}
      }
      \subfloat[Light Side \textit{w/} Flash]{
        \label{fig:light_side_flash}
        \includegraphics[width=0.47\linewidth]{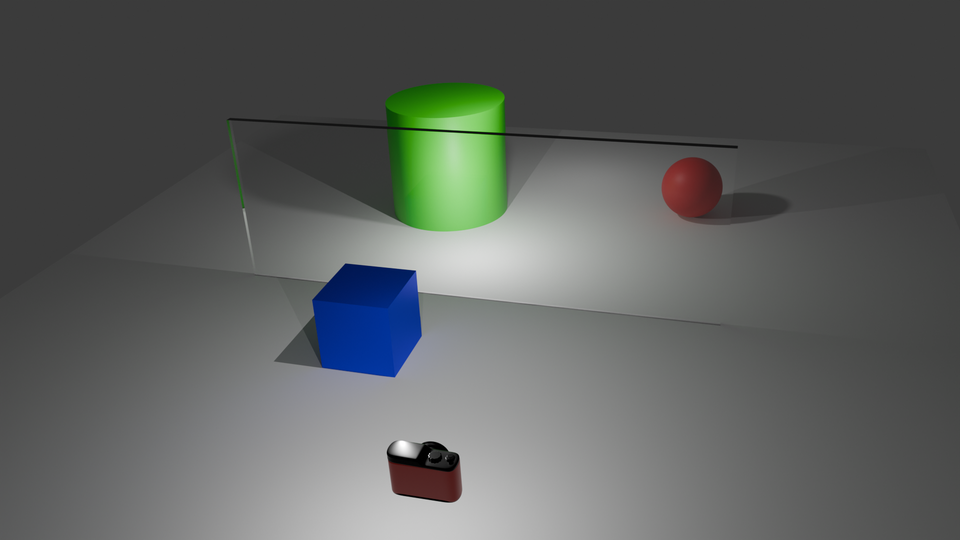}
      }
    \end{minipage}
    
    \begin{minipage}[b]{\linewidth}
      \subfloat[Dark Side \textit{w/o} Flash]{
        \label{fig:dark_side_no_flash}
        \includegraphics[width=0.47\linewidth]{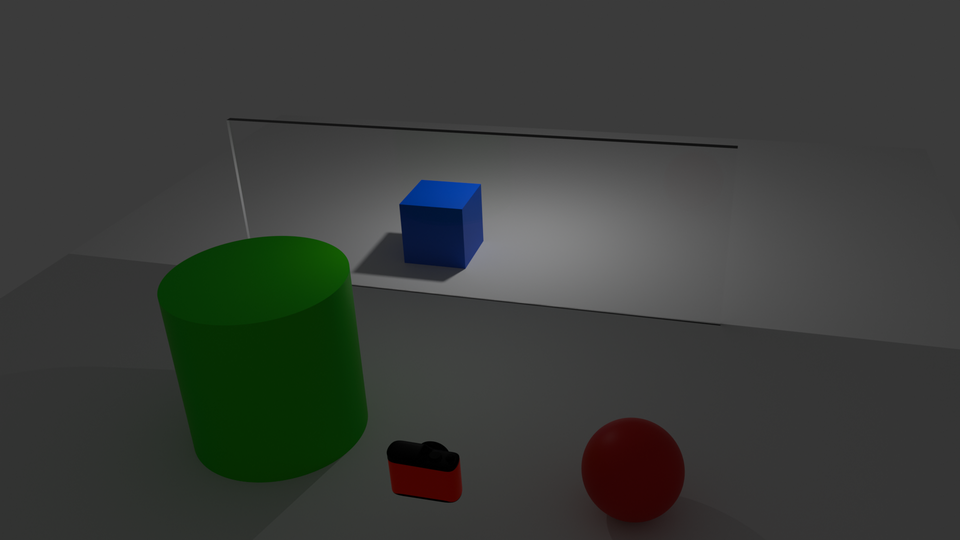}
      }
      \subfloat[Dark Side \textit{w/} Flash]{
        \label{fig:dark_side_flash}
        \includegraphics[width=0.47\linewidth]{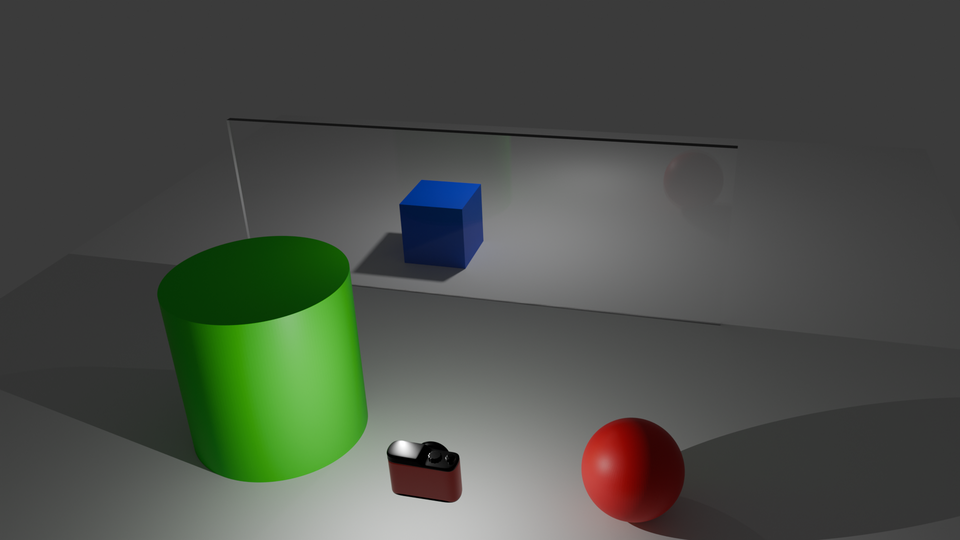}
      }
    \end{minipage}

    \caption{Scenes rendering by Blender for appearance ($1st$ scene) and disappearance ($2nd$ scene) of reflection: Distinct reflection on the glass surface will disappear when taking photographs from the high-lighted side and applying flash towards the darker side (from Fig.~\ref{fig:light_side_no_flash} to Fig.~\ref{fig:light_side_flash}). Conversely, \tao{in some situations,} clear reflection will appear on the glass surface when we take photographs from the low-lighted side and apply flash towards the light side (from Fig.~\ref{fig:dark_side_no_flash} to Fig.~\ref{fig:dark_side_flash}).}
    \label{fig:blender_rendering}
  \end{figure}


\tao{This observation can be illustrated in Fig.~\ref{fig:blender_rendering}, 
\hh{and the camera's views of the observation are further shown in our supplementary material.}
\tao{We rendered these examples with different illumination intensities with Blender~\cite{blender}}.}
\tao{Specifically, Fig.~\ref{fig:light_side_no_flash} shows that a prominent reflection (blue box) will appear on the glass surface, if the lighting intensity of the space in front of the glass surface is obviously stronger than that of the space behind the glass surface.}
Fig.~\ref{fig:light_side_flash} shows that \tao{while a flash accompanying the camera is applied, the lighting} intensity of the darker side \taore{behind} the glass surface \tao{will be enhanced}. 
\taore{Then,} the reflection \tao{appeared on the} glass surface will become indistinguishable and even be overshadowed by the transmission layer. 
\taore{Therefore, the no-flash and flash image pair shown in Fig.~\ref{fig:light_side_no_flash} and~\ref{fig:light_side_flash} is able to aid in reflection cue extraction and glass surface detection.}
\taore{Similarly, in} Fig.~\ref{fig:dark_side_no_flash}, the camera in the darker side of the glass surface cannot capture \taore{obvious} reflection due to \tao{insufficient lighting}. 
Then, as shown in Fig.~\ref{fig:dark_side_flash}, while a flash is applied, the dark space in front of the glass surface will be illuminated, \tao{resulting in prominent reflections} \tao{on the glass surface}. \taore{Thus, the no-flash and flash-image pair shown in Fig.~\ref{fig:dark_side_no_flash} and~\ref{fig:dark_side_flash} can contribute to glass surface detection.}
%

Several real-world \tao{scenes without and with flash} are shown in Fig.~\ref{fig:instrcution_result_show}.
As shown in the $1st$ \tao{scene}, when there is no reflection on the glass surface, GSDNet~\cite{lin2021rich} will under-detect the glass surface at the right \tao{side} of the scene, and even humans \tao{is hard to distinguish} whether there is glass at the rightmost \hh{glass-like region}. 
However, \tao{while} a flash is applied in the same scene, the appearance of a reflection prompts \tao{our method} to identify the glass surface, which highlights the importance of applying a flash. 
Additionally, as shown in the $3rd$ \tao{scene}, after applying the flash, the original \taore{obvious} reflection on the glass surface disappears, while the non-glass \tao{homogeneous region in the left side of the image} does not exhibit such an obvious difference.

In this paper, we propose a \textit{\textbf{N}o-flash and \textbf{F}lash \textbf{
Glass} Surface Detection \textbf{Net}work (\textit{\textbf{NFGlassNet}})}, which contains two novel modules: Reflection Contrast Mining Module (RCMM) \tao{for accurately extracting reflections} \taore{from non-flash and flash image pairs}, and Reflection Guided Attention Module (RGAM)  \tao{for identifying} glass surface \taore{guided by the extracted} reflection cues. 
%
In addition, for learning our \textit{NFGlassNet}, we construct a large scale \textbf{N}o-flash and \textbf{F}lash \textbf{G}lass Surface \textbf{D}ataset (NFGD) based on the reflection dynamics (appearance and disappearance) in flash/no-flash imagery, which consists of approximately 3.3k no-flash and flash image pairs captured with a Canon EOS 80D equipped with a flash. 
Unlike existing no-flash and flash datasets~\cite{aksoy2018dataset,lei2021robust,hong2024light} proposed for low-level tasks (\eg, reflection removal~\cite{lei2021robust,hong2024light, PPFPNet2024}, denoising~\cite{aksoy2018dataset}), our NFGD contains a large number of glass surface instances across diverse scenes. 
\taore{Notably, for deploying our method, a program can be customed to control the target camera (\eg, phone) to capture no-flash and flash image pairs in sequence (no-flash shot followed by a flash shot).}

\tao{In general}, the main contributions of our work \tao{can be} summarized as follows:
\begin{itemize}
    \item We propose a novel network, called \textit{NFGlassNet}, which leverages the reflection dynamics present in flash/no-flash imagery for glass surface detection. 

    \item We propose a Reflection Contrast Mining Module (RCMM) for \tao{extracting reflection from non-flash and flash image pairs}, and a Reflection Guided Attention Module (RGAM) \taore{for accurately detecting glass surfaces guided by the reflection cues}.

    \item We \taore{construct a new} glass surface \taore{detection} dataset, \tao{named NFGD}, \tao{consisting of $3312$ no-flash and flash image pairs along with corresponding annotations.}
 
    \item Extensive experiments conducted on \taore{various} real-world scenes demonstrate the superior performance of our method \tao{compared with the} state-of-the-art methods.
\end{itemize}

\newcommand{\newsubheight}{0.0858}
\begin{figure*}
	\renewcommand{\tabcolsep}{0.8pt}
	\renewcommand\arraystretch{0.6}
    \begin{center}
        \begin{tabular}{cccccccc}
                \includegraphics[height=\newsubheight\linewidth, keepaspectratio]{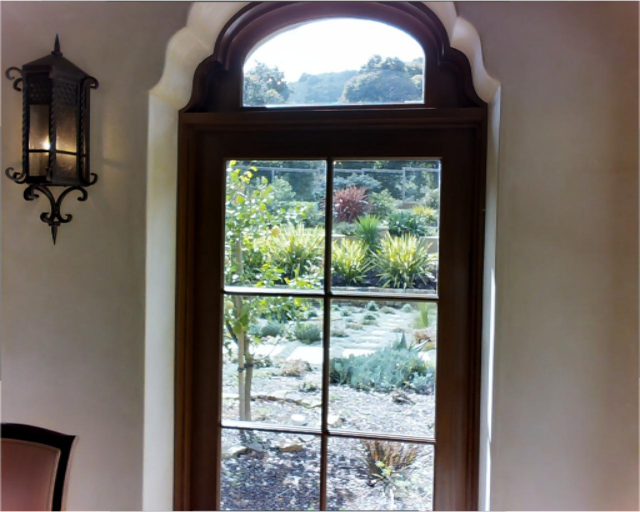}&
                \includegraphics[height=\newsubheight\linewidth, keepaspectratio]{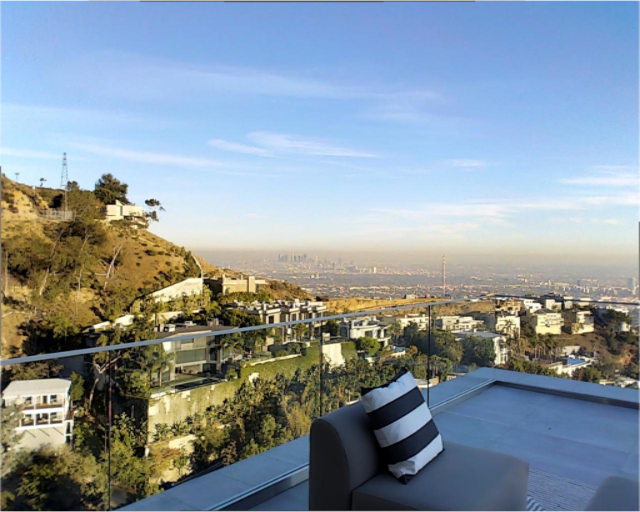}&
                \hspace{1.0pt}
                \includegraphics[height=\newsubheight\linewidth, keepaspectratio]{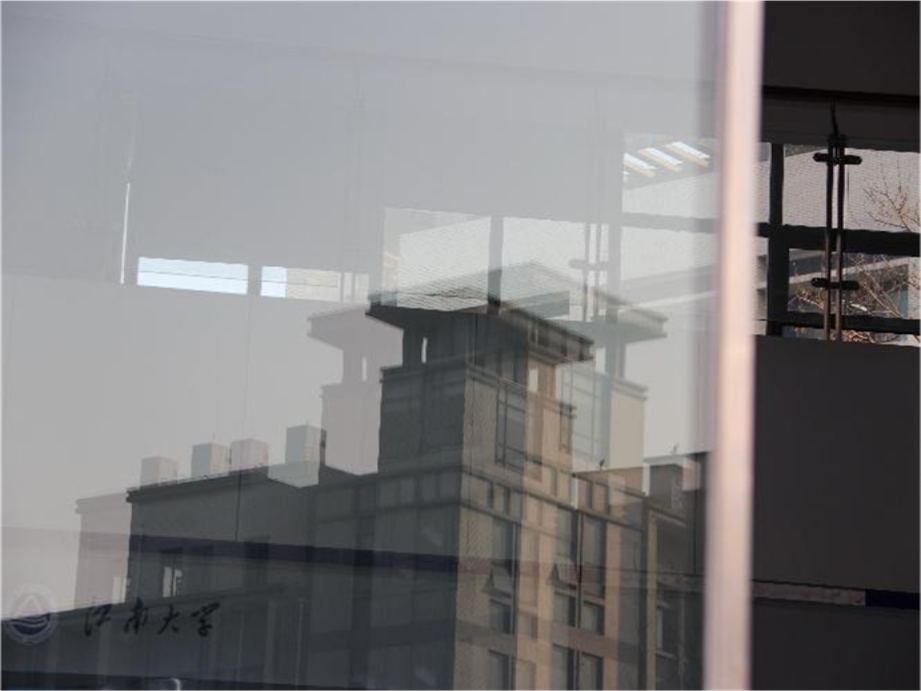}&
                \includegraphics[height=\newsubheight\linewidth, keepaspectratio]{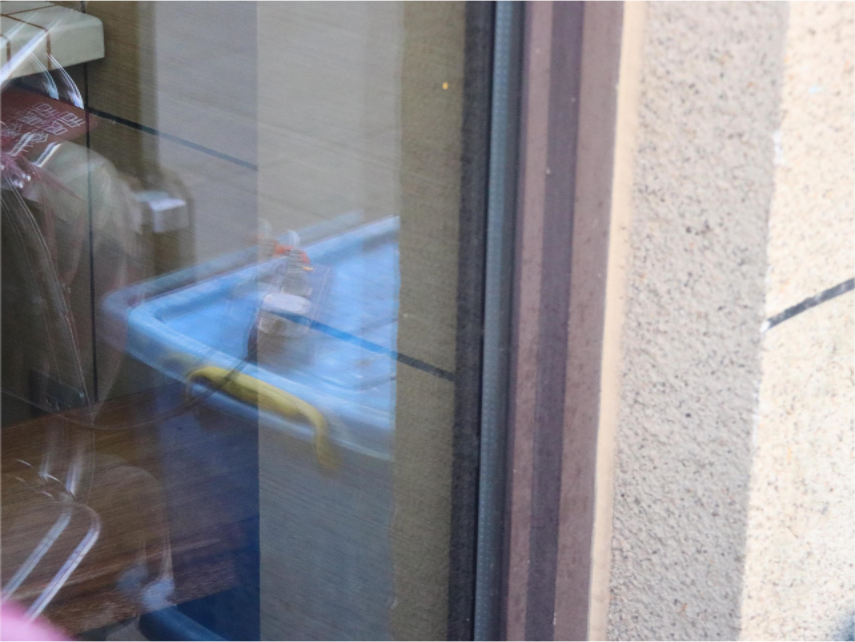}&
                \hspace{1.0pt}
                \includegraphics[height=\newsubheight\linewidth, keepaspectratio]{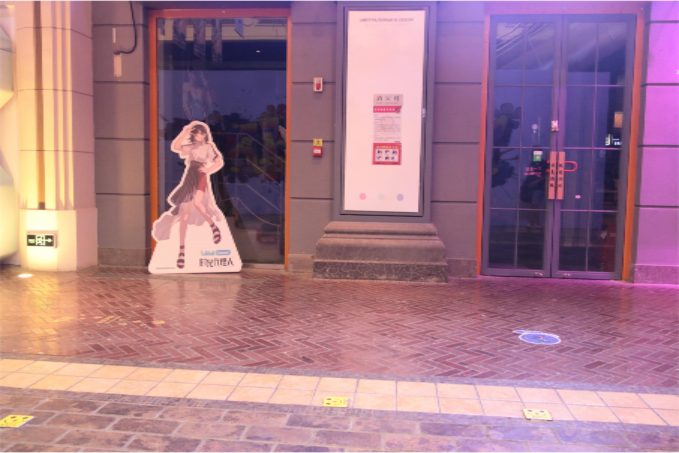}&
                \includegraphics[height=\newsubheight\linewidth, keepaspectratio]{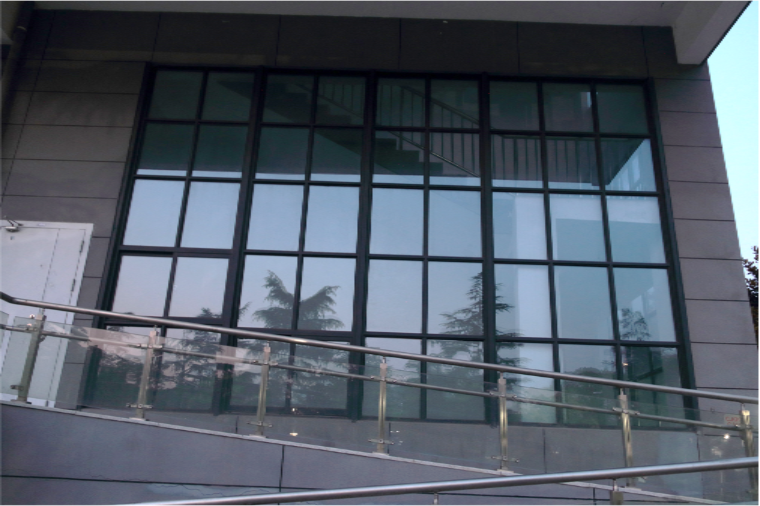}&
                \includegraphics[height=\newsubheight\linewidth, keepaspectratio]{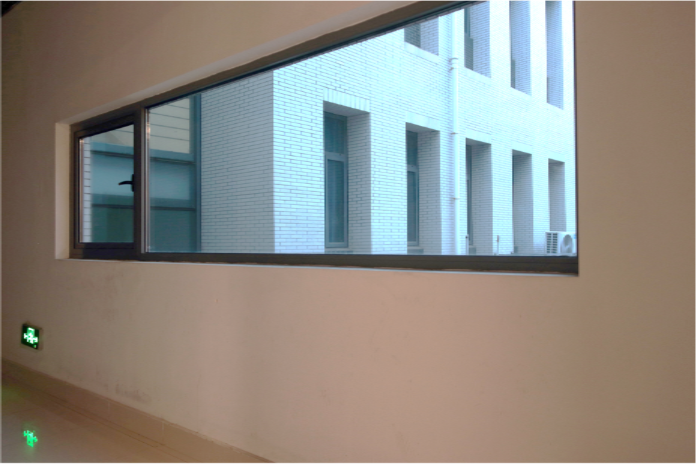}&
                \includegraphics[height=\newsubheight\linewidth, keepaspectratio]{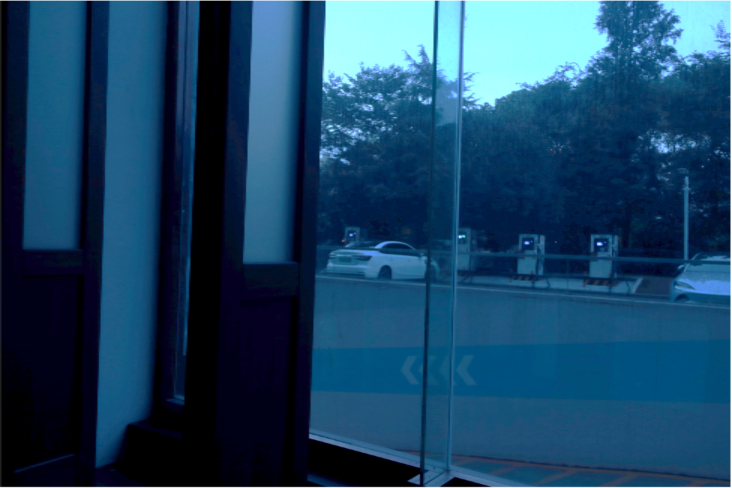}\\
  
                \includegraphics[height=\newsubheight\linewidth, keepaspectratio]{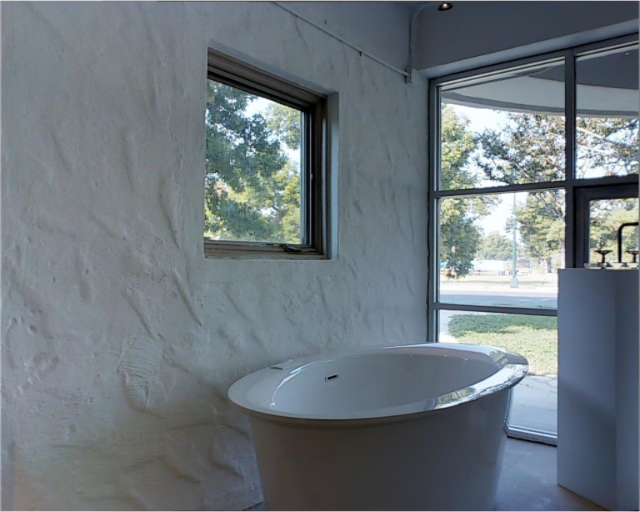}&
                \includegraphics[height=\newsubheight\linewidth, keepaspectratio]{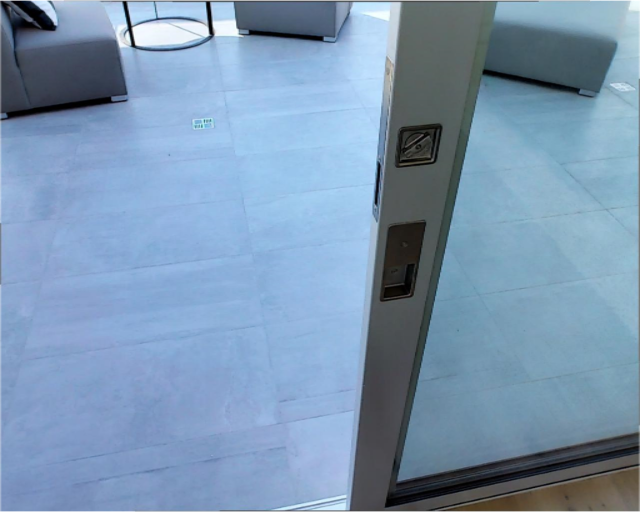}&
                \hspace{1.0pt}
                \includegraphics[height=\newsubheight\linewidth, keepaspectratio]{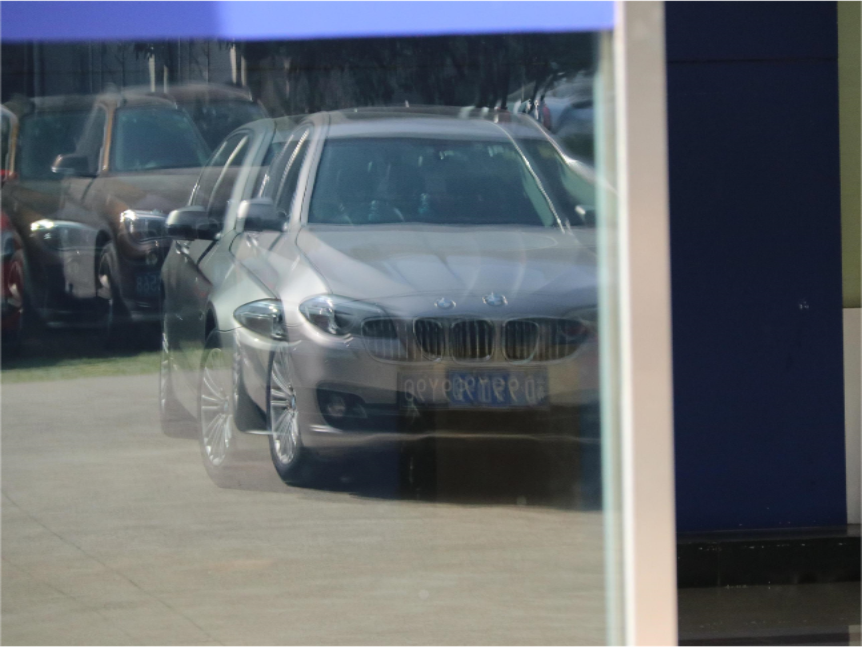}&
                \includegraphics[height=\newsubheight\linewidth, keepaspectratio]{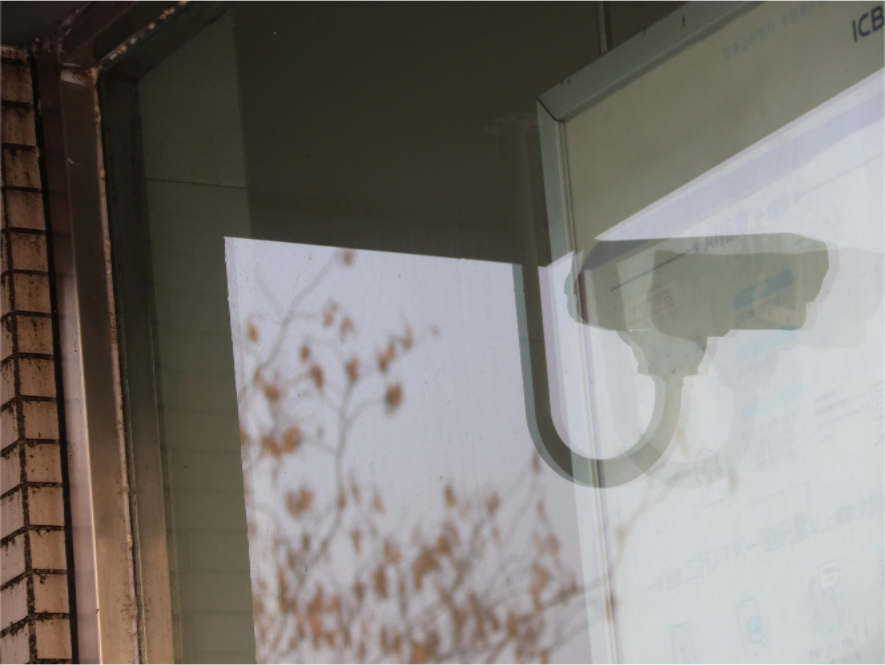}&
                \hspace{1.0pt}
                \includegraphics[height=\newsubheight\linewidth, keepaspectratio]{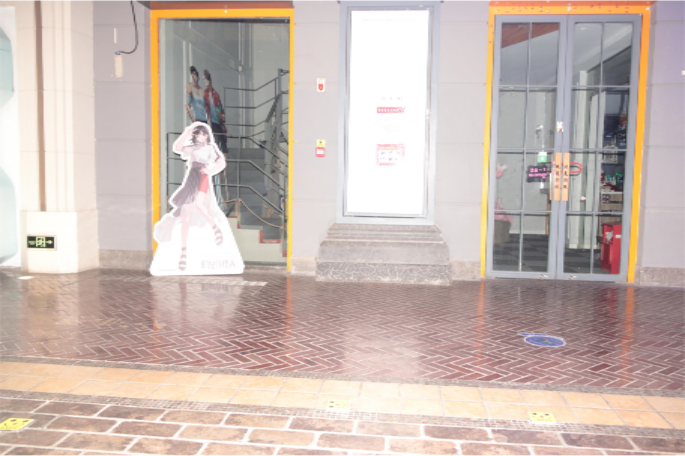}&
                \includegraphics[height=\newsubheight\linewidth, keepaspectratio]{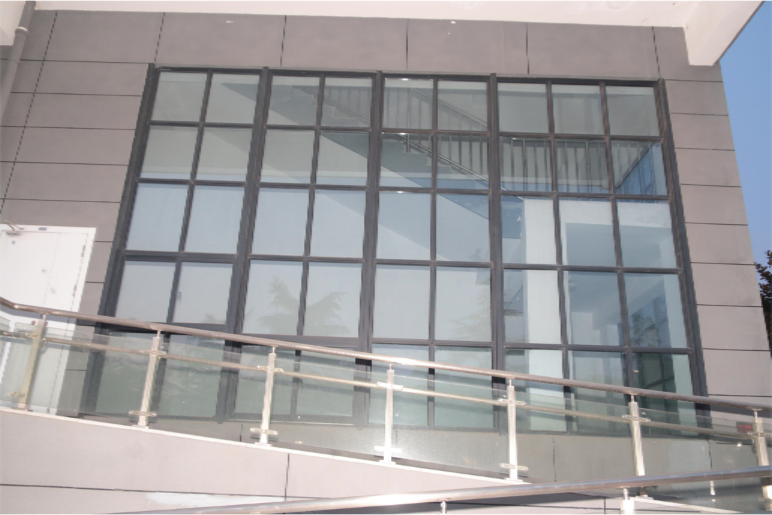}&
                \includegraphics[height=\newsubheight\linewidth, keepaspectratio]{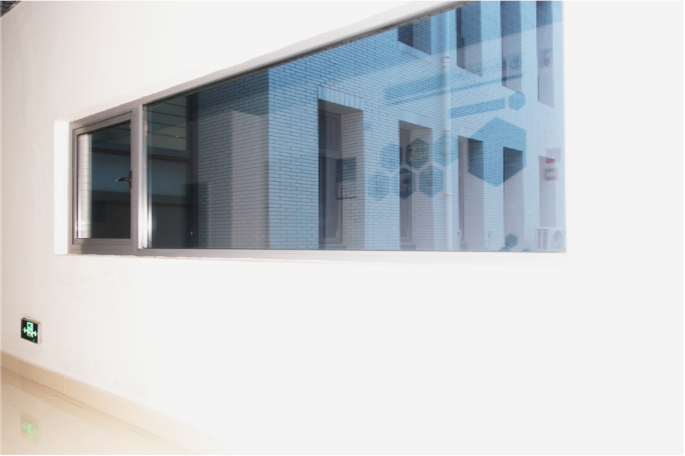}&
                \includegraphics[height=\newsubheight\linewidth, keepaspectratio]{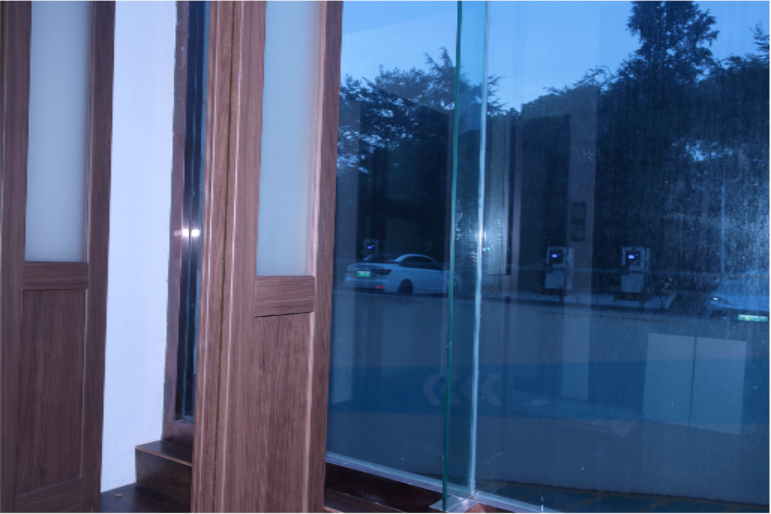}\\ 

                \multicolumn{2}{c}{\scriptsize{GSD~\cite{lin2021rich}}}&
                \multicolumn{2}{c}{\scriptsize{GSGD~\cite{Yan2025ghosting}}}&
                \multicolumn{4}{c}{\scriptsize{NFGD (ours)}}\\
                
        \end{tabular}
    \end{center}
 	\vspace{-0.016\textwidth}
    \caption{Comparison between GSD~\cite{lin2021rich}, GSGD~\cite{Yan2025ghosting} and our NFGD. \tao{Each no-flash and flash image pair from} our NFGD exhibits the appearance and disappearance \tao{phenomenon} of reflections on glass surfaces. \tao{Our NFGD encompasses a wide range of shots across various real-world scenes.}}
    \label{fig:examples_from_dataset}
\end{figure*}

\section{Related Work}
\subsection{Glass Surface Detection}
Mei~\etal~\cite{mei2020don} first proposed the glass surface detection problem and explored contextual glass features from a large receptive field. 
He~\etal~\cite{he2021enhanced} focused on glass boundary information and proposed a point-based graph convolution module to accurately distinguish between glass and non-glass regions. 
Yu~\etal~\cite{yu2022progressive} leveraged a mutual guidance approach between high-level and low-level features to obtain more accurate glass features for refining the glass mask. Lin~\etal~\cite{lin2021rich} considered the intrinsic properties of glass and first utilized reflection cues for glass surface detection.
Furthermore, Yan~\etal~\cite{Yan2025ghosting} \tao{proposed to leverage ghosting effect appeared on glass surfaces to accurately detect glass surfaces}.
Qi~\etal~\cite{qi2024glass} noted that Gaussian blurriness is a unique property of glass surfaces and utilized average pooling to simulate this property for glass surface detection. 
However, these previous methods based on single images may fail when these glass cues disappear (\eg, unclear glass boundaries, absence of reflections on the glass surface, image blur, \etc). 

Recently, Lin~\etal~\cite{DGSDNet} proposed an RGB-D glass surface detection method based on the phenomenon of depth missing in glass regions. 
Mei~\etal~\cite{mei2022polarization} proposed an RGB-P glass surface detection method based on the polarization cues of glass surfaces.
Lin~\etal~\cite{lin2022semantic} observed that glass surfaces are usually accompanied by walls or curtains, leading to a semantics-prior method (RGB-S) based on the contextual information around glass surfaces. 
Huo~\etal~\cite{huo2023glass} noted that the thermal conductivity of glass differs from that of other materials, thus utilizing an extra thermal sensor (RGB-T) to locate glass surface regions. 
Yan~\etal~\cite{yan2024nrglassnet} recognized that near-infrared can effectively suppress reflections on glass surfaces, \tao{and propose NRGlassNet, which takes as input} RGB-NIR \tao{image pairs} for glass detection. 
Although these multi-modal methods improve detection performance, additional sensors often incur high costs and require pixel alignment between the two modalities. 
Our method (RGB-Flash) \tao{can fully} utilize reflection cues from no-flash and flash image pairs to detect glass surfaces without extra costs and preprocessing.

\subsection{Transparent Object Detection}
Xie~\etal~\cite{xie2020segmenting,xie2021segmenting} proposed a convolution-based network named TransLab~\cite{xie2020segmenting} and a convolution union transformer-based network named Trans2Seg~\cite{xie2021segmenting} for transparent object detection.
They also constructed a large-scale dataset, Trans10K-V2~\cite{xie2021segmenting}, which includes two categories of transparent objects, stuff and things. Zhang~\etal~\cite{zhang2022trans4trans} developed a dual prediction head utilizing a shared transformer-based encoder and non-shared decoders for deploying transparent object detection on mobile devices. 
Zhu~\etal~\cite{zhu2021transfusion} and Liang~\etal~\cite{liang2023monocular} aimed to reconstruct the depth missing regions caused by transparent objects in depth maps. 
They proposed a boundary-aware method and a line segment-prior method, respectively, to detect transparent objects before depth reconstruction. 
Moreover, there are also methods that explore boundary cues to improve detection from polarization~\cite{kalra2020deep} and light field~\cite{xu2015transcut} images. 
All these methods achieve good performance in transparent object detection, as transparent objects typically possess easily observable geometric boundaries. 
However, glass surfaces usually do not have this property, thus require \tao{prior} effective cues to guide the detection of glass surfaces.

\subsection{Mirror \tao{D}etection}
Mei~\etal~\cite{yang2019my} first provided a large benchmark for mirror detection by utilizing the discontinuity of textures between mirror and non-mirror regions for mirror detection. 
Lin~\etal~\cite{lin2020progressive} noted the strict symmetry of objects inside and outside the mirror, detecting the mirror by exploring the similarity between adjacent feature cells in the feature space. 
Huang~\etal~\cite{huang2023symmetry} further observed the loose symmetry of mirrors due to specific perspectives, and they explored this loose symmetry by using the image and its mirrored version as inputs to enhance mirror detection. 
Guan~\etal~\cite{guan2022learning} noted that mirrors often appear alongside sinks or dressers, proposing semantics as a prior for mirror detection. 
Tan~\etal~\cite{tan2022mirror} leveraged the intrinsic visual properties of mirrors to facilitate mirror detection. 
\hhh{Zha~\etal~\cite{DPRNet2024} argued that the frequencies inside and outside the mirror are distinctive, and exploited low-frequency features to locate the mirror region and high-frequency features to refine detection results.} 
Xie~\etal~\cite{xie2024csfwinformer} further adopted the Discrete Cosine Transform (DCT) from a frequency perspective to extract mirror features across different frequency domains, improving the accuracy of mirror detection. 
However, unlike mirrors, glass \tao{surfaces do not} have clear \tao{boundaries}, and \tao{the reflection layer always mixed with corresponding transmission layer}, causing mirror detection methods \tao{being less effective in serving glass detection.}

\taore{\subsection{Flash and No-flash Image Pairs for Computer Vision Tasks}
Petschnigg \etal~\cite{petschnigg2004digital} proposed a variety of methods that analyze and combine the strengths of such flash/no-flash image pairs for image restoration and enhancement, such as Flash-to-ambient detail transfer and continuous flash intensity adjustment.
Flash and No-flash image pairs have also been leveraged in saliency detection~\cite{he2014saliency}, reflection removal~\cite{chang2020siamese}, and denoising images captured from low-light conditions~\cite{xia2021deep,xu2024laplacian}. 
Moreover, Cao \etal~\cite{cao2020stereoscopic} proposed a method for recovering shape and albedo of Lambertian surfaces from a stereo flash/no-flash pair.
Notably, Maralan \etal~\cite{maralan2023computational} proposed a physically motivated intrinsic formulation for flash photograph formation, and developed flash decomposition and generation methods for flash and no-flash photographs, respectively. 
However, this method exhibits limited generalization ability and struggles to effectively process flash and no-flash photographs in the presence of glass surfaces.}

\section{No-flash and Flash Glass Dataset \tao{(NFGD)}}
Lin~\etal~\cite{lin2021rich} and Yan~\etal~\cite{Yan2025ghosting} have proposed the GSD~\cite{lin2021rich} and GSGD~\cite{Yan2025ghosting} for glass surface detection, \tao{images from} which \tao{always contains prominent reflections on the glass surface}. However, not every image in GSD exhibits reflections on glass surfaces. Moreover, close-up shots capture double-layer reflections (ghosting effects) in GSGD. \tao{For learning our network}, we propose NFGD, \tao{which consists of $\sim$3.3k no-flash and flash image pairs along with corresponding annotations. Our NFGD} complements the scenes with a lack of reflections on glass surfaces in GSD, and offers a more varied range of shots (close-up, medium, and long shots) compared to GSGD. \taore{Several} examples from each dataset exhibited in Fig.~\ref{fig:examples_from_dataset} demonstrate the superiority of our NFGD.

\subsection{Dataset Construction}

\tao{No-flash and flash image pairs from our NFGD are} captured with a \tao{Canon 80D} DSLR camera \tao{from} various semantic scenes, including bathrooms, lounges, laboratories, hospitals, libraries, shopping malls, and schools. 
We also design a camera script to \taore{capture} a pair of no-flash and flash images within a very short time, \taore{ensuring} no pixel offset between the image pairs. To achieve greater variability in the illumination of images with flash, we captured images with different illuminations to enhance the robustness of our dataset. Specifically, we adjusted image brightness using varying flash intensities, aperture sizes, exposure times, and ISO settings. We also utilized an additional external hot-shoe flash (maximum brightness: $60$ GN at ISO $100$) to apply flash, using powers of $\times 1/2$, $\times 1/4$, and $\times 1/8$ to obtain different flash intensities. 
\taore{Our dataset encompasses scenes acquired under diverse ambient light conditions, including midday sunshine and overcast dusk. We prioritize adjusting the aperture size and exposure time to ensure the captured images accurately represent real-world observed brightness. This methodology enriches the dataset's scene diversity while preserving image fidelity.}

\renewcommand{\newsubwidth}{0.46}
\begin{figure}[h]
\centering
\subfloat[]{\includegraphics[width=1.4in]{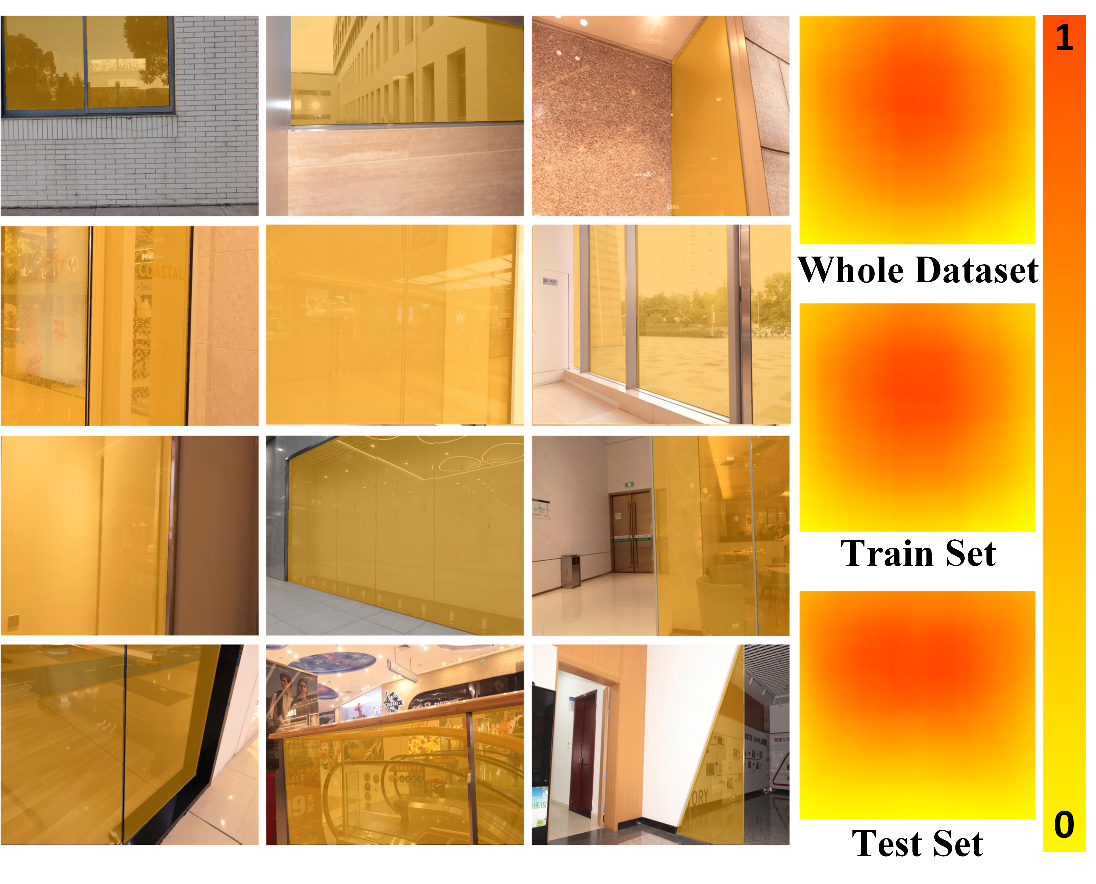}%
\label{fig:location_distribution}}
\hfil
\subfloat[]{\includegraphics[width=1.6in]{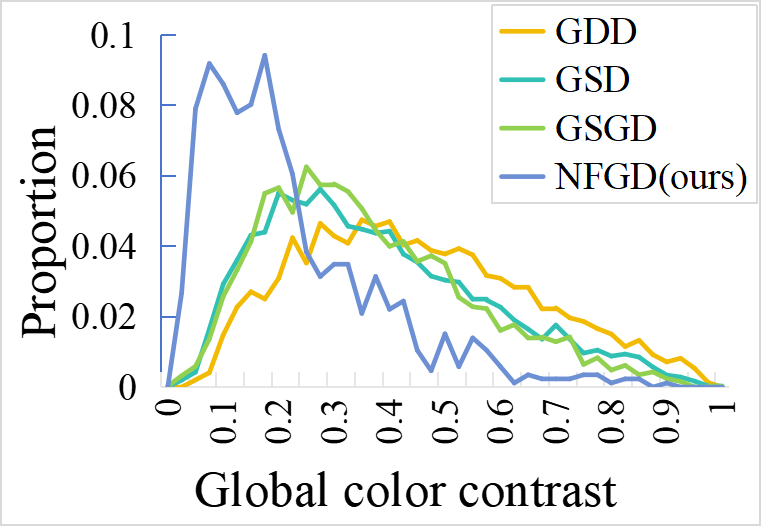}%
\label{fig:color_contrast}}
\caption{Statistics of our NFGD. (a) Glass surface location distribution. (b) Global color contrast.}
\label{fig:dataset_statistics}
\end{figure}

\subsection{Dataset Analysis}
\begin{itemize}
    \item Glass Location. The glass location distribution is the average of all glass surface regions in the dataset. We adopt the same method as~\cite{yu2022progressive,mei2022polarization} to evaluate our NFGD. As shown in Fig.~\ref{fig:location_distribution}, glass surfaces in our dataset \tao{primarily} concentrate at the middle and top regions, which is consistent in training and testing sets. Moreover, we employed shots from different distances and at various tilt angles to diversify the locations and shapes of glass instances, which could avoid the ``center bias" problem due to natural observation tendencies.

    \item \lyw{Contrast Distribution. As noted in \cite{lin2021rich}, lower color contrast between glass and non-glass regions implies a higher visual similarity, which in turn increases the difficulty of glass surface detection. To quantify this, we compute the $\chi^2$ distance to measure color contrast across multiple glass-surface datasets (GDD~\cite{mei2020don}, GSD~\cite{lin2021rich}, and GSGD~\cite{Yan2025ghosting}). Specifically, we use the no-flash images to calculate the global color contrast, as illustrated in Fig.~\ref{fig:color_contrast}. Overall, NFGD contains a larger proportion of images with low color contrast ($<0.3$), indicating that our dataset poses a more challenging detection scenario.}

\end{itemize}

\begin{figure*}[t]
\centerline{\includegraphics[width=0.82\textwidth]{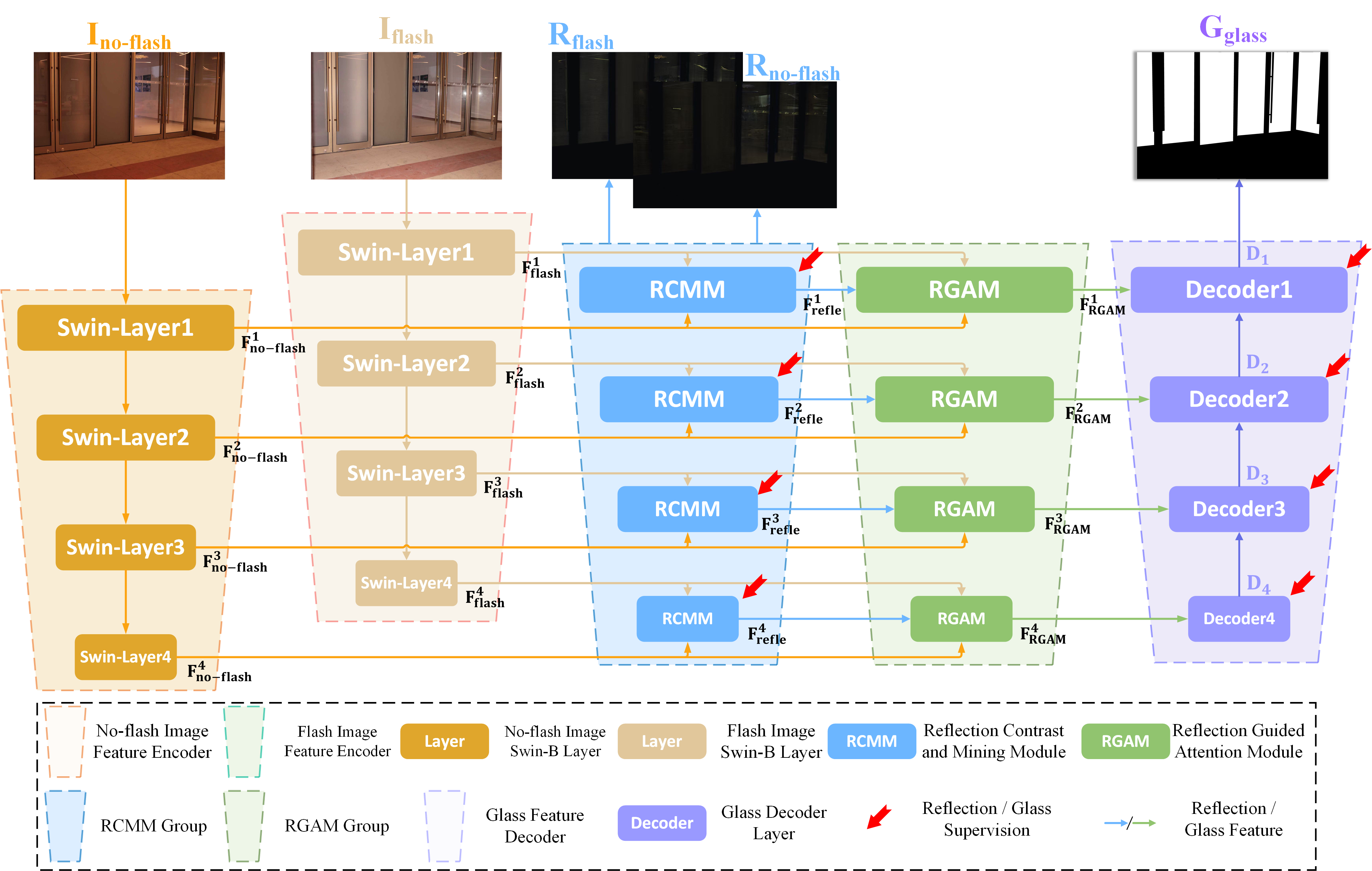}}
\caption{The architecture of our proposed \textit{\textbf{N}o-flash and \textbf{F}lash \textbf{Glass} Surface Detection \textbf{Net}work (\textit{NFGlassNet})}.}
\label{fig:NFGlassNet}
\end{figure*}

\section{Methodology}
In this section, we introduce our \textit{NFGlassNet} for glass surface detection \taore{with the aid of} cues from the appearance and disappearance \tao{phenomenon} of reflections caused by \tao{a} flash. In the first subsection, we present the overall pipeline of the network. In the second and third subsections, we \tao{elaborates on} our Reflection Contrast Mining Module (RCMM) and Reflection Guided Attention Module (RGAM), respectively.

\subsection{Overall Pipeline}
\taore{For multimodal glass surface detection~\cite{DGSDNet,mei2021depth,mei2022polarization,huo2023glass,yan2024nrglassnet}, mirror detection~\cite{mei2021depth,zhou2023utlnet} and salient object detection~\cite{HENet2024,DGPINet2024,LASNet2023,DepthInjection2023,EMANet2022,DeepDPS2023,CAINet2024}, images from auxiliary sensors often show significant semantic and textural differences compared to RGB counterparts.}
\taore{In each flash/no-flash image pair, semantic and texture information remains largely consistent, with the exception of reflection regions.}
Therefore, we do not need a complex encoder architecture like the multimodal mirror detection method~\cite{zhou2023utlnet} to fuse two modalities with significant differences in the encoder stage. 

As illustrated in the Fig.~\ref{fig:NFGlassNet}, our \tao{network} takes a pair of no-flash and flash images as input. 
\tao{It employs a pipeline consisting of encoder, extraction and fusion, and then decoder.} 
Due to the diverse shapes and varying sizes of different glass surfaces, a large-field contextual exploration is essentially needed to acquire a more precise \tao{and robust} detection of glass surfaces.
Consequently, we utilize Swin-Transformer V2~\cite{liu2022swin}  rather than ResNeXt~\cite{xie2017aggregated} \tao{as the backbone network}, as the former is better at modeling long-range dependencies.

Given \tao{the input no-flash and flash image pair} $\mathbf{I_{\text{no-flash}}} \in \mathbb{R}^{H\times W\times 3}$ and $\mathbf{I_{\text{flash}}} \in \mathbb
{R}^{H\times W\times 3}$, we extract feature sets $\mathbf{F_{\text{no-flash}}}$ and $\mathbf{F_{\text{flash}}}$ \tao{from them} using two identical Swin-Transformer V2~\cite{liu2022swin} backbones \tao{without sharing} weights.
Specifically, \tao{the definitions of} $\mathbf{F_{\text{no-flash}}} = \{\mathbf{F^\mathrm{i}_{\text{no-flash}}} \in \mathbb{R}^{\frac{H}{2^{i-1}} \times \frac{W}{2^{i-1}} \times (C \times 2^{i-1})} \mid i \in \{1,2,3,4 \}\}$ and $\mathbf{F_{\text{flash}}}$ \tao{are similar}.

The features $\mathbf{F_{\text{no-flash}}}$ and $\mathbf{F_{\text{flash}}}$ are \tao{then} fed into the \hh{RCMM group and RGAM group to extract and fuse features}. 
%
\hh{Each group contains four layers of \tao{corresponding} modules \tao{with different scales}, where the input feature size of each layer \tao{adapts} to the multi-scale features extracted from the encoder. In the RCMM group, }
$\mathbf{F_{\text{no-flash}}}$ and $\mathbf{F_{\text{flash}}}$ pass through the RCMM and yield the prediction of the reflection images $\mathbf{R}_\text{no-flash}$, $\mathbf{R}_\text{flash}$, and \tao{the} reflection features $\mathbf{F_{\text{refle}}}$. 
These reflection features, along with $\mathbf{F_{\text{no-flash}}}$ and $\mathbf{F_{\text{flash}}}$, are then sent into RGAM \hh{group}, where each RGAM adopts an optimized attention map to fuse reflection and glass features \tao{to obtain features more relevant to glass surfaces, denoted as} $\mathbf{F_{\text{RGAM}}}$.
\tao{Finally, a straightforward cascaded decoder is used to predict the final glass surface mask for the input image pair.}

\begin{figure}[ht]
\centerline{\includegraphics[width=0.46\textwidth]{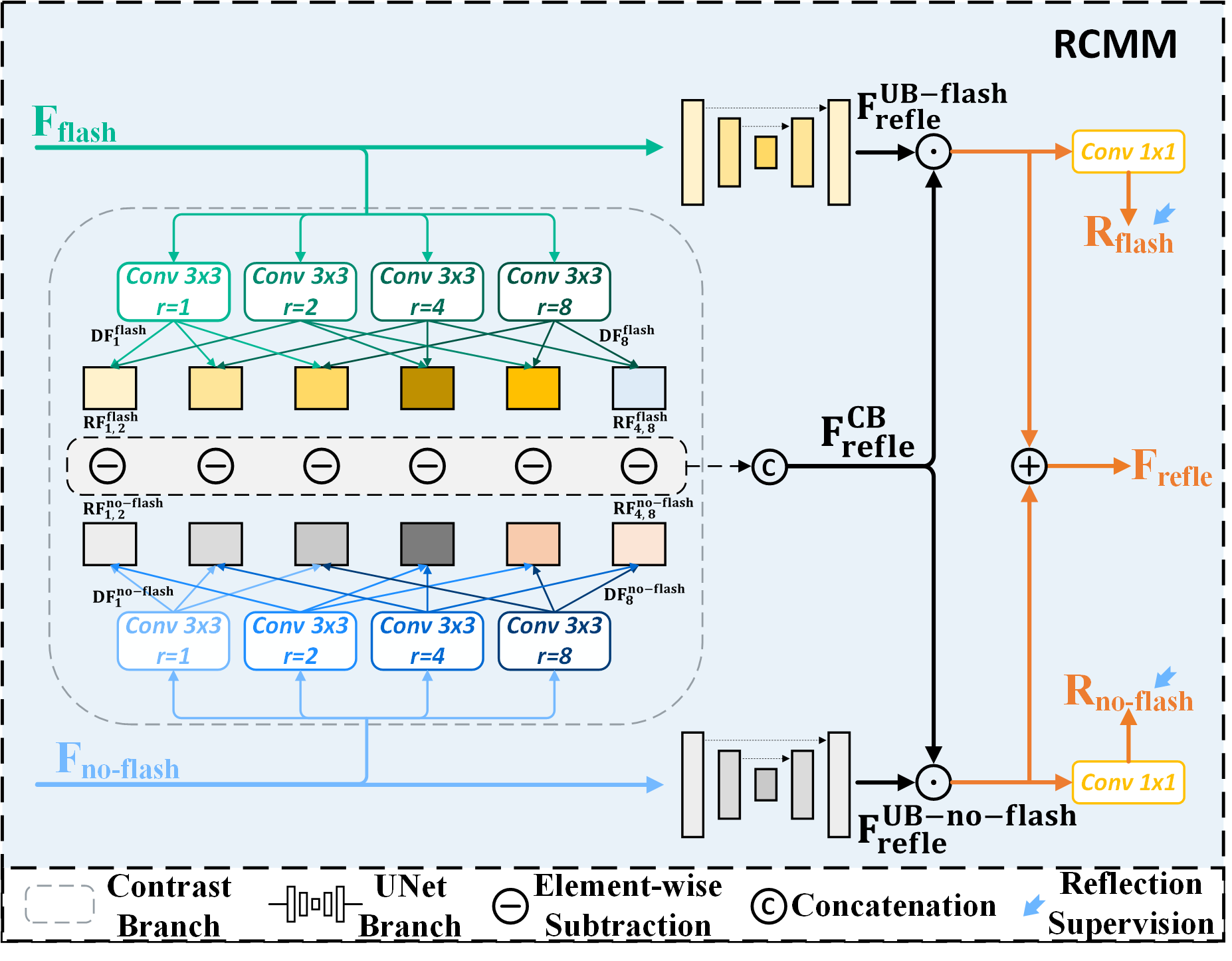}}
\caption{The architecture of the Reflection Contrast Mining Module (RCMM).}
\label{fig:RCMM}
\end{figure}

\subsection{Reflection Contrast Mining Module (RCMM)}
Fig.~\ref{fig:instrcution_result_show} shows that the distinct difference between the no-flash and flash images in the same scene lies in the appearance \tao{and} disappearance of reflections. 
Therefore, \tao{we propose RCMM to extract reflection from the input non-flash and flash image pair,} as shown in Fig.~\ref{fig:RCMM}. \tao{First,} a contrast branch in the middle of RCMM \tao{is used to} captures the \tao{contrast difference} (\ie, \tao{the} reflections) between the \tao{no-flash image and the flash image} by subtracting their feature maps. 
\tao{Second}, two UNet branches \tao{are employed} to extract reflections from each individual \tao{input} image to complement the reflection features \tao{extracted by} the contrast branch.

\tao{Reflections on glass surfaces don't have a fixed shape and exhibit a huge variation in scale. They can appear anywhere on the glass surface and blend with the transmission layer.} 
Therefore, it is necessary to \tao{capture potential reflection using multiple receptive fields with different sizes}. 
\tao{In our RCMM}, $\mathbf{F_{\text{no-flash}}}$ and $\mathbf{F_{\text{flash}}}$ are fed to \tao{an array of} dilated convolutions (denoted as $\mathrm{DConv}_{r_i}^{\text{flash/no-flash}}(\cdot)$) with different dilation rates $\mathbf{r_i} \in \{1,2,4,8\}$ to capture features. 
The extracted features $\mathbf{DF}_{r_i}^{\text{flash}}$ and $\mathbf{DF}_{r_i}^{\text{no-flash}}$ can be expressed as:
 \begin{align}
     \mathbf{DF}_{r_i}^{\text{flash}}    &= \mathrm{DConv}_{r_i}^{\text{flash}}(\mathbf{F_{\text{flash}}}), \\
     \mathbf{DF}_{r_i}^{\text{no-flash}} &= \mathrm{DConv}_{r_i}^{\text{no-flash}}(\mathbf{F_{\text{no-flash}}}).
 \end{align}
These features are then fused \taoR{via element-wise addition} according to their \taoR{pairwise combinations,} \tao{as shown in Fig.~\ref{fig:RCMM}}, to obtain richer representations from different receptive field sizes. \tao{The pairwise fused features from $\mathbf{F_{\text{flash}}}$ and $\mathbf{F_{\text{no-flash}}}$ are denoted as} $\mathbf{RF}_{r_i,r_j}^{\text{flash}}$ and $\mathbf{RF}_{r_i,r_j}^{\text{no-flash}}$, \tao{respectively, and} \tao{$\mathbf{RF}_{r_i,r_j}^{\text{flash}}$} can be \tao{defined} as:
\begin{equation}
    \mathbf{RF}_{r_i, r_j}^{\text{flash}} = \mathbf{DF}_{r_i}^{\text{flash}} + \mathbf{DF}_{r_j}^{\text{flash}},  (r_i, r_j) \in \binom{\{1, 2, 4, 8\}}{2}.
    \label{eq:rf}
\end{equation}
\tao{The definition for} $\mathbf{RF}_{r_i, r_j}^{\text{no-flash}}$ \tao{is similar}.

Subsequently, we subtract the \tao{features $\mathbf{RF}_{r_i, r_j}^{\text{flash}}$ and $\mathbf{RF}_{r_i, r_j}^{\text{no-flash}}$, which have the same dilation rate combination, in order to obtain differences. All differences from the corresponding dilation rate combination} are then concatenated to produce the reflection features of the contrast branch $\mathbf{F}_{\text{refle}}^{\text{CB}}$. This process can be written as:
\begin{equation}
    \mathbf{F}_{\text{refle}}^{\text{CB}} =
    \mathrm{Concat}\left( \left\{ \mathbf{RF}_{r_i, r_j}^{\text{flash}} - \mathbf{RF}_{r_i, r_j}^{\text{no-flash}} \right\} \right).
    \label{eq:f_contrast}
\end{equation}


\taoR{Notably, our contrast branch differs from RCAM proposed in~\cite{lin2021rich}. Specifically, RCAM initially employs a series of dilated convolutions with increasing dilation rates to extract multi-scale context features, and then directly computes contrasted features via element-wise subtraction across all permuted pairs of different context scales within a single image.
In contrast, our branch first fuses the outputs of different pair-wise dilated convolutions via element-wise addition to yield rich representations. It then derives contrasted features by performing element-wise subtraction between the corresponding rich representations of a pair of flash and no-flash images.}


\tao{On the other hand, we also directly extract reflections from $\mathbf{F_{\text{no-flash}}}$ and $\mathbf{F_{\text{flash}}}$ \tao{through the top and bottom branches of RCMM}.}
\tao{Specifically, inspired} by ~\cite{sun2019multi}, we employ a simple UNet~\cite{ronneberger2015u} (denoted as $\mathcal{U}^{\text{flash/no-flash}}(\cdot)$) to extract \tao{features of reflection} \tao{from $\mathbf{F_\text{flash}}$ and $\mathbf{F_{\text{no-flash}}}$ as follows:}
\begin{equation}
    \mathbf{F}_{\mathbf{\text{refle}}}^{\text{UB-flash}}=\mathcal{U}^{\text{flash}}(\mathbf{F_{\text{flash}}}),
    \label{eq:f_unet}
\end{equation}
and $\mathbf{F}_{\mathbf{\text{refle}}}^{\text{UB-no-flash}}$ follows the same process.

\tao{In the final stage, these two types of reflection features are fused to produce the output reflection feature as follows:}
\begin{equation}
    \begin{aligned}
        \mathbf{F}_{\mathbf{\text{refle}}} &= \mathbf{F}_{\text{refle}}^{\text{no-flash}} + \mathbf{F}_{\text{refle}}^{\text{flash}} \\
        &= (\mathbf{F}_{\text{refle}}^{\text{CB}} \odot \mathbf{F}_{\mathbf{\text{refle}}}^{\text{UB-flash}}) + (\mathbf{F}_{\text{refle}}^{\text{CB}} \odot \mathbf{F}_{\mathbf{\text{refle}}}^{\text{UB-no-flash}}).
    \end{aligned}
\end{equation}
Concurrently, we employ two 1×1 convolutional layers to extract corresponding reflection images $\mathbf{R}_\text{flash}$ and $\mathbf{R}_\text{no-flash}$ from the features $\mathbf{F}_{\mathbf{\text{refle}}}$ and $\mathbf{F}_{\mathbf{\text{no-refle}}}$ respectively.

\textbf{Pseudo Ground Truth for Reflection:}
To effectively supervise $\mathbf{F}_{\mathbf{\text{refle}}}$, we generate the pseudo ground truth, $\mathbf{\hat{R}}_\text{flash}$ and $\mathbf{\hat{R}}_\text{no-flash}$ \tao{for the $\mathbf{R}_\text{flash}$ and $\mathbf{R}_\text{no-flash}$, respectively}. The process diagram can be found in our \taore{supplemental}.
Existing SOTA reflection detection methods (\eg, LANet~\cite{dong2021location}) always assume that \tao{glass surfaces (or reflections) cover the entire image, \tao{while our method only cares about reflections on glass surfaces, \taore{which locate in local regions of the input image}. Thus, existing reflection detection methods} cannot be directly used to supervise the reflection cue extraction of our method, as shown in Fig.~\ref{fig:reflections_extracted_from_glass}.}
\tao{Specifically, we first} apply \taore{the ground-truth} glass mask to cover the input image pair, and then \tao{adopt} LANet~\cite{dong2021location} to obtain \tao{the} pseudo \taore{ground-truth reflections ($\mathbf{\hat{R}_{no-flash}}$, $\mathbf{\hat{R}_{flash}}$) from the masked no-flash and flash images}. 
%

\renewcommand{\newsubwidth}{0.158}
\begin{figure}[ht]
	\renewcommand{\tabcolsep}{0.8pt}
	\renewcommand\arraystretch{0.6}
        \begin{center}
            \begin{tabular}{cccccc}
                \includegraphics[width=\newsubwidth\linewidth]{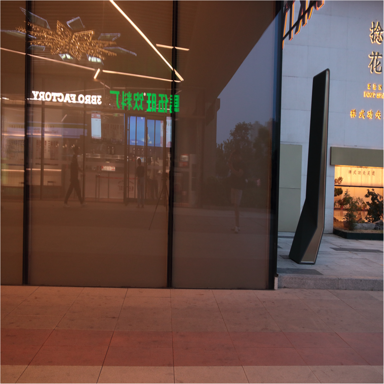}&
                \includegraphics[width=\newsubwidth\linewidth]{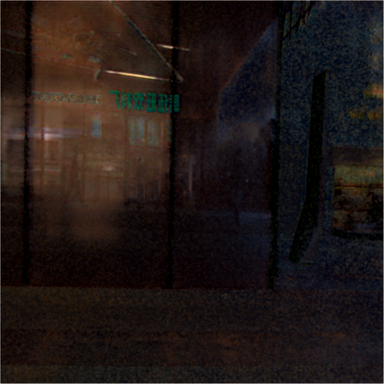}&
                \includegraphics[width=\newsubwidth\linewidth]{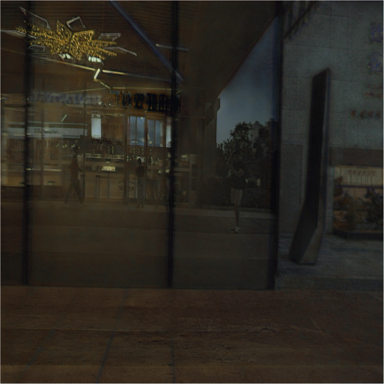}&
                \includegraphics[width=\newsubwidth\linewidth]{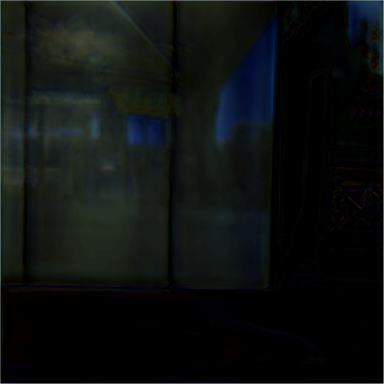}&
                \includegraphics[width=\newsubwidth\linewidth]{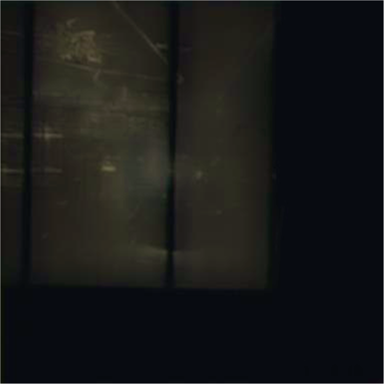}&
                \includegraphics[width=\newsubwidth\linewidth]{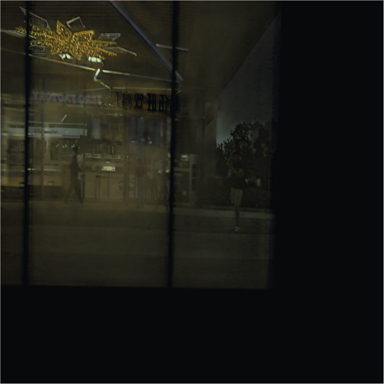}\\ 
  
                \includegraphics[width=\newsubwidth\linewidth]{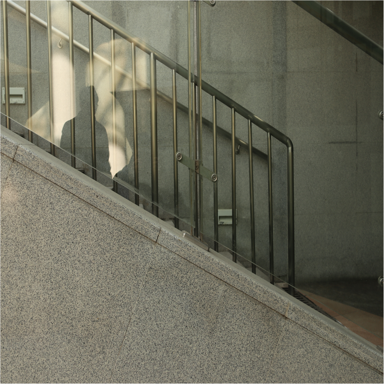}&
                \includegraphics[width=\newsubwidth\linewidth]{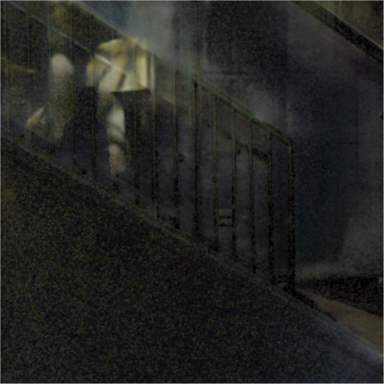}&
                \includegraphics[width=\newsubwidth\linewidth]{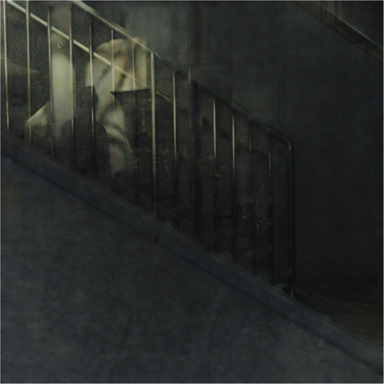}&
                \includegraphics[width=\newsubwidth\linewidth]{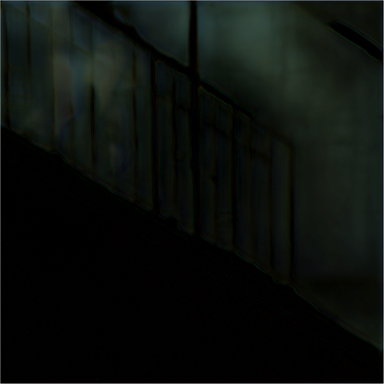}&
                \includegraphics[width=\newsubwidth\linewidth]{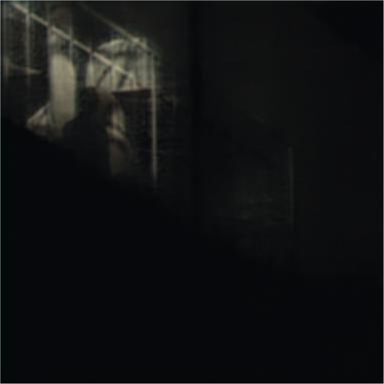}&
                \includegraphics[width=\newsubwidth\linewidth]{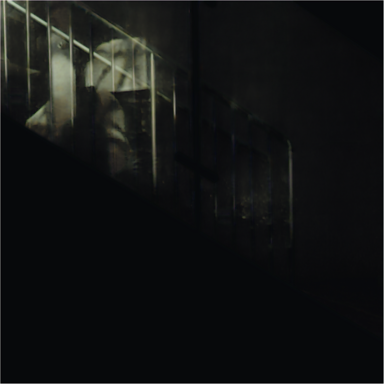}\\ 

               \fontsize{7.0pt}{\baselineskip}\selectfont{Image}&
                \fontsize{7.0pt}{\baselineskip}\selectfont{IBCLN~\cite{li2020single}}&
                \fontsize{7.0pt}{\baselineskip}\selectfont{LANet~\cite{dong2021location}}&
                \fontsize{7.0pt}{\baselineskip}\selectfont{GSDNet~\cite{lin2021rich}}&
                \fontsize{7.0pt}{\baselineskip}\selectfont{Ours}&
                \fontsize{7.0pt}{\baselineskip}\selectfont{Pseudo GT}\\
                
            \end{tabular}
        \end{center}
 	\vspace{-0.016\textwidth}
        \caption{Reflection extractions from two SOTA methods (IBCLN~\cite{li2020single} and LANet~\cite{dong2021location}) in \textit{SIRR}, and GSDNet~\cite{lin2021rich} in glass surface detection. Note that all three methods mis-detect non-glass regions as glass reflections, and reflections of GSDNet~\cite{lin2021rich} are too weak in real scenes. Our method \taore{is able to correctly} detect reflections in \taore{these} challenging scenes.}
        \label{fig:reflections_extracted_from_glass}
\end{figure}

\subsection{Reflection Guided Attention Module (RGAM)}

\begin{figure}[t]
\centerline{\includegraphics[width=0.42\textwidth]{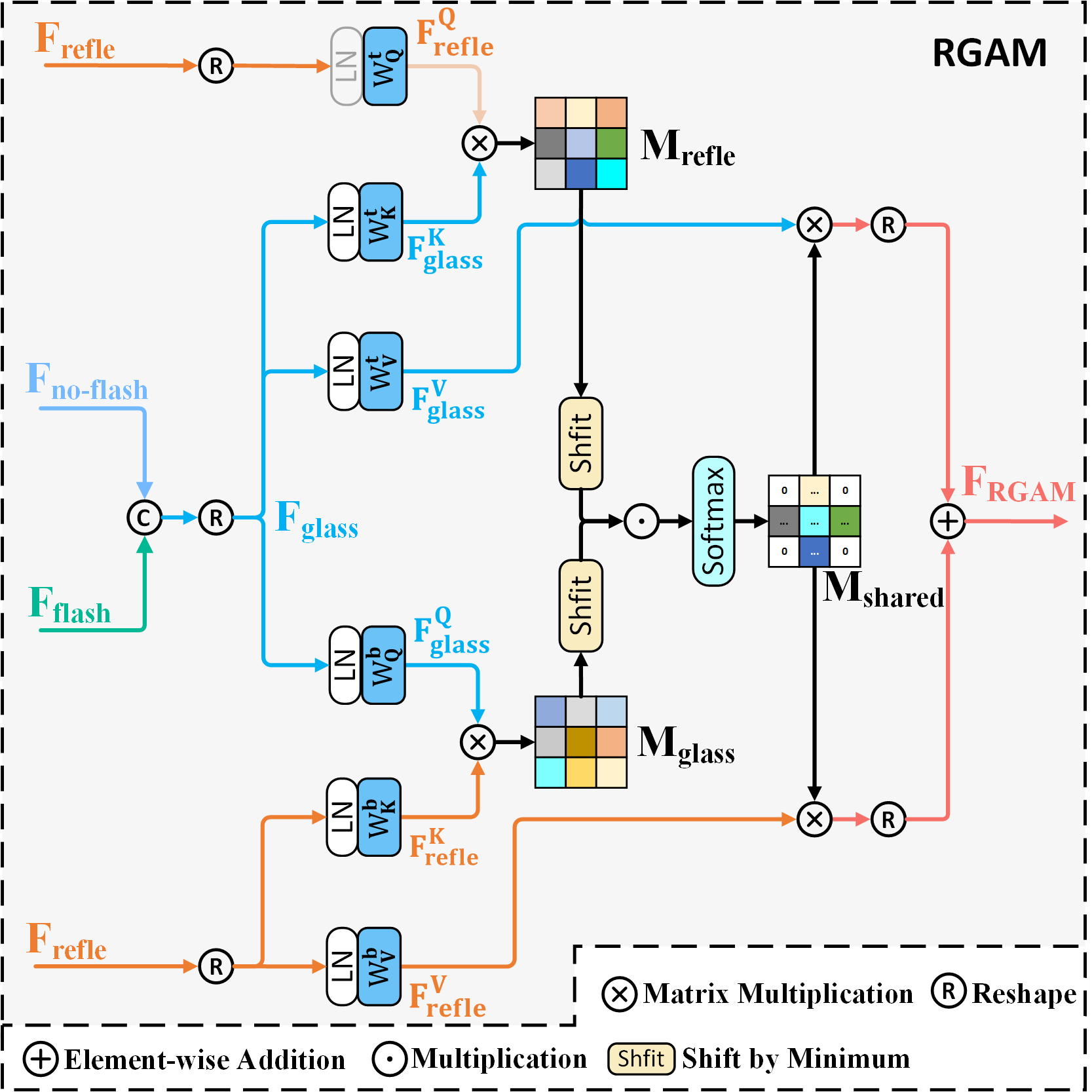}}
\caption{The architecture of our Reflection Guided Attention Module (RGAM).}
\label{fig:RGAM}
\end{figure}

\tao{As shown in Fig.~\ref{fig:RGAM}, our RGAM aims to fuse the extracted reflection features and features extracted from no-flash and flash images. Thus, \hh{to \tao{fully} explore the two \tao{kinds of} features}, we adopt dual cross-attention branches to independently search for reflection features ($\mathbf{F_{\text{refle}}}$) from the concatenated feature ($\mathbf{F_{\text{glass}}}$) of $\mathbf{F_{\text{no-flash}}}$ and $\mathbf{F_{\text{flash}}}$, and the concatenated features ($\mathbf{F_{\text{glass}}}$) from the reflection features ($\mathbf{F_{\text{refle}}}$). In this way, two attention maps named $\mathbf{M_{\text{refle}}}$ and $\mathbf{M_\text{glass}}$ can be obtained.}

\taoR{Reflections can occur on various smooth surfaces, and certain glass-like regions characterized by smooth surfaces and frames resembling glass borders are easily misidentified as actual glass.}
Therefore, the two attention maps, $\mathbf{M_{\text{refle}}}$ and $\mathbf{M_\text{glass}}$, obtained from the top and bottom branches (cross-attentions) \taoR{are multiplied to yield} the enhanced (\taoR{i.e.}, shared) attention map $\mathbf{M_{\text{shared}}}$.
This enhanced attention map is then multiplied back to each \taoR{branch to refine the respective features}. 
Finally, the \taoR{refined} features from both branches are added to \taoR{yield} the fused feature map, $\mathbf{F_{\text{RGAM}}}$.

Specifically, we employ \tao{two cross-attention in parallel}, which allows the two features to serve as the $\textit{query}$ ($\textbf{Q}$) in turn to ensure that the two query features (\ie, $\mathbf{F_{\text{refle}}}$ and $\mathbf{F_{\text{glass}}}$) are fully and equally explored. 
Unlike the serial single-branch structure used by ~\cite{lee2024guided} for feature fusion, our parallel dual cross-attention branches avoid insufficient exploration of features due to the order of the two $\textit{queries}$.

$\mathbf{F_{\text{glass}}}$ is the concatenation of $\mathbf{F_{\text{flash}}}$ and $\mathbf{F_{\text{no-flash}}}$, \tao{defined as}:
\begin{equation}
    \mathbf{F_{\text{glass}}}=\mathrm{Conv_{1 \times 1}}(\mathrm{Concat}(\mathbf{F_{\text{no-flash}}} + \mathbf{F_{\text{flash}}})),
\end{equation}
where $\mathrm{Conv_{1 \times 1}}$ denotes the $1 \times 1$ point-wise convolution to adjust channel numbers. 

\tao{Then, we} reshape $\mathbf{F_{\text{glass}}}$ and $\mathbf{F_{\text{refle}}} \in \mathbb{R}^{\hat{H} \times \hat{W} \times \hat{C}}$ to $\mathbb{R}^{\hat{H} \hat{W} 
\times N \times \frac{\hat{C}}{N}}$, where $N$ is the number of heads in the multi-head attention. 
After they pass through Layer Normalization ($\mathrm{LN}(\cdot)$), two sets of learnable matrices with the same structure but non-shared weights ($\mathrm{W_Q^\text{t/b}}$, $\mathrm{W_K^\text{t/b}}$, and $\mathrm{W_V^\text{t/b}}$) are applied to generate their respective tokens. $\mathbf{F_{\text{refle}}}$ and $\mathbf{F_{\text{glass}}}$ take turns as the \tao{$\textit{query}$} ($\textbf{Q}$). 
When $\mathbf{F_{\text{refle}}}$ serves as the \tao{$\textit{query}$} ($\textbf{Q}$), and $\mathbf{F_{\text{glass}}}$ \tao{serves} as the $\textit{key}$ ($\textbf{K}$) and $\textit{value}$ ($\textbf{V}$), this process can be described as:
\begin{align}
    \label{eq:f_refle}
    \mathbf{F_{\text{refle}}^\text{Q}} &= \mathrm{W_Q^\text{t}}(\mathrm{LN}(\mathrm{reshape}(\mathbf{F_{\text{refle}}}))), \\
    \mathbf{F_{\text{glass}}^\text{K}} &= \mathrm{W_K^\text{t}}(\mathrm{LN}(\mathrm{reshape}(\mathbf{F_{\text{glass}}}))), \\
    \mathbf{F_{\text{glass}}^\text{V}} &= \mathrm{W_V^\text{t}}(\mathrm{LN}(\mathrm{reshape}(\mathbf{F_{\text{glass}}}))).
\end{align}
\tao{While} \hh{$\mathbf{F_{\text{glass}}}$ acts as the \tao{$\textit{query}$}, the process of obtaining $\mathbf{F_{\text{glass}}^\text{Q}}$, $\mathbf{F_{\text{refle}}^\text{K}}$ and $\mathbf{F_{\text{refle}}^\text{V}}$ \tao{through the bottom cross-attention} follows the similar way.}

Next, \taoR{our method generates} attention maps $\mathbf{M_\text{refle}}, \mathbf{M_\text{glass}} \in \mathbb{R}^{N \times \hat{H}\hat{W} \times \hat{H}\hat{W}}$ for $\mathbf{F_{\text{refle}}}$ and $\mathbf{F_{\text{glass}}}$, respectively, \tao{which} can be written as:
\begin{align}
    \mathbf{M_\text{refle}} &= \mathbf{F_{\text{refle}}^\text{Q}} \otimes (\mathbf{F_{\text{glass}}^\text{K}})^\text{T}, \\
\mathbf{M_\text{glass}} &= \mathbf{F_{\text{glass}}^\text{Q}} \otimes (\mathbf{F_{\text{refle}}^\text{K}})^\text{T}.
\end{align}
Subsequently, a shift operation is applied to $\mathbf{M_\text{refle}}$ and $\mathbf{M_\text{glass}}$, respectively, where the minimum value is subtracted from all elements in the matrix \tao{in order} to shift \taore{values} to the positive range starting from $0$. After this, the two shifted attention maps are multiplied element-wise and \tao{then} normalized through the softmax function, which yields the final shared \tao{(enhanced)} attention map ($\mathbf{M_\text{shared}}$). This process can be expressed as:
\begin{align}
    \mathbf{M_\text{shared}} &= \mathrm{softmax}( (\mathbf{M_\text{refle}} - \mathrm{Min}(\mathbf{M_\text{refle}})) \notag \\
    &\quad \odot (\mathbf{M_\text{glass}} - \mathrm{Min}(\mathbf{M_\text{glass}}))),
    \label{eq:m_shared}
\end{align}
where $\mathrm{Min}(\cdot)$ denotes the minimum value of the matrix. After the shift operation, the feature cells corresponding to regions that are similar to $\textbf{Q}$ have values close to $1$, while those dissimilar to $\textbf{Q}$ have values close to $0$. 
\hh{Since the element-wise multiplication of two feature cells with values close to $1$ results in a value still near $1$, while the multiplication of cells close to $0$ results in values even closer to $0$, regions exhibiting both reflection and glass frame features will have their feature cell values enhanced, while regions lacking both features will be suppressed.}
As a result, our shared attention map is better at enhancing features and suppressing noise than the two respective attention map. 

Finally, we \tao{multiply} $\mathbf{M_\text{shared}}$ \tao{back} to $\mathbf{F_{\text{glass}}^\text{V}}$ and $\mathbf{F_{\text{refle}}^\text{V}}$, \tao{and} reshape the \tao{results} into $\mathbb{R}^{\hat{H} \times \hat{W} \times \hat{C}}$. Then, element-wise addition \tao{is performed} to obtain the more refined glass feature $\mathbf{F_\text{RGAM}}$ \tao{as follows}:
\begin{equation}
    \begin{aligned}
        \mathbf{F_\text{RGAM}} = &\ \mathrm{reshape}(\mathbf{M_\text{shared}} \otimes \mathbf{F_{\text{glass}}^\text{V}}) \\
    + &\ \mathrm{reshape}(\mathbf{M_\text{shared}} \otimes \mathbf{F_{\text{refle}}^\text{V}}).
    \end{aligned}
\end{equation}

\subsection{Decoder}
\tao{In order to} accurately decode \taore{features ($\mathbf{F_\text{RGAM}}$) output by RGAMs} to obtain the final refined glass mask, we propose a progressive decoder. Since the glass features have been sufficiently explored and fused in RCMM and RGAM, the decoder should be as simple as possible. Similar to GSDNet~\cite{lin2021rich} and PGSNet~\cite{yu2022progressive}, we upsample ($\mathrm{Up(\cdot)}$) deep features and concatenate them with shallow features along the channel dimension to decode each layer's features, then \tao{pass} the decoded features to even shallower layers. The process at each decoder layer ($\mathbf{D_\mathrm{i}}$) can be represented as:
\begin{equation}
    \mathbf{D_\mathrm{i}} = 
    \begin{cases}
        \mathrm{Conv}(\mathrm{Concat}(\mathrm{Up}(\mathbf{D_\mathrm{i+1}}), \mathbf{F_{RGAM}^\mathrm{i}}))  &\mathrm{i}=1,2,3 \\
        \mathrm{Up}(\mathbf{F_{RGAM}^\mathrm{4}}) &\mathrm{i}=4,
    \end{cases}
\end{equation}
where $\mathrm{Conv}(\cdot)$ denotes a $3 \times 3$ convolution. Then, we obtain the glass mask $\mathbf{G}_\mathrm{i} = \mathrm{Conv_{1 \times 1}}(\mathbf{D}_\mathrm{i})$ for $\mathrm{i} \in \{1, 2, 3, 4\}$ layers of the Decoder. Finally, we upsample and adjust the channels of $\mathbf{D_\mathrm{1}}$ to acquire the final glass mask ($\mathbf{G_\text{glass}}$) as follows:
\begin{equation}
    \mathbf{G_\text{glass}}=\mathrm{Conv}(\mathrm{Up(\mathbf{D}_\mathrm{1})}).
\end{equation}


\subsection{Loss Functions}
Our loss function consists of supervision for the glass and reflection. 
The total loss $\mathcal{L}$ can be expressed as:
\begin{equation}
    \mathcal{L} = \sum_{\mathrm{i}=1}^{4}(\mathcal{L}_{glass}(\mathbf{G}_\mathrm{i}, \mathbf{\hat{G}}_\mathrm{i}) +
    \lambda \sum_{\mathrm{j}\in \mathrm{S}}\mathcal{L}_{refle}(\mathbf{R}_\mathrm{i}^\mathrm{j}, \mathbf{\hat{R}}_\mathrm{i}^\mathrm{j})),
\end{equation}
where $\lambda$ is a hyper-parameter, $\mathrm{S} = \{\text{flash, no-flash}\}$, and $\mathbf{R}_\mathrm{i}^\mathrm{j}$ represents the reflection prediction decoded from $\mathbf{F}_{\text{refle}}^{\text{flash}}$ or $\mathbf{F}_{\text{refle}}^{\text{no-flash}}$ at each layer like CeilNet~\cite{fan2017generic}. 
To supervise the reflection result, we adopt the $\mathrm{SSIM}$ loss and $L_1$ loss, as:
\begin{equation}
    \mathcal{L}_{refle}(\mathbf{R}_\mathrm{i}^\mathrm{j}, \mathbf{\hat{R}}_\mathrm{i}^\mathrm{j}) = 1-\mathrm{SSIM}(\mathbf{R}_\mathrm{i}^\mathrm{j}, \mathbf{\hat{R}}_\mathrm{i}^\mathrm{j}) + \Vert \mathbf{R}_\mathrm{i}^\mathrm{j}-\mathbf{\hat{R}}_\mathrm{i}^\mathrm{j} \Vert_1.
\end{equation}
In order to obtain accurate glass surface regions and boundaries, we utilize $\mathrm{IoU}$ loss~\cite{qin2019basnet} and $\mathrm{BCE}$ loss~\cite{de2005tutorial} for glass supervision, as:
\begin{align}
    \mathcal{L}_{glass}(\mathbf{G}_\mathrm{i}, \mathbf{\hat{G}}_\mathrm{i}) 
    =& 1 - \frac{\mathbf{G}_\mathrm{i} \mathbf{\hat{G}}_\mathrm{i}}{\mathbf{G}_\mathrm{i} + \mathbf{\hat{G}}_\mathrm{i} - \mathbf{G}_\mathrm{i} \mathbf{\hat{G}}_\mathrm{i}} \notag \\
    &-( \mathbf{\hat{G}}_\mathrm{i} \log{\mathbf{G}_\mathrm{i}} + (1 - \mathbf{\hat{G}}_\mathrm{i}) \log(1 - \mathbf{G}_\mathrm{i})).
\end{align}

\section{Experiments}
\subsection{Experimental Settings}
\subsubsection{Implementation Details}
We train our model using PyTorch on an NVIDIA RTX 4090 GPU (24GB). We resize the no-flash and flash images to $384 \times 384$ for input and use Swin Transformer V2~\cite{liu2022swin} pre-trained on ImageNet\hh{~\cite{deng2009imagenet}} as our backbone network. We use the AdamW~\cite{LoshchilovH19} optimizer with an initial learning rate of 1e-5. The batch size is set to $2$, and the hyper-parameter $\lambda$ is set to $0.8$. We train our network for $150$ epochs, which takes approximately $25$ hours. The inference time \tao{for each} image pair is around $0.2$ seconds. To avoid overfitting, we apply the same data augmentation methods \hh{(\eg, random cropping, horizontal flipping, rotation)} as \tao{the} previous \tao{work}~\cite{lin2021rich} during training. Notably, we do not use CRF~\cite{krahenbuhl2011efficient} for post-processing of the inferred images, as our method focuses more on correctly identifying glass and non-glass regions.

\subsubsection{Evaluation Metrics}
For evaluating our network with other competing methods comprehensively, we adopt five widely used metrics to assess the glass detection performance, \tao{including} intersection over union (IoU), F-measure \tao{(F$_{\beta}$)}, mean absolute error (MAE), balanced error rate (BER), and pixel accuracy (ACC). \hh{The detailed formulations of these metrics can be further found in the supplementary material.}

\subsection{Comparison with the State-of-the-art Methods}

\renewcommand{\tabcolsep}{3.0pt}
\renewcommand\arraystretch{1}
\begin{table*}[ht]
  \caption{Quantitative comparison between our method and $19$ State-of-the-art methods for Glass Surface Detection, Salient Object Detection, and Mirror Detection. We report IoU, F$_\beta$,  MAE, BER, and Acc \tao{values}. Best and second-best results are marked in \Best{Red} and \SecondBest{Cyan}, respectively.} 
  \label{tab:results_subsets}
  \vspace{-3mm}
  \centering
  \begin{tabular}{cllccccccc}
    \toprule
    {Task} &Method &Year &IoU  $\uparrow$   &F$_\beta$  $\uparrow$       &MAE  $\downarrow$     &BER $\downarrow$    &ACC $\uparrow$  &Params(M) &FLOPs(G) \\
    \midrule
    \multirow{5}{*}{Glass Detection (RGB-Only)}  
    &Translab~\cite{xie2020segmenting}  &ECCV’20  &77.14   &0.869 &0.101 &0.121  &0.873 &42.19   &49.26  \\ 
    &GDNet~\cite{mei2020don}  &CVPR’20  &78.83   &0.872 &0.097 &0.115  &0.881   &207.95 &231.50   \\
    &GSDNet~\cite{lin2021rich}  &CVPR’21  &80.21   &0.887 &0.094 &0.108  &0.887   &83.71  &92.86  \\
    &EBLNet~\cite{he2021enhanced}  &ICCV’21  &79.92   &0.878 &0.096 &0.115  &0.885   &111.71 &322.16  \\
    &GhostingNet~\cite{Yan2025ghosting}  &TPAMI’25  &81.98   &0.898 &0.088 &0.096  &0.899 &280.33   &322.33  \\
    \midrule
    \multirow{5}{*}{Glass Surface Detection (Multimodal)}  
    &DGSDNet (RGB-D)~\cite{DGSDNet}  &AAAI’25  &81.14   &0.890 &0.092 &0.091  &0.890   &80.54 &35.06  \\
    &GlassSemNet (RGB-S)~\cite{lin2022semantic}  &NeurIPS’22  &81.37   &0.894 &0.091 &0.098  &0.892   &361.33 &1412.03  \\
    &PGSNet (RGB-P)~\cite{mei2022polarization}  &CVPR’22  &80.09   &0.881 &0.094 &0.112  &0.886   &302.84  &291.42 \\
    &RGB-T~\cite{huo2023glass}  &TIP’23  &{83.93}   &{0.909} &0.083&0.095  &{0.912}  &81.25   &42.68  \\
    &NRGlassNet (RGB-NIR)~\cite{yan2024nrglassnet}  &KBS’24  &82.85   &0.899 &0.088 &0.085  &0.903   &203.37  &195.07 \\
    &NightGSD (RGB-NIR)~\cite{yan2026when}  &TMLR’26 &\SecondBest{84.73}   &0.908  &  0.090 & 0.093 &\Best{0.925}  & 234.88 & 469.98 \\
    \midrule
    \multirow{5}{*}{SOD (Multimodal)}
    &SPNet (RGB-D)~\cite{zhou2021specificity} &ICCV'21 &82.68 &0.897 &0.084 &0.089 &0.898 &81.05 &202.74 \\
    &CIRNet (RGB-D)~\cite{cong2022cir} &TIP'22 &82.93 &0.894 &0.105 &0.081 &0.903 &103.16 &42.60 \\
    &WaveNet (RGB-T)~\cite{WaveNet} &TIP'23 &83.74 &0.902 &\SecondBest{0.082} &0.080 &0.909 &84.88 &64.02 \\
    &HENet (RGB-D)~\cite{HENet2024} &TCSVT'24 &74.32 &0.849 &0.156 &0.145 &0.868 &10.43 &10.75 \\
    &LASNet (RGB-T)~\cite{LASNet2023} &TCSVT'23 &77.46 &0.875 &0.114 &0.118 &0.879 &93.57 &111.69 \\
    \midrule
    \multirow{2}{*}{Mirror Detection}  
    &PMDNet~\cite{lin2020progressive}  &CVPR’21  &80.95   &0.886 &0.093 &0.103  &0.891 &147.66   &119.27  \\
    &CSFwinformer~\cite{xie2024csfwinformer}  &TIP’24  &83.86   &0.903 &0.096 &\SecondBest{0.079} &0.904 &188.62   &209.83  \\
    \midrule
    \multirow{1}{*}{Semantic Segmentation}
    &SAM3 (zero-shot)~\cite{carion2025sam3}& ICLR'26 & 82.71 & \SecondBest{0.924} & 0.107&0.099&0.897&840.51& 5036.94\\
    \midrule
    \multirow{1}{*}{Glass Surface Detection}
    &Ours (RGB-Flash)  &-  &\Best{86.46}   &\Best{0.926} &\Best{0.080} &\Best{0.076} &\SecondBest{0.922} &208.43 &140.20  \\
    \bottomrule
  \end{tabular}
  \vspace{-3mm}
\end{table*}

\renewcommand{\tabcolsep}{4.0pt}
\renewcommand\arraystretch{1}
\begin{table}[t]
\caption{
Study of the \tao{illumination} intensity effects on the seven methods that performed well in Tab.~\ref{tab:results_subsets}. We replaced the original no-flash image with the flash image as input \tao{of the competing methods~\cite{lin2021rich,Yan2025ghosting,xie2024csfwinformer}}. \tao{For our method and the \tao{multimodal image-based} methods~\cite{huo2023glass,WaveNet,yan2026when}, we exchange the roles of the two images of each input no-flash and flash image pair.} Results that are better than the corresponding \tao{values exhibited} in Tab.~\ref{tab:results_subsets} are marked in \Better{cyan}.
}
\label{tab:results_reverse_input}
  \vspace{-2mm}
  \centering
    \begin{tabular}{clccccc}
    \toprule
     {Task} &Method  &IoU$\uparrow$ &F$_\beta$$\uparrow$ &MAE$\downarrow$ &BER$\downarrow$ &ACC$\uparrow$ \\
    \midrule
     \multirow{3}{*}{GSD}
     &GSDNet~\cite{lin2021rich} &78.61 &0.871 &0.098 &0.119 &0.878 \\
    &GhostingNet~\cite{Yan2025ghosting} &81.53 &0.894 &0.092 &0.097 &0.892 \\
    &NightGSD~\cite{yan2026when} &84.10   &0.903 &  0.097 & 0.101 &0.925\\
    \midrule
    \multirow{2}{*}{SOD} &RGB-T~\cite{huo2023glass} &83.41 &0.901 &0.097 &0.089 &0.908 \\
    &WaveNet~\cite{WaveNet} &83.26 &0.901 &0.098 &0.088 &0.905 \\
    \midrule
    \multirow{1}{*}{MD} &CSFWinformer~\cite{xie2024csfwinformer} &\Better{83.98} &\Better{0.912} &\Better{0.095}&\Better{0.076}&\Better{0.906} \\
    \midrule
    \multirow{1}{*}{SS} &SAM3 (zero-shot)~\cite{carion2025sam3} &\Better{84.05} &\Better{0.929} &\Better{0.099} &\Better{0.091} &\Better{0.905} \\
    \midrule
    \multirow{1}{*}{GSD} &Ours &86.14 &0.923 &0.080 &0.081 &0.922 \\
    \bottomrule
  \end{tabular}
  \vspace{-2mm}
\end{table}

\subsubsection{Quantitative Comparison}
As shown in Tab.~\ref{tab:results_subsets}, $19$ State-of-the-art (SOTA) methods \tao{are chosen} from the past five years for \tao{comparing with our method}, including $5$ \tao{single image-based} glass surface detection (GSD) methods~\citep{xie2020segmenting,mei2020don,lin2021rich,he2021enhanced,Yan2025ghosting}, $6$ \tao{multimodal image-based} GSD methods~\citep{mei2021depth,lin2022semantic,mei2022polarization,huo2023glass,yan2024nrglassnet, yan2026when}, $5$ Salient Object Detection (SOD) methods~\citep{zhu2021transfusion,cong2022cir,WaveNet,HENet2024,LASNet2023}, $2$ Mirror Detection (MD) methods~\citep{lin2020progressive,xie2024csfwinformer} and $1$ Semantic Segmentation (SS) vision foundation model~\citep{carion2025sam3}. We retrained these methods on our dataset with their codes, models, and corresponding training strategies to ensure a fair comparison. For all multi-modal methods \tao{listed} in Tab.~\ref{tab:results_subsets}, we use the no-flash image as the RGB input and the flash image as the \tao{image from another modality}. For the \tao{other} comparison methods, we only use the no-flash image as the input in the training and inference process. \lyw{For the most recently proposed vision foundation model SAM3~\citep{carion2025sam3}, we use the text prompt `glass'.}
\lywre{The quantitative comparison results demonstrate that our method achieves state-of-the-art performance across all evaluation metrics. Notably, our method outperforms previous glass surface detection methods while maintaining considerably lower computational complexity than NightGSD. Specifically, it ranks first in IoU (86.46\%), $F_{\beta}$ (0.926), MAE (0.052), and BER (0.053), while achieving the second-best ACC score (0.922), which is only 0.3\% lower than NightGSD (0.925).}
\tao{The qualitative results of our method and competing methods evaluated on two multimodal datasets: RGB-T~\citep{huo2023glass} and RGB-NIR~\citep{yan2024nrglassnet} are shown in our supplementary material.}

To validate whether \tao{good} illumination intensity \tao{can improve} the detection performance of the \tao{competing} methods, we select \lywre{seven} methods \tao{that} perform well \tao{across} different tasks. As shown in Tab.~\ref{tab:results_reverse_input}, \tao{for evaluating} GSDNet~\cite{lin2021rich}, GhostingNet~\cite{Yan2025ghosting}, CSFwinformer~\cite{xie2024csfwinformer} and \lywre {SAM3}~\cite{carion2025sam3}, we conduct experiments using only the flash image as input. For RGB-T~\cite{huo2023glass}, \lywre{NightGSD}~\cite{yan2026when}, \tao{and} \taoq{WaveNet~\cite{WaveNet}}, the input setting is reversed compared to \tao{that of} Tab.~\ref{tab:results_subsets}, where we use the flash image as the RGB input and the no-flash image as the \tao{image from another modality}. For our \textit{NFGlassNet}, we reverse the input setting similarly. 
\lywre{The results show that simply replacing the original no-flash image with the flash image does not consistently improve the competing methods, and in most cases even degrades performance. In contrast, our method remains robust when exchanging the roles of the flash and no-flash images, indicating that the proposed module can effectively leverage information from both images regardless of their input order.}

\lyw{In addition, we evaluate our method on the single-image glass-surface detection benchmarks GDD~\cite{mei2020don} and GSD~\cite{lin2021rich}, with the results summarized in Table \ref{tab:result_GDD_GSD}. For fairness, we retrained all the methods on our device using the same training parameters and we retained the dual-branch network structure, where both branches take the same RGB image as input.}
\lywre{Although several recent methods specifically designed for single-image glass surface detection achieve higher performance, our method consistently produces competitive results across both datasets, outperforming earlier glass surface detection methods. These results indicate that the proposed architecture generalizes well beyond the paired-image setting and does not rely solely on flash information to achieve strong performance.}

\taore{
Moreover, we also evaluate the performance of our network on other multimodal datasets. Specifically, Tab.~\ref{tab:results_RGBNIR} shows the quantitative results of our method and competing methods evaluated on two multimodal datasets: RGB-T~\cite{huo2023glass} and RGB-NIR~\cite{yan2024nrglassnet}. It demonstrated that our \textit{NFGlassNet} ranks second-best in these two datasets. The performance of our network is only weaker than that of the network specifically designed for the particular dataset. We have only selected a few methods that perform well on our dataset (exhibited in Tab.~\ref{tab:results_subsets}) as the competing methods in this study.
}

\begin{table*}
\centering
\caption{Quantitative comparison between our method and $10$ state-of-the-art methods on the single-image glass surface detection (GSD) datasets, GDD and GSD. \taore{TOD denotes transparent object detection.} Best detection results are marked in \Best{red}.}
\begin{tabular}{clcccccccccccc}
\toprule
\multirow{2.45}{*}{\centering Task} &\multirow{2.45}{*}{Method} &\multirow{2.45}{*}{Venue} &\multicolumn{5}{c}{GDD\cite{mei2020don}}&&\multicolumn{5}{c}{GSD\cite{lin2021rich}} \\
\cmidrule{4-8}  \cmidrule{10-14}
&&&IoU$\uparrow$ &$F_\beta$$\uparrow$ &MAE$\downarrow$ &BER$\downarrow$ &ACC$\uparrow$ &&IoU$\uparrow$ &$F_\beta$$\uparrow$ &MAE$\downarrow$ &BER$\downarrow$ &ACC$\uparrow$ \\
\midrule
GSD & GDNet\cite{mei2020don} & CVPR'20& 81.47&0.895&0.098&0.087&0.919 && 76.77&0.864&0.076&0.097&0.882\\
GSD & GSDNet\cite{lin2021rich} &CVPR'21& 88.07&0.932&0.059&0.057&{0.949} &&83.67&0.903&0.055&0.061&{0.931}\\
GSD & EBLNet\cite{he2021enhanced} & ICCV'21& 88.72&0.940&0.055&0.053&0.944 && 82.34&0.889&0.064&0.073&0.911\\
GSD & RFENet\cite{fan2023RFENet} & IJCAI'23& 86.93&0.928&0.067&0.065&0.932 && 83.59&{0.904}&{0.049}&0.064&0.914\\
GSD & GhostingNet\cite{Yan2025ghosting} & TPAMI'25&{89.30}&{0.943}&{0.054}&{0.051}&0.944 && {83.77}&{0.904}&0.055&{0.061}& {0.928}\\
GSD & GlassWizard~\cite{li2025glasswizard}&ICCV'25&\Best{92.10}&\Best{0.961}&\Best{0.041}&\Best{0.039} &0.959&&\Best{89.10}&\Best{0.942}&\Best{0.035}&\Best{0.041}&\Best{0.952}\\
GSD &  MSNet~\cite{cheng2026msnet}&AAAI'26&91.50&0.955&0.043&0.041&- & &87.80&0.916&0.042&0.047&-\\
GSD & NightGSD (RGB-NIR)~\cite{yan2026when} &TMLR'26 & {91.27} & {0.946} & {0.044} & {0.045} &\Best{0.963}&& {87.91} & 0.921 &{0.039} &{0.047}& 0.935\\

\midrule
GSD & ours & - & {90.18} &{0.946} &{0.052} & {0.047} & {0.958} && {86.16} & {0.924} & {0.044} & {0.056} & 0.927\\
\bottomrule
\end{tabular}

\label{tab:result_GDD_GSD}
\end{table*}


\begin{table*}[t]
\caption{Qualitative results of our method and the competing methods evaluated on the RGB-T dataset~\cite{huo2023glass} and the RGB-NIR dataset~\cite{yan2024nrglassnet}. The Best and second-best results are marked in \Best{Red} and \SecondBest{Cyan}, respectively.} 
\label{tab:results_RGBNIR}
\centering 
\begin{tabular}{clcccccccccccc} 
\toprule 
\multirow{3}{*}{Modality} &\multirow{3}{*}{Method} &\multirow{3}{*}{Task} &\multicolumn{5}{c}{RGB-T~\cite{huo2023glass}}&&\multicolumn{5}{c}{RGB-NIR~\cite{yan2024nrglassnet}} \\
 \cmidrule{4-8}  \cmidrule{10-14}
  &&&IoU$\uparrow$ &F$_\beta$$\uparrow$ &MAE$\downarrow$ &BER$\downarrow$ &ACC$\uparrow$ &&IoU$\uparrow$ &F$_\beta$$\uparrow$ &MAE$\downarrow$ &BER$\downarrow$ &ACC$\uparrow$ \\
\midrule 
\multirow{3}{*}{RGB-T}
&RGB-T~\cite{huo2023glass} &GSD &\Best{87.13} &\Best{0.911} &\Best{0.026} &\Best{0.037} &\Best{0.895}&&85.36 &0.926 &0.056 &0.057 &0.952 \\
&WaveNet~\cite{WaveNet} &SOD &84.84 &0.902 &0.046 &0.046 &0.880&&86.48 &0.927 &0.052 &0.052 &0.960 \\
&SPNet~\cite{SPNet} &SOD &79.52 &0.882 &0.093 &0.064 &0.844&&83.23 &0.908 &0.065 &0.062 &0.942\\
\midrule 
\multirow{2}{*}{RGB-D}
&PDNet~\cite{mei2021depth} &MD &81.12 &0.883 &0.076 &0.063 &0.881&&84.62 &0.921 &0.058 &0.058 &0.947 \\
&CIRNet~\cite{cong2022cir} &SOD &83.26 &0.901 &0.070 &0.075 &0.905&&83.07 &0.913 &0.098 &0.066 &0.931 \\
\midrule
\multirow{1}{*}{RGB-NIR} 
&NRGlassNet~\cite{yan2024nrglassnet} &GSD &85.06 &0.898 &0.043 &0.048 &0.889&&\Best{89.81} &\Best{0.941} &\Best{0.038} &\Best{0.040} &\Best{0.974} \\
\midrule
\multirow{1}{*}{RGB-Flash} 
&Ours &GSD &\SecondBest{85.30} &\SecondBest{0.903} &\SecondBest{0.040} &\SecondBest{0.042} &\SecondBest{0.891} &&\SecondBest{88.39} &\SecondBest{0.934} &\SecondBest{0.043} &\SecondBest{0.044} &\SecondBest{0.966}\\
\bottomrule 
\end{tabular}
\vspace{-3mm}
\end{table*}

\subsubsection{Qualitative Comparison}
As shown in Fig.~\ref{fig:glass_comparision_result}, we compare our method with $7$ SOTA methods~\cite{mei2020don,lin2021rich,xie2024csfwinformer,lin2022semantic,yan2024nrglassnet,mei2021depth,huo2023glass}. The $1$st scene shows that the bottom-left glass surface without a complete window frame \tao{in the no-flash image would} be under-detected by GDNet~\cite{mei2020don}, GSDNet~\cite{lin2021rich} and RGB-D~\cite{mei2021depth}. The $2$nd, $3$rd, $4$th and $5$th scenes show that most methods mis-detect the rectangular frame (\eg, the rightmost open door shop in the $2$nd scene, the bottom-right rectangular region in the $3$rd scene, and the wall and floor near the glass frame in the $5$th scene) as a glass region, as they \tao{have shapes similar to that of} glass surfaces. 

The $6$th and $7$th scenes show challenging examples where glass surface \tao{frames} are difficult to identify. GDNet~\cite{mei2020don}, CSFWinformer~\cite{xie2024csfwinformer}, RGB-S~\cite{lin2022semantic}, and RGB-D~\cite{mei2021depth} under-detect glass surfaces, because the boundaries of the handrail and door are confusing and misleading. 
\tao{Since} the boundary of the glass surface in \tao{the} $8$th scene is transparent, GDNet~\cite{mei2020don}, RGB-S~\cite{lin2022semantic} and RGB-D~\cite{mei2021depth} under-detect \tao{the} glass surface in this scene. 

In the $9$th scene, the \tao{complex and unique appearance of the} rightmost \tao{glass surface compared to other glass surfaces confuses} GDNet~\cite{mei2020don}, RGB-S~\cite{lin2022semantic} and RGB-D~\cite{mei2021depth}, \tao{which may be due to these methods cannot} exploit reflection cues. 
The last scene shows a complex background behind the glass surfaces, \hh{with} \tao{a billboard in front of the glass surface on the left side of the image. Moreover, the} billboard with \tao{appearance similar to} the background \tao{fool}\hh{s} most \tao{competing} methods.

\lywre{These examples demonstrate that our method accurately detects glass surfaces by exploiting the appearance and disappearance of reflections in no-flash and flash image pairs.}

\renewcommand{\newsubwidth}{0.087}
\begin{figure*}
	\renewcommand{\tabcolsep}{0.8pt}
	\renewcommand\arraystretch{0.6}
        \begin{center}
            \begin{tabular}{ccccccccccc}
                \includegraphics[width=\newsubwidth\linewidth]{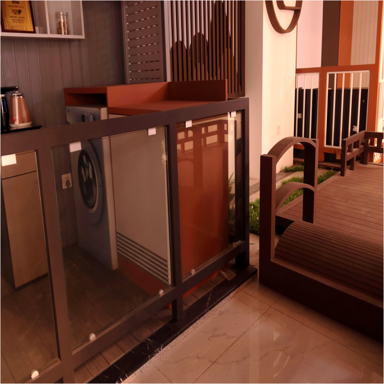}&
                \includegraphics[width=\newsubwidth\linewidth]{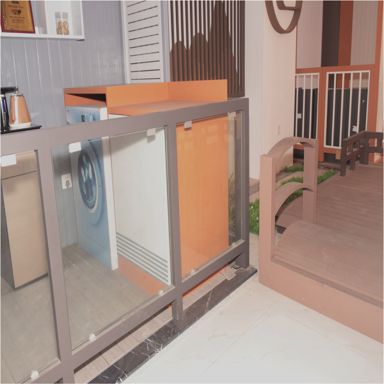}&
                \includegraphics[width=\newsubwidth\linewidth]{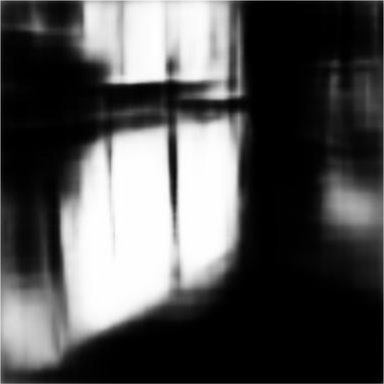}&
                \includegraphics[width=\newsubwidth\linewidth]{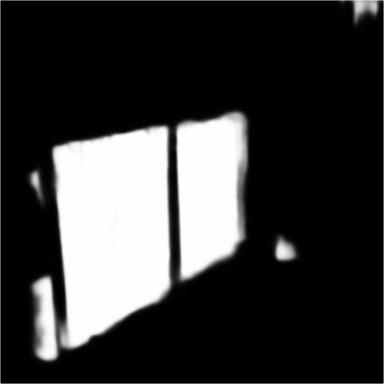}&
                \includegraphics[width=\newsubwidth\linewidth]{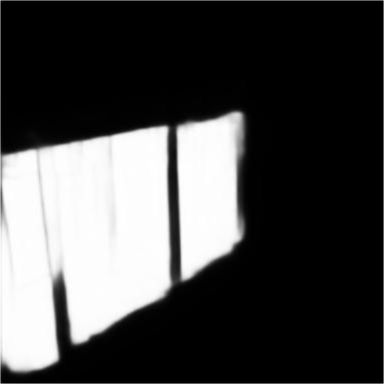}&
                \includegraphics[width=\newsubwidth\linewidth]{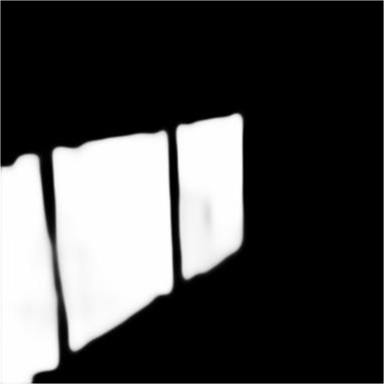}&
                \includegraphics[width=\newsubwidth\linewidth]{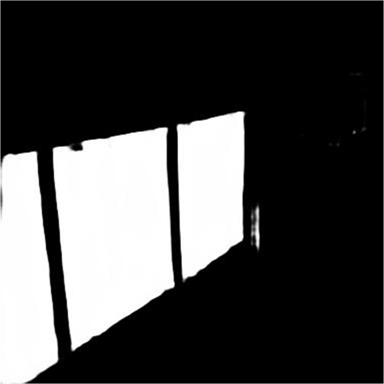}&
                \includegraphics[width=\newsubwidth\linewidth]{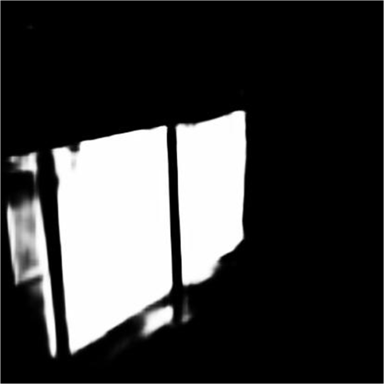}&
                \includegraphics[width=\newsubwidth\linewidth]{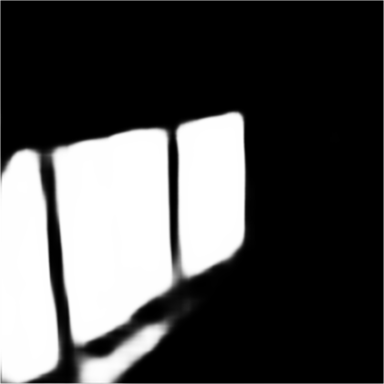}&
                \includegraphics[width=\newsubwidth\linewidth]{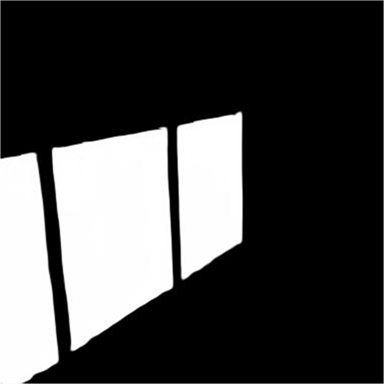}&
                \includegraphics[width=\newsubwidth\linewidth]{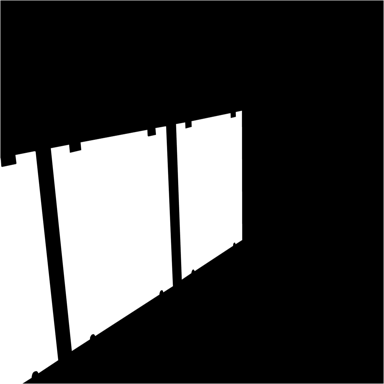} \\ 

                \includegraphics[width=\newsubwidth\linewidth]{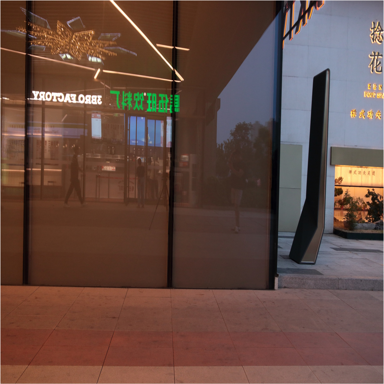}&
                \includegraphics[width=\newsubwidth\linewidth]{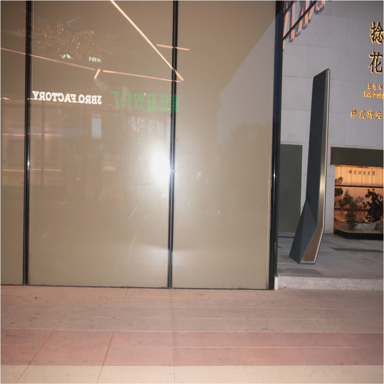}&
                \includegraphics[width=\newsubwidth\linewidth]{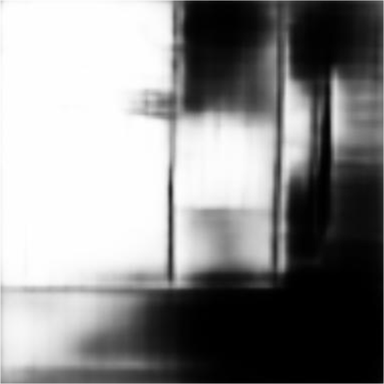}&
                \includegraphics[width=\newsubwidth\linewidth]{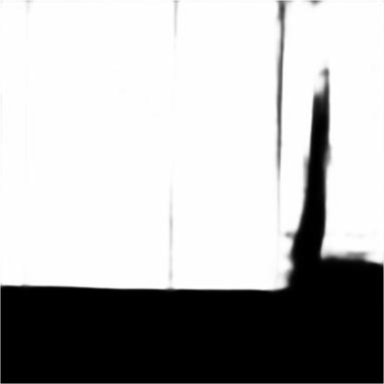}&
                \includegraphics[width=\newsubwidth\linewidth]{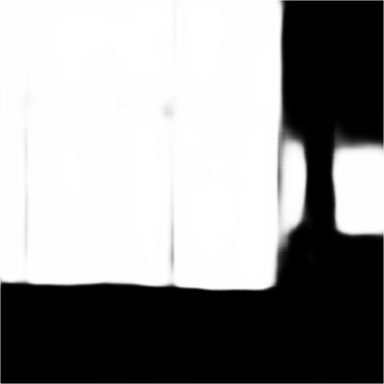}&
                \includegraphics[width=\newsubwidth\linewidth]{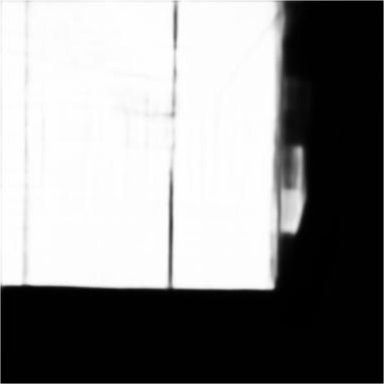}&
                \includegraphics[width=\newsubwidth\linewidth]{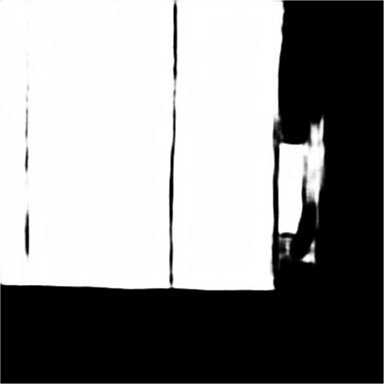}&
                \includegraphics[width=\newsubwidth\linewidth]{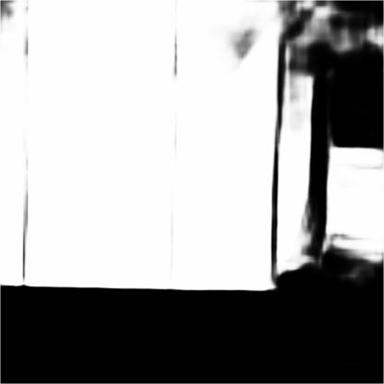}&
                \includegraphics[width=\newsubwidth\linewidth]{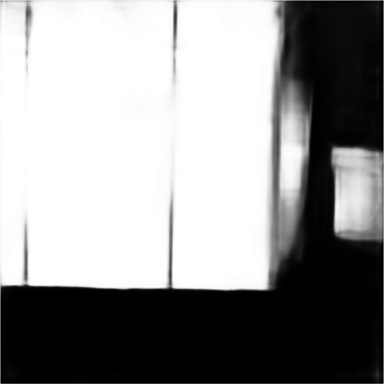}&
                \includegraphics[width=\newsubwidth\linewidth]{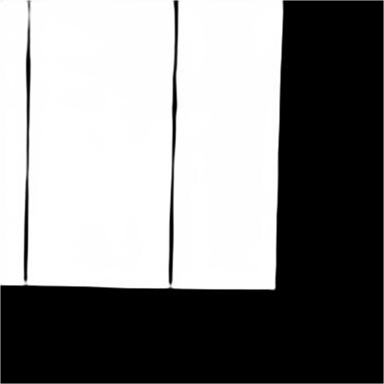}&
                \includegraphics[width=\newsubwidth\linewidth]{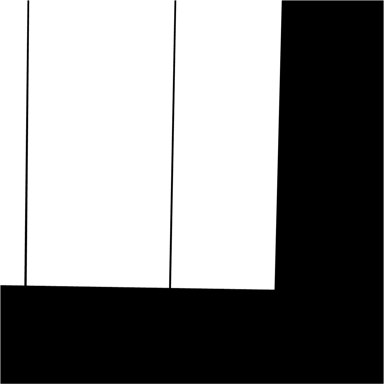} \\ 
                
                \includegraphics[width=\newsubwidth\linewidth]{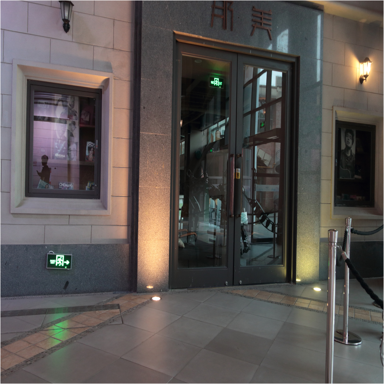}&
                \includegraphics[width=\newsubwidth\linewidth]{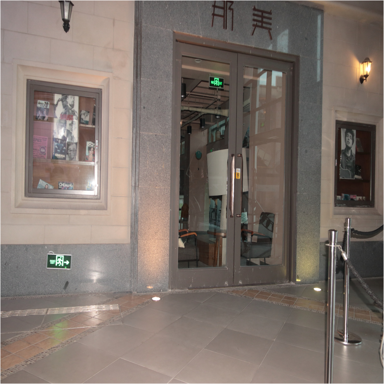}&
                \includegraphics[width=\newsubwidth\linewidth]{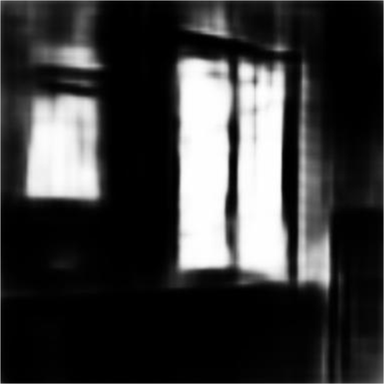}&
                \includegraphics[width=\newsubwidth\linewidth]{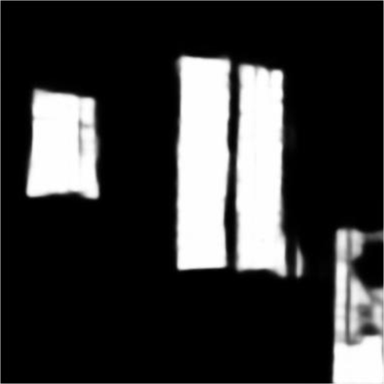}&
                \includegraphics[width=\newsubwidth\linewidth]{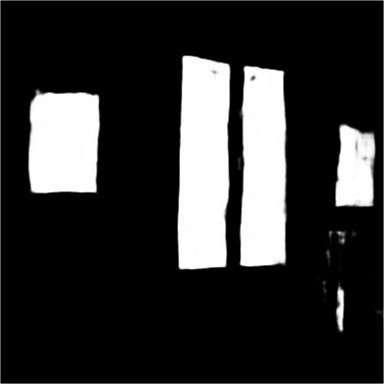}&
                \includegraphics[width=\newsubwidth\linewidth]{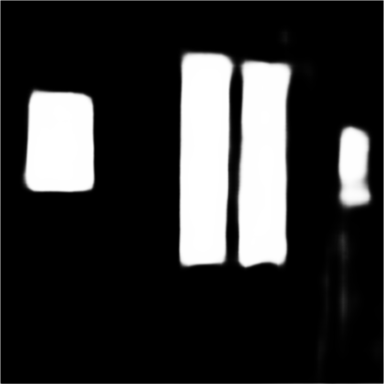}&
                \includegraphics[width=\newsubwidth\linewidth]{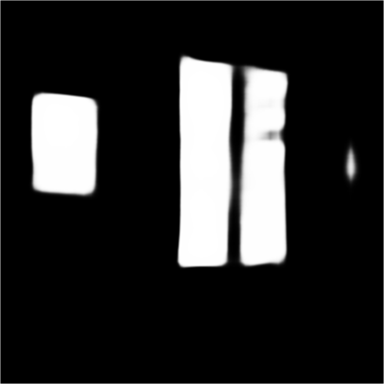}&
                \includegraphics[width=\newsubwidth\linewidth]{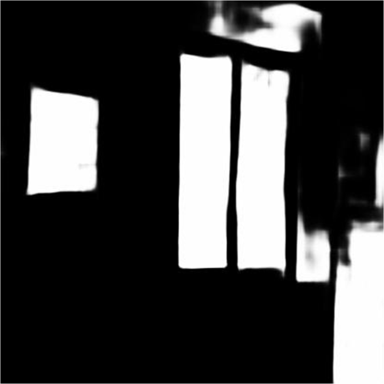}&
                \includegraphics[width=\newsubwidth\linewidth]{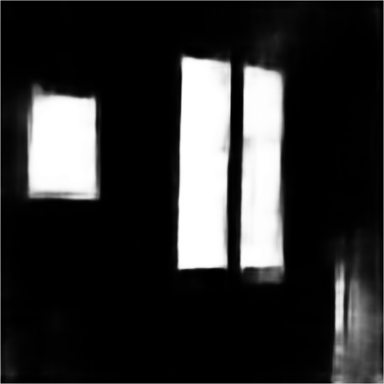}&
                \includegraphics[width=\newsubwidth\linewidth]{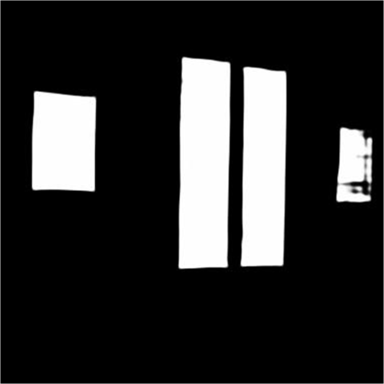}&
                \includegraphics[width=\newsubwidth\linewidth]{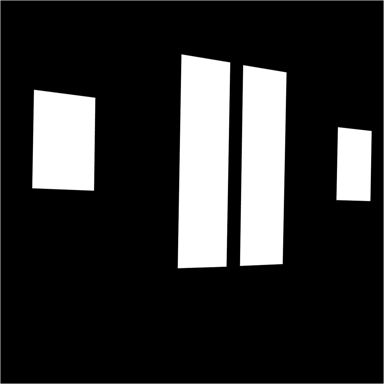} \\ 
                
                \includegraphics[width=\newsubwidth\linewidth]{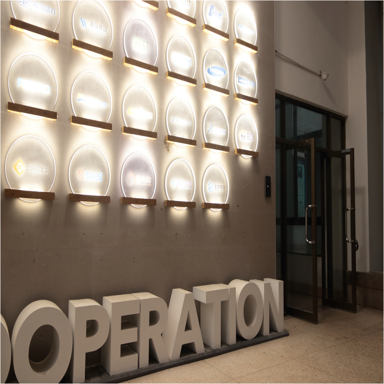}&
                \includegraphics[width=\newsubwidth\linewidth]{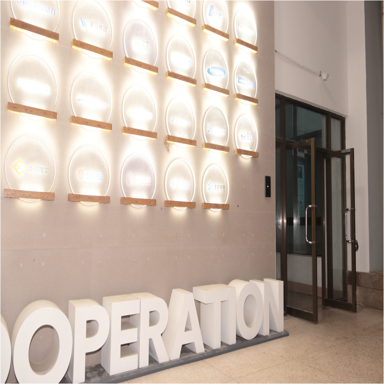}&
                \includegraphics[width=\newsubwidth\linewidth]{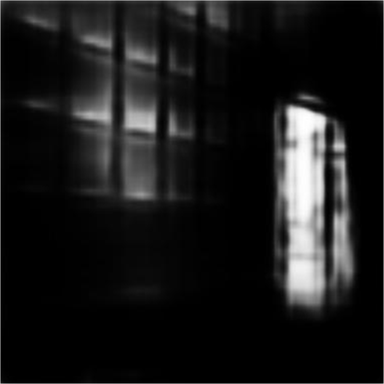}&
                \includegraphics[width=\newsubwidth\linewidth]{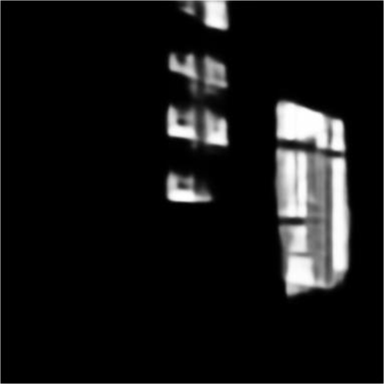}&
                \includegraphics[width=\newsubwidth\linewidth]{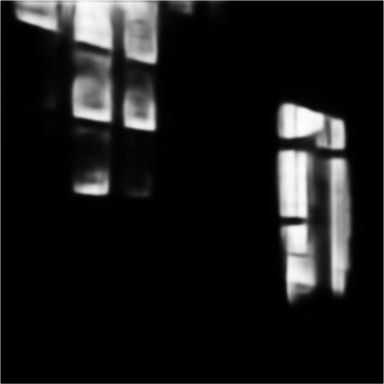}&
                \includegraphics[width=\newsubwidth\linewidth]{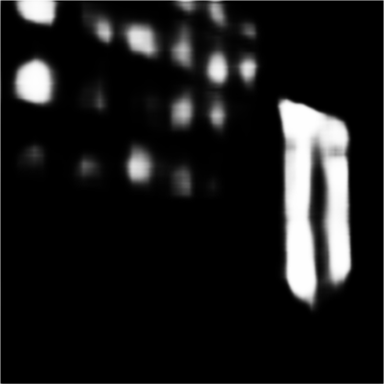}&
                \includegraphics[width=\newsubwidth\linewidth]{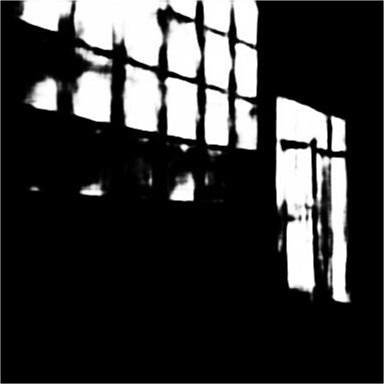}&
                \includegraphics[width=\newsubwidth\linewidth]{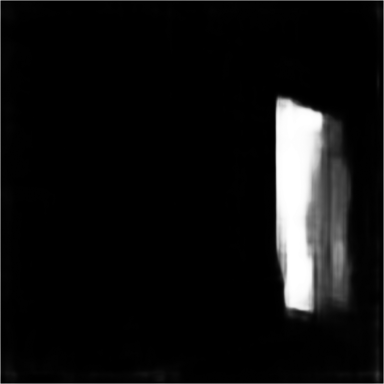}&
                \includegraphics[width=\newsubwidth\linewidth]{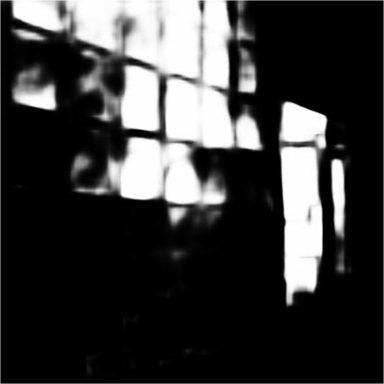}&
                \includegraphics[width=\newsubwidth\linewidth]{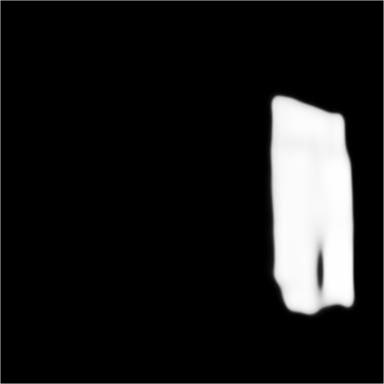}&
                \includegraphics[width=\newsubwidth\linewidth]{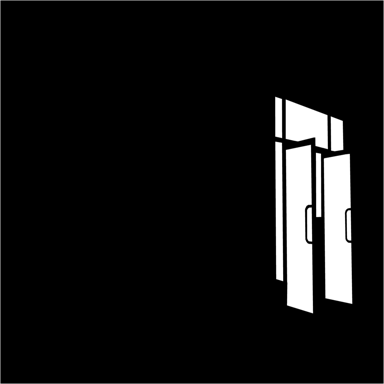} \\ 

                \includegraphics[width=\newsubwidth\linewidth]{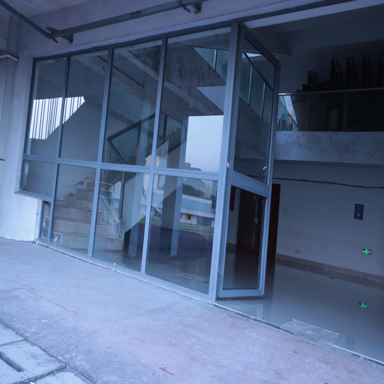}&
                \includegraphics[width=\newsubwidth\linewidth]{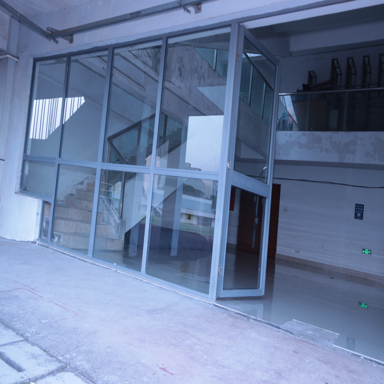}&
                \includegraphics[width=\newsubwidth\linewidth]{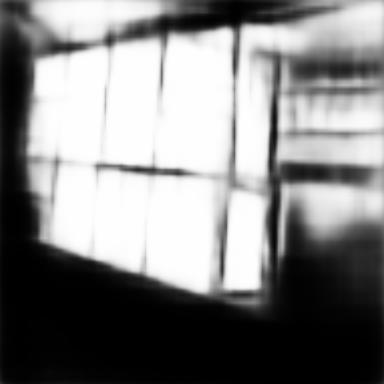}&
                \includegraphics[width=\newsubwidth\linewidth]{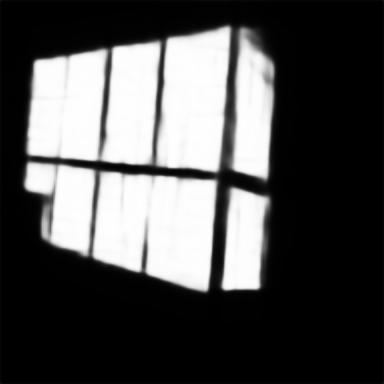}&
                \includegraphics[width=\newsubwidth\linewidth]{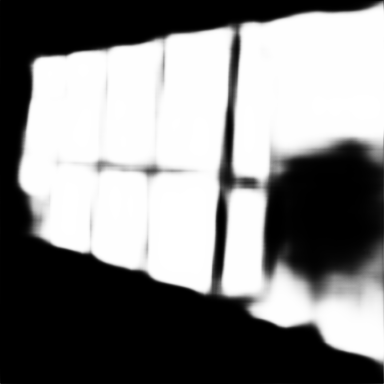}&
                \includegraphics[width=\newsubwidth\linewidth]{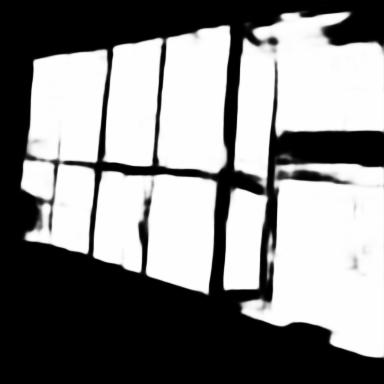}&
                \includegraphics[width=\newsubwidth\linewidth]{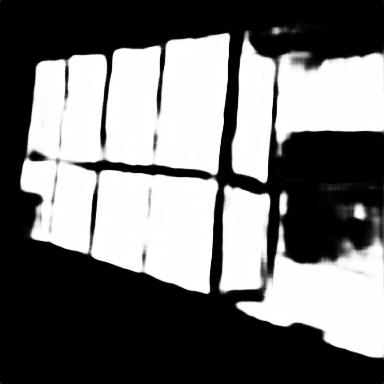}&
                \includegraphics[width=\newsubwidth\linewidth]{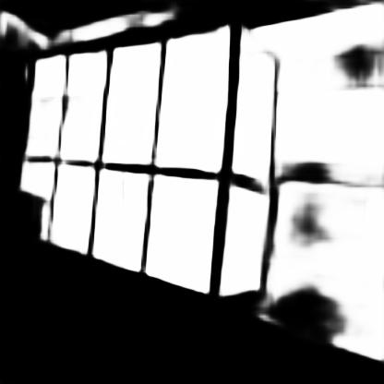}&
                \includegraphics[width=\newsubwidth\linewidth]{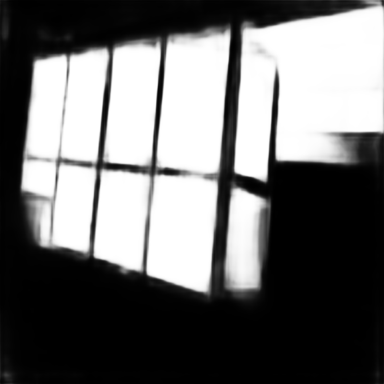}&
                \includegraphics[width=\newsubwidth\linewidth]{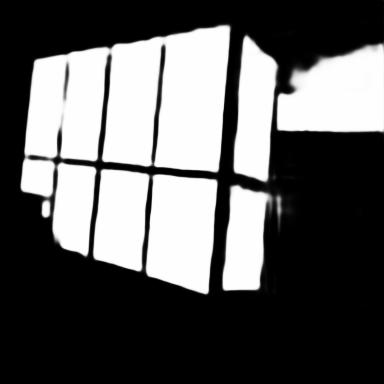}&
                \includegraphics[width=\newsubwidth\linewidth]{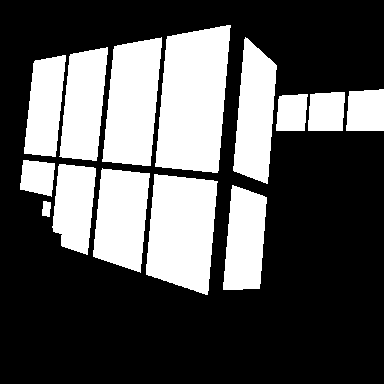} \\ 

                \includegraphics[width=\newsubwidth\linewidth]{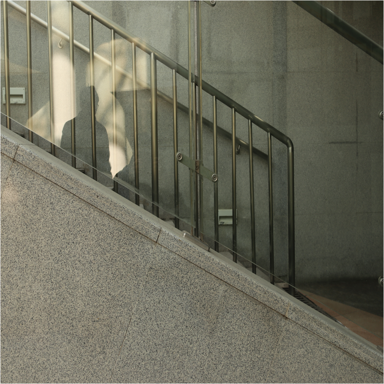}&
                \includegraphics[width=\newsubwidth\linewidth]{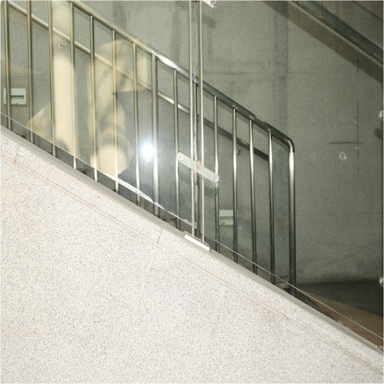}&
                \includegraphics[width=\newsubwidth\linewidth]{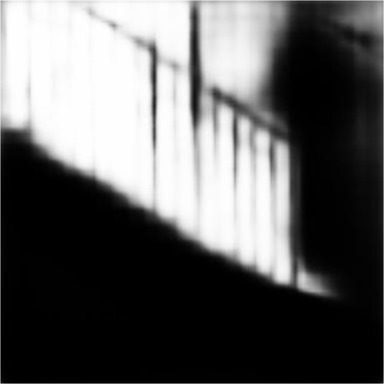}&
                \includegraphics[width=\newsubwidth\linewidth]{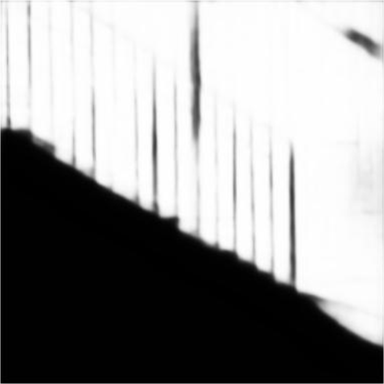}&
                \includegraphics[width=\newsubwidth\linewidth]{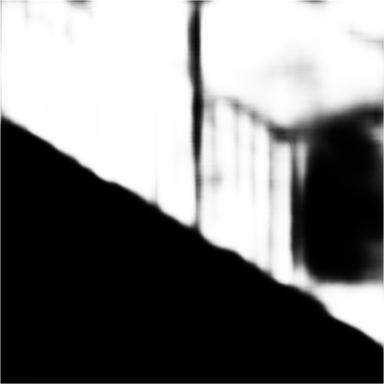}&
                \includegraphics[width=\newsubwidth\linewidth]{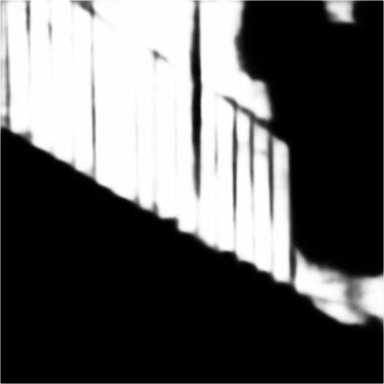}&
                \includegraphics[width=\newsubwidth\linewidth]{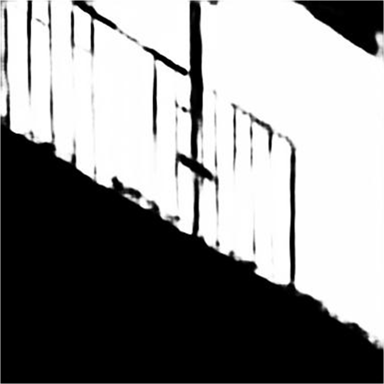}&
                \includegraphics[width=\newsubwidth\linewidth]{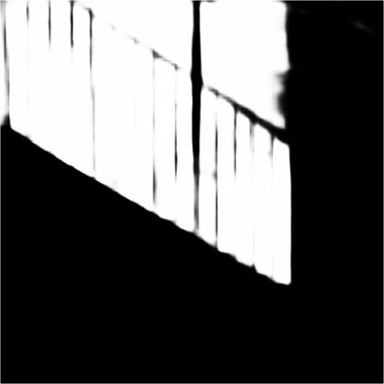}&
                \includegraphics[width=\newsubwidth\linewidth]{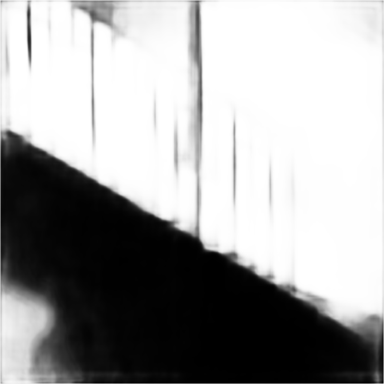}&
                \includegraphics[width=\newsubwidth\linewidth]{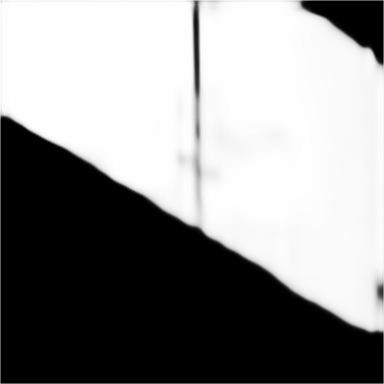}&
                \includegraphics[width=\newsubwidth\linewidth]{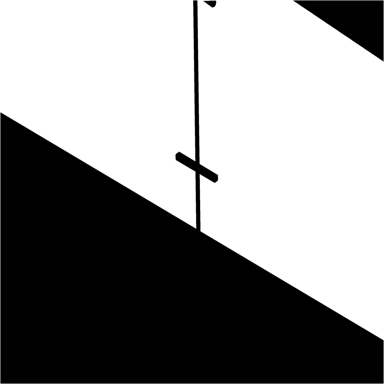} \\ 

                \includegraphics[width=\newsubwidth\linewidth]{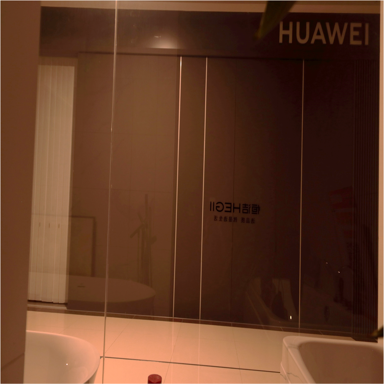}&
                \includegraphics[width=\newsubwidth\linewidth]{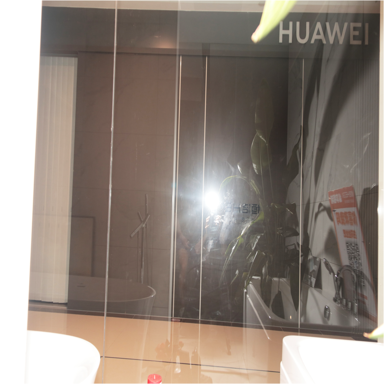}&
                \includegraphics[width=\newsubwidth\linewidth]{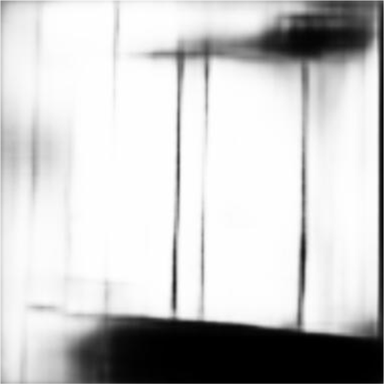}&
                \includegraphics[width=\newsubwidth\linewidth]{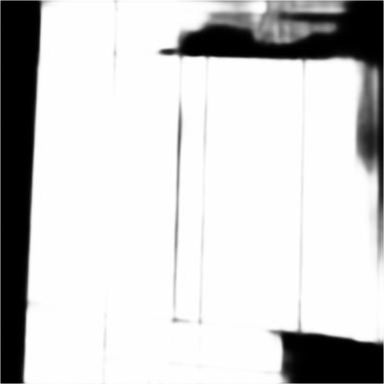}&
                \includegraphics[width=\newsubwidth\linewidth]{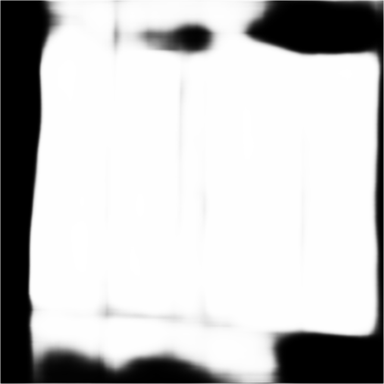}&
                \includegraphics[width=\newsubwidth\linewidth]{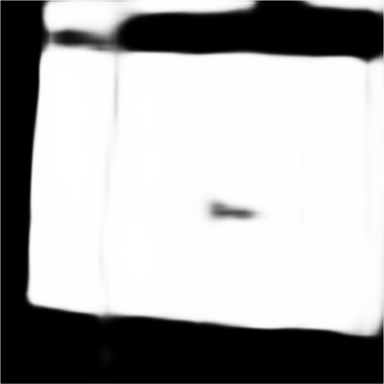}&
                \includegraphics[width=\newsubwidth\linewidth]{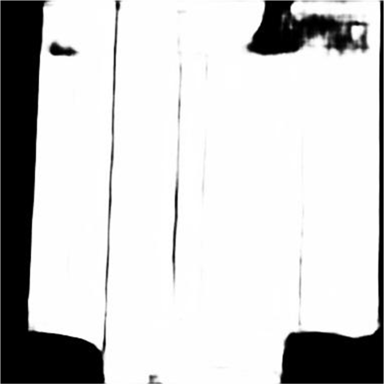}&
                \includegraphics[width=\newsubwidth\linewidth]{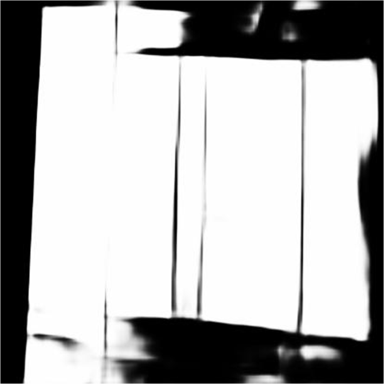}&
                \includegraphics[width=\newsubwidth\linewidth]{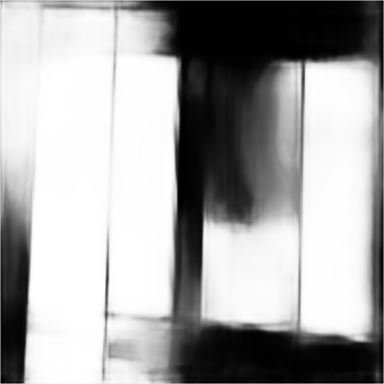}&
                \includegraphics[width=\newsubwidth\linewidth]{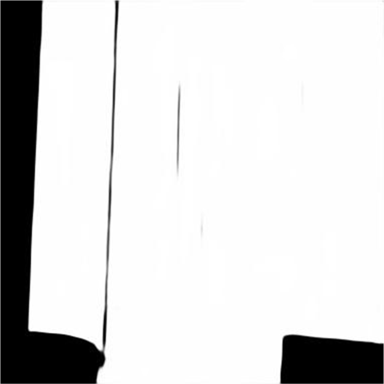}&
                \includegraphics[width=\newsubwidth\linewidth]{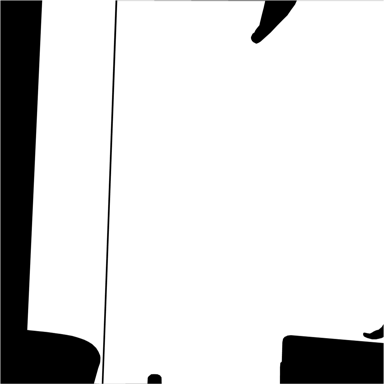} \\

                \includegraphics[width=\newsubwidth\linewidth]{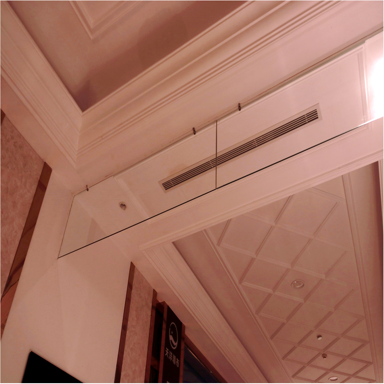}&
                \includegraphics[width=\newsubwidth\linewidth]{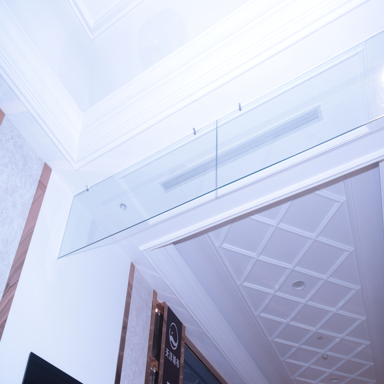}&
                \includegraphics[width=\newsubwidth\linewidth]{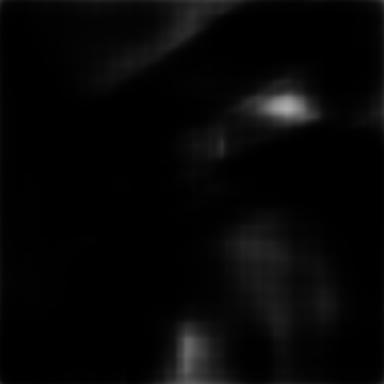}&
                \includegraphics[width=\newsubwidth\linewidth]{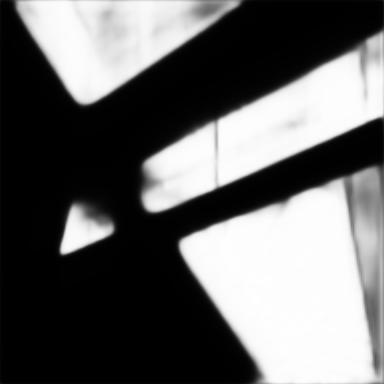}&
                \includegraphics[width=\newsubwidth\linewidth]{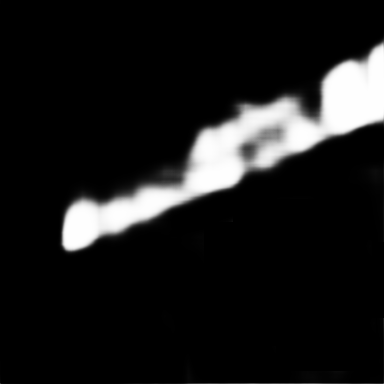}&
                \includegraphics[width=\newsubwidth\linewidth]{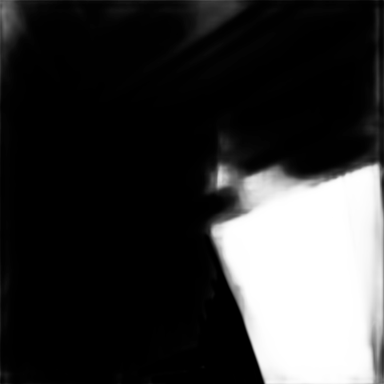}&
                \includegraphics[width=\newsubwidth\linewidth]{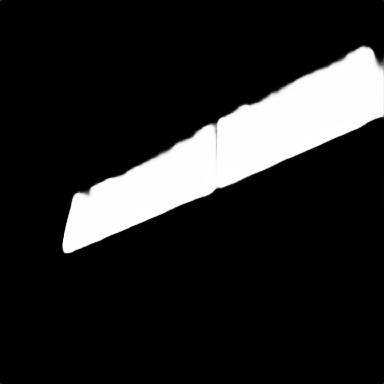}&
                \includegraphics[width=\newsubwidth\linewidth]{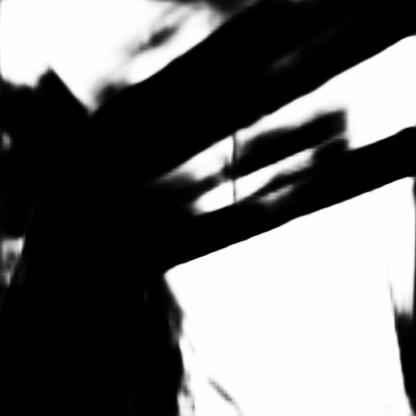}&
                \includegraphics[width=\newsubwidth\linewidth]{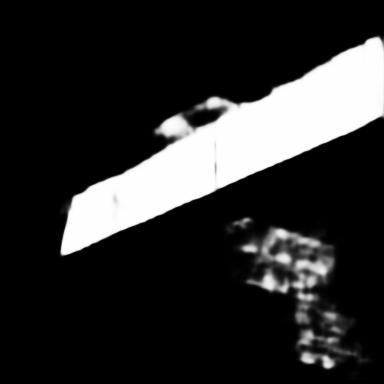}&
                \includegraphics[width=\newsubwidth\linewidth]{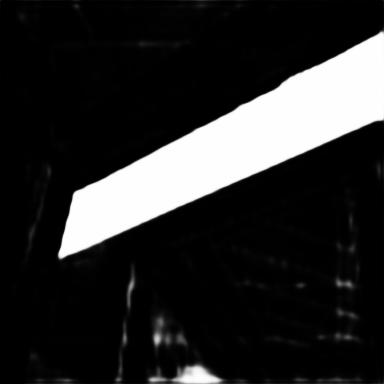}&
                \includegraphics[width=\newsubwidth\linewidth]{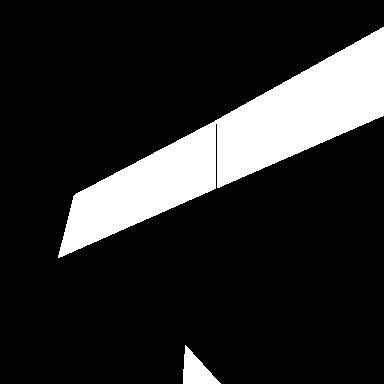} \\ 

                \includegraphics[width=\newsubwidth\linewidth]{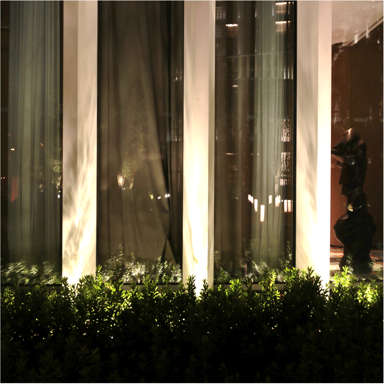}&
                \includegraphics[width=\newsubwidth\linewidth]{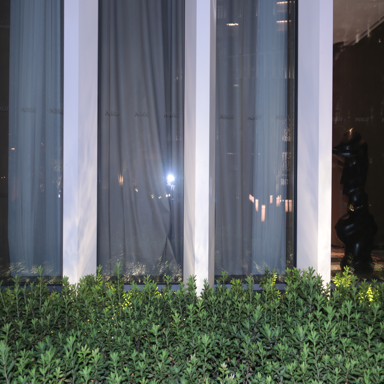}&
                \includegraphics[width=\newsubwidth\linewidth]{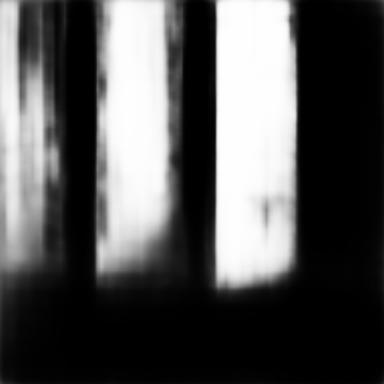}&
                \includegraphics[width=\newsubwidth\linewidth]{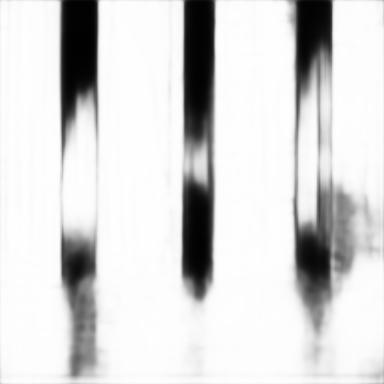}&
                \includegraphics[width=\newsubwidth\linewidth]{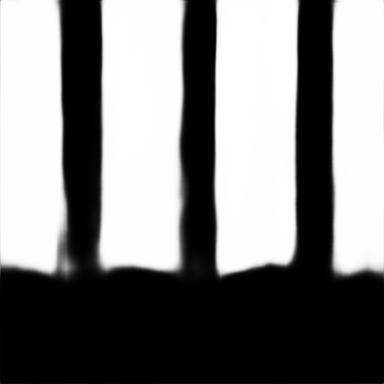}&
                \includegraphics[width=\newsubwidth\linewidth]{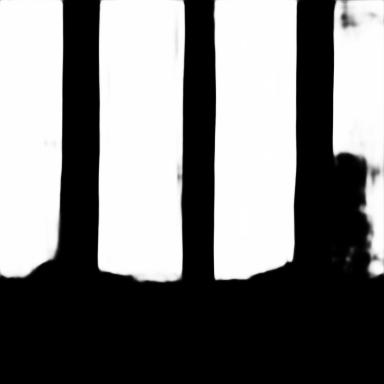}&
                \includegraphics[width=\newsubwidth\linewidth]{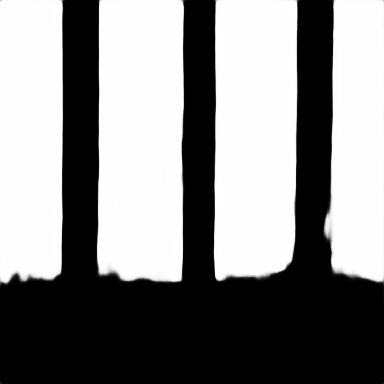}&
                \includegraphics[width=\newsubwidth\linewidth]{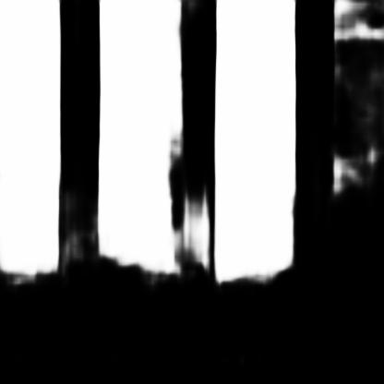}&
                \includegraphics[width=\newsubwidth\linewidth]{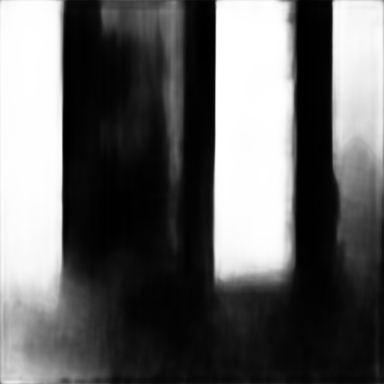}&
                \includegraphics[width=\newsubwidth\linewidth]{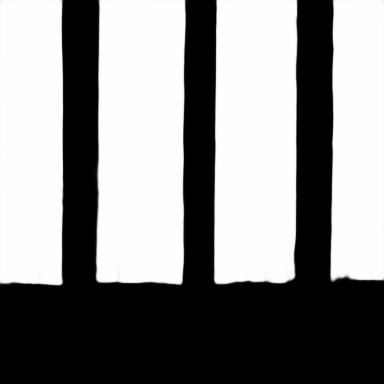}&
                \includegraphics[width=\newsubwidth\linewidth]{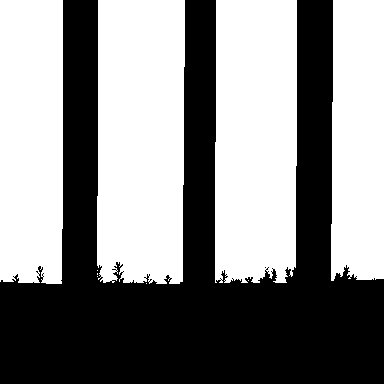} \\ 

                \includegraphics[width=\newsubwidth\linewidth]{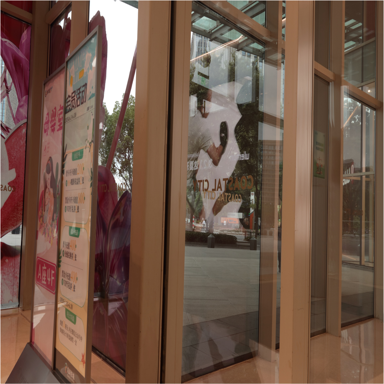}&
                \includegraphics[width=\newsubwidth\linewidth]{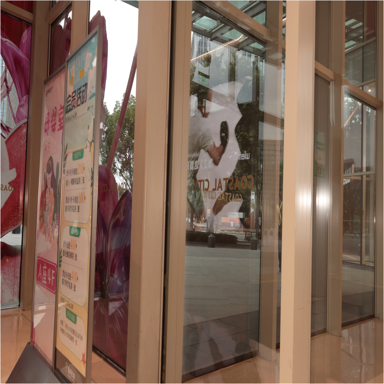}&
                \includegraphics[width=\newsubwidth\linewidth]{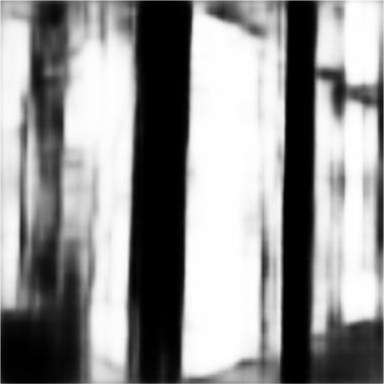}&
                \includegraphics[width=\newsubwidth\linewidth]{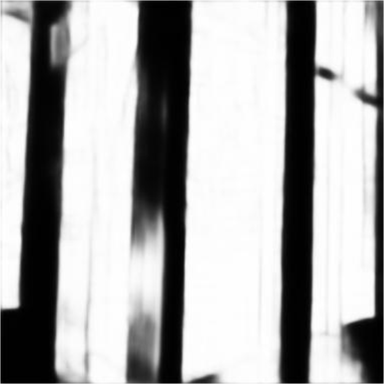}&
                \includegraphics[width=\newsubwidth\linewidth]{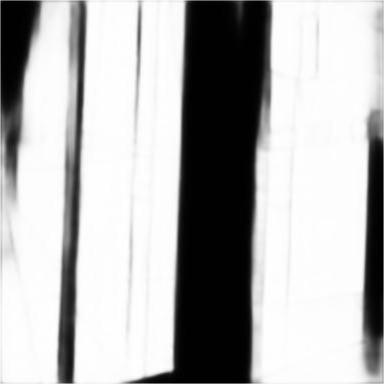}&
                \includegraphics[width=\newsubwidth\linewidth]{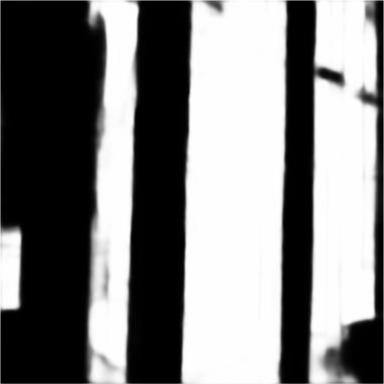}&
                \includegraphics[width=\newsubwidth\linewidth]{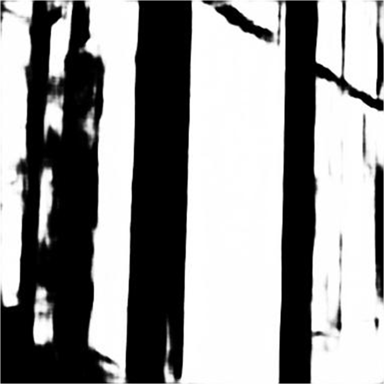}&
                \includegraphics[width=\newsubwidth\linewidth]{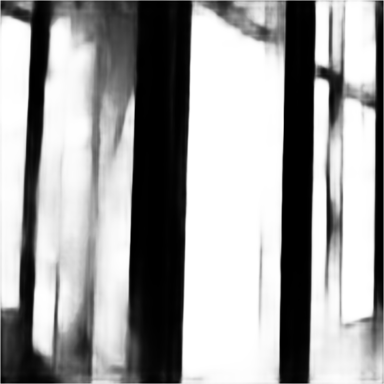}&
                \includegraphics[width=\newsubwidth\linewidth]{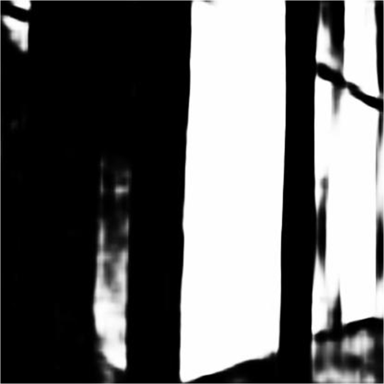}&
                \includegraphics[width=\newsubwidth\linewidth]{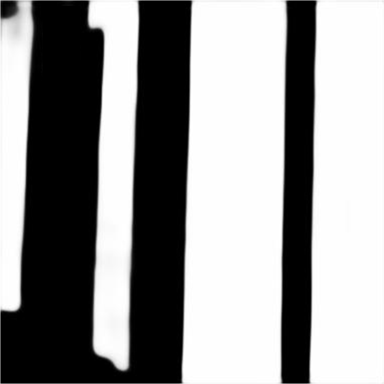}&
                \includegraphics[width=\newsubwidth\linewidth]{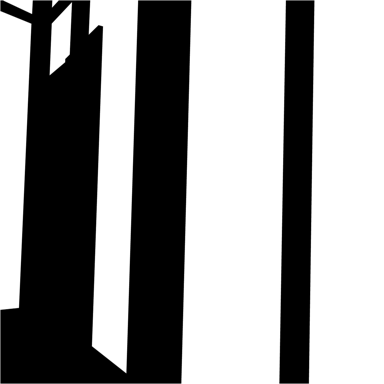} \\

                \fontsize{7.0pt}{\baselineskip}\selectfont{No-flash Image}&
                \fontsize{7.0pt}{\baselineskip}\selectfont{Flash Image}&
                \fontsize{7.0pt}{\baselineskip}\selectfont{GDNet~\cite{mei2020don}}&
                \fontsize{7.0pt}{\baselineskip}\selectfont{GSDNet~\cite{lin2021rich}}&
                \fontsize{7.0pt}{\baselineskip}\selectfont{CSFwin~\cite{xie2024csfwinformer}}&
                \fontsize{7.0pt}{\baselineskip}\selectfont{RGB-S~\cite{lin2022semantic}}&
                \fontsize{7.0pt}{\baselineskip}\selectfont{RGB-NIR~\cite{yan2024nrglassnet}}&
                \fontsize{7.0pt}{\baselineskip}\selectfont{RGB-D~\cite{mei2021depth}}&
                \fontsize{7.0pt}{\baselineskip}\selectfont{RGB-T~\cite{huo2023glass}}&
                \fontsize{7.0pt}{\baselineskip}\selectfont{Ours}&
                \fontsize{7.0pt}{\baselineskip}\selectfont{GT} \\
                
            \end{tabular}
        \end{center}
 	      \vspace{-0.016\textwidth}
        \caption{Visual comparison of the glass surface detection results against state-of-the-art methods on our NFGD dataset. The first three methods based on single (\ie, no-flash) image detection, while the next four methods are multimodal detection \tao{methods} (with the flash image as another modality input).}
        \label{fig:glass_comparision_result}
        \vspace{-3mm}
\end{figure*}

\subsection{Ablation Studies}
We conduct ablation studies on our proposed \textit{NFGlassNet} to validate the effectiveness of the key modules, \tao{as} shown in Tab.~\ref{tab:ablation_study_modules}. \tao{``Base" refers to the variant of our network after removing all the RCMMs and RGAMs, \lywre{and ``Base (Single branch)" uses flash images and non-flash images concatenated along the channel as input.}}
Then, we add either RCMM or RGAM to the ``Base" model, which results in an improvement of IoU \tao{metric} by $2.89\%$ and $0.28\%$, respectively. The significant performance improvement brought by RCMM demonstrates the effectiveness of reflection cues in glass surface detection. Additionally, we replace RCMM with dual UNet, and RGAM with CBAM~\cite{woo2018cbam}, \tao{respectively,} in our model. \tao{Then, the} IoU \tao{value} drops by $2.19\%$ and $1.47\%$, respectively, compared to our \tao{complete} model. \tao{This demonstrates} the rationality and effectiveness of our \tao{network}. 
\hh{Visual examples of incorporating the RCMM and RGAM are shown in Fig.~\ref{fig:examples_AB_study}.}
Further ablation \tao{study on the structure of RCMM and RGAM} are \tao{reported} in the \lywre{Appendix~\ref{sec:abl}.}

\renewcommand{\tabcolsep}{3.8pt}
\renewcommand\arraystretch{1}
\begin{table}[t]
\caption{
Ablation study of our method (RCMM and RGAM) on our NFGD dataset. ``Base" refers to the no-flash and flash backbone networks with decoders, excluding RCMM and RGAM. ``Base (Single branch)" refers to the use of flash images and non-flash images concatenated along the channel as input. In ``Base+RGAM," we utilized the inferred reflection results from LANet, then encoded them to match the dimensions and channel numbers of RCMM's output to replace RGAM's original input.} 
  \label{tab:ablation_study_modules}
  \vspace{-2mm}
  \centering
    \begin{tabular}{lccccc}
    \toprule
    Method  &IoU$\uparrow$ &F$_\beta$$\uparrow$ &MAE$\downarrow$ &BER$\downarrow$ &ACC$\uparrow$ \\
    \midrule
    Base &81.96 &0.897 &0.119 &0.118 &0.896 \\
    Base (Single branch) &77.22 &0.836 &0.161 &0.198 &0.846 \\
    Base + RCMM &84.85 &0.914 &0.089 &0.089 &0.916 \\
    Base + RGAM &82.24 &0.902 &0.114 &0.099 &0.899 \\
    \midrule
    Base + dual UNet + RGAM &84.27 &0.905 &0.089 &0.097 &0.913 \\
    Base + RCMM + CBAM~\cite{woo2018cbam} &84.99 &0.914 &0.087 &0.094 &0.921 \\
    \midrule
    Base + RCMM + RGAM (Ours) &\Best{86.46} &\Best{0.926} &\Best{0.080} &\Best{0.076} &\Best{0.922} \\
    \bottomrule
  \end{tabular}
  \vspace{-2mm}
\end{table}

\renewcommand{\newsubwidth}{0.105}
\begin{figure*}[ht]
	\renewcommand{\tabcolsep}{0.8pt}
	\renewcommand\arraystretch{0.6}
        \begin{center}
            \begin{tabular}{ccccccccc}
                
                \includegraphics[width=\newsubwidth\linewidth]{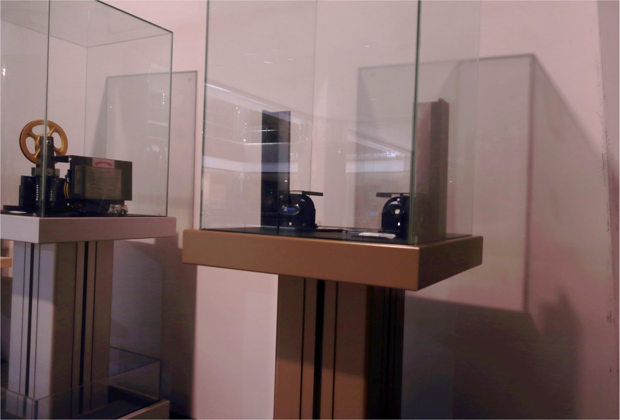}&
                \includegraphics[width=\newsubwidth\linewidth]{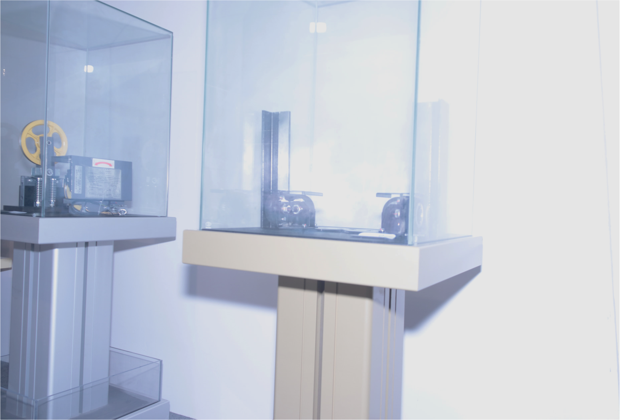}&
                \includegraphics[width=\newsubwidth\linewidth]{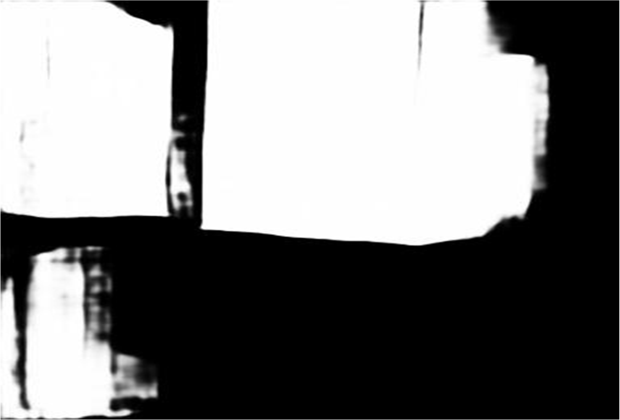}&
                \includegraphics[width=\newsubwidth\linewidth]{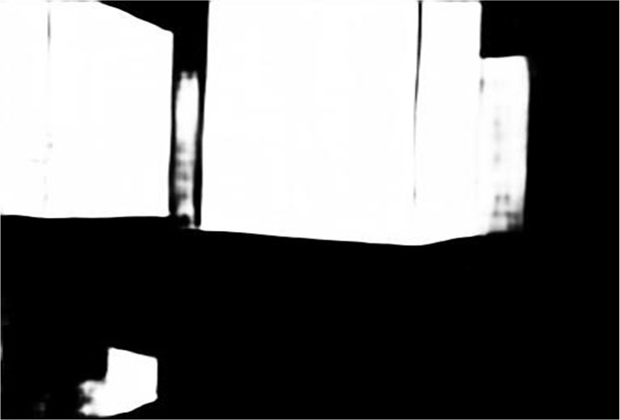}&
                \includegraphics[width=\newsubwidth\linewidth]{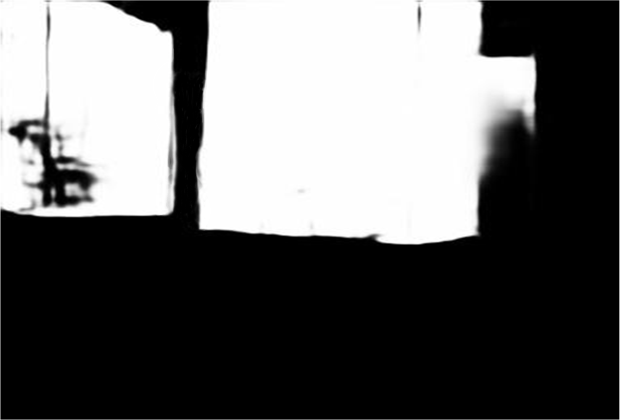}&
                \includegraphics[width=\newsubwidth\linewidth]{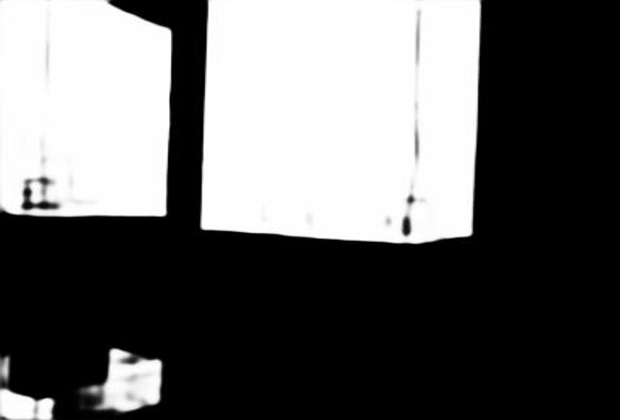}&
                \includegraphics[width=\newsubwidth\linewidth]{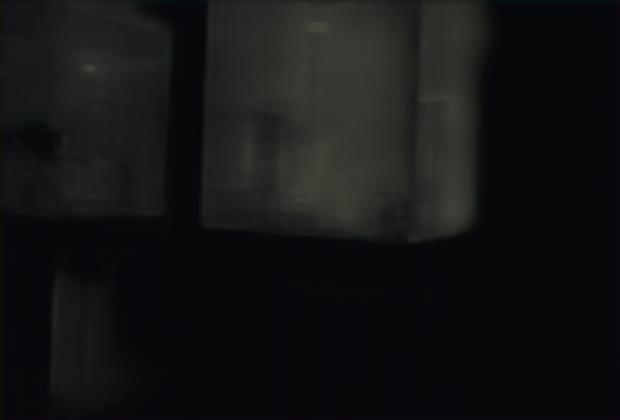}&
                \includegraphics[width=\newsubwidth\linewidth]{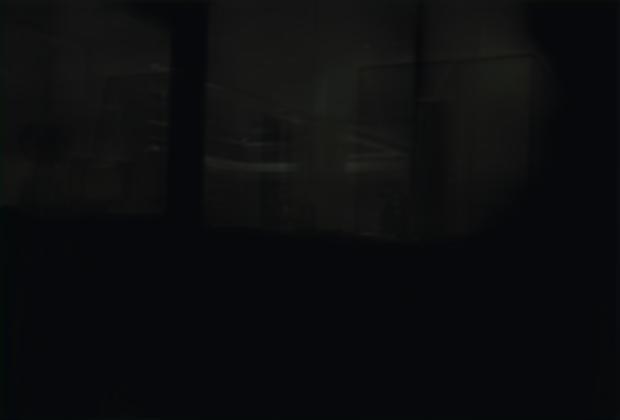}&
                \includegraphics[width=\newsubwidth\linewidth]{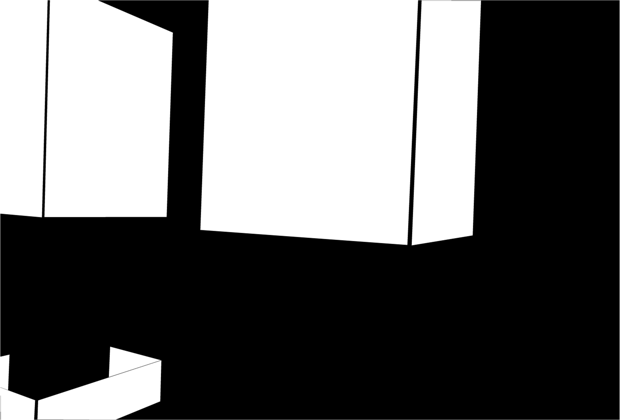}\\ 
                

                \includegraphics[width=\newsubwidth\linewidth]{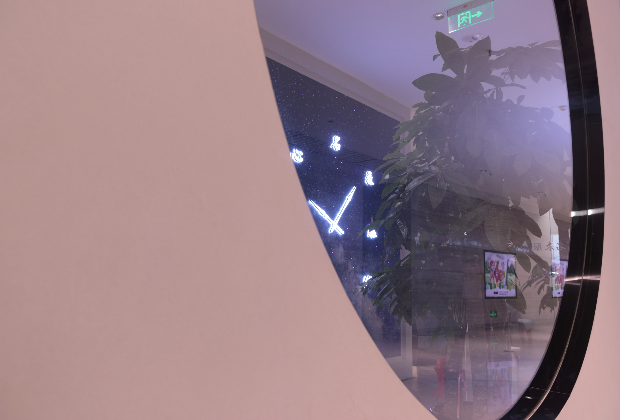}&
                \includegraphics[width=\newsubwidth\linewidth]{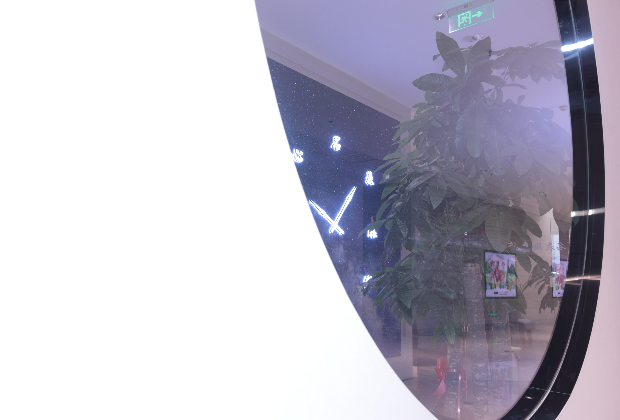}&
                \includegraphics[width=\newsubwidth\linewidth]{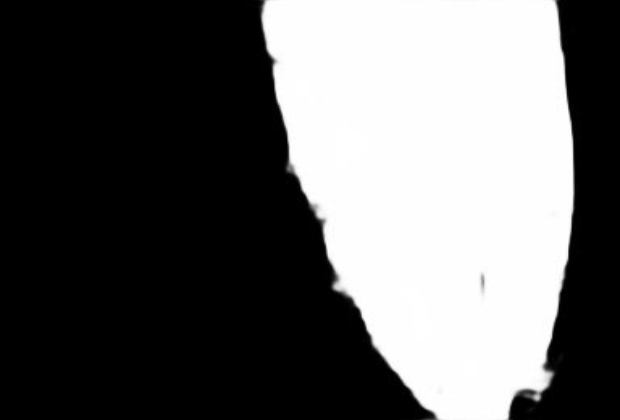}&
                \includegraphics[width=\newsubwidth\linewidth]{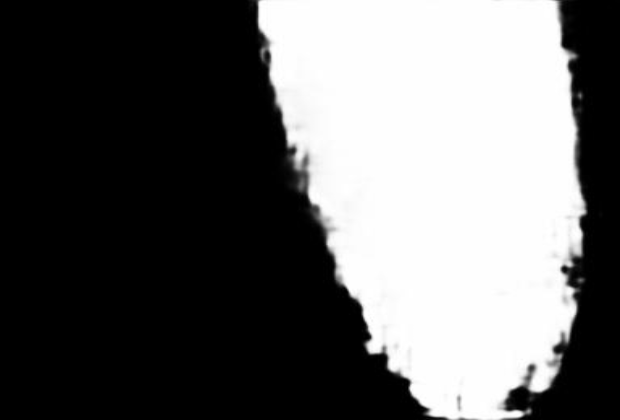}&
                \includegraphics[width=\newsubwidth\linewidth]{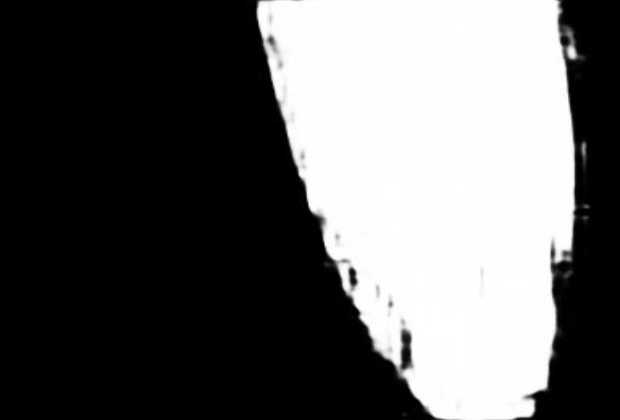}&
                \includegraphics[width=\newsubwidth\linewidth]{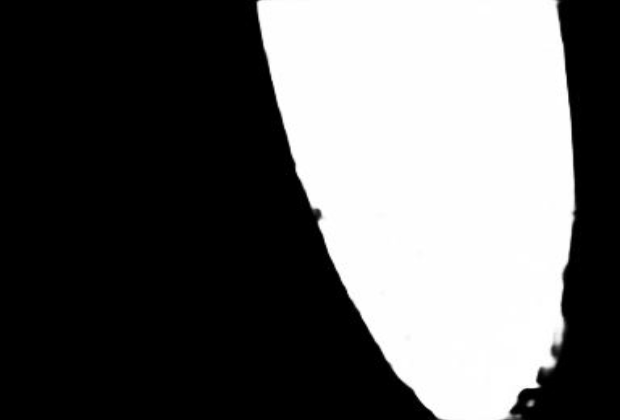}&
                \includegraphics[width=\newsubwidth\linewidth]{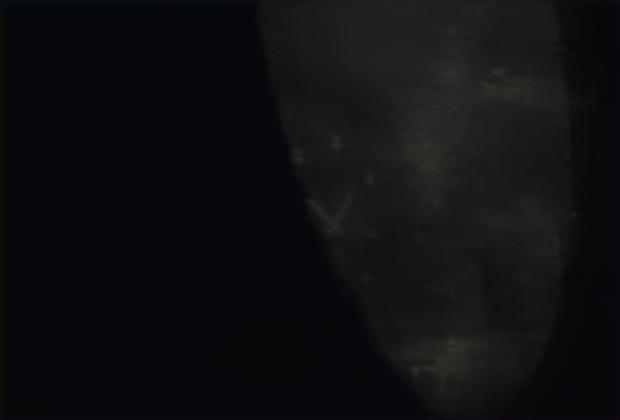}&
                \includegraphics[width=\newsubwidth\linewidth]{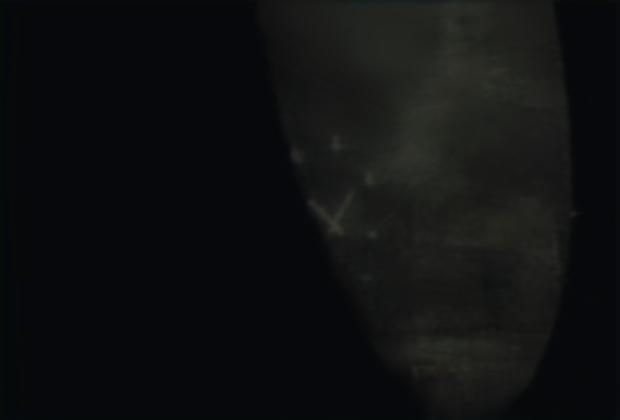}&
                \includegraphics[width=\newsubwidth\linewidth]{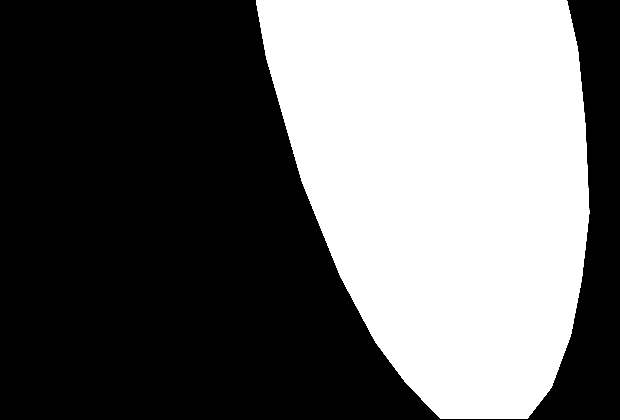}\\
                
                \scriptsize{No-flash Image}&
                \scriptsize{Flash Image}&
                \scriptsize{Base}&
                \scriptsize{Base + RCMM}&
                \scriptsize{Base + RGAM}&
                \scriptsize{Ours}&
                \scriptsize{$\mathbf{R}_\text{no-flash}$}&
                \scriptsize{$\mathbf{R}_\text{flash}$}&
                \scriptsize{GT}\\
                
            \end{tabular}
        \end{center}
        \vspace{-0.016\textwidth}
        \caption{Visual examples of the ablation study. The ``Base" model \tao{means the variation of our \textit{NFGlassNet} without RCMMs and RGAMs, which} is prone to over-detection in most scenes. \tao{While} incorporating RCMMs, the model \tao{can utilize reflection} cues to locate glass regions, thereby reducing over-detection. The \tao{variation of our network (``Base + RGAM") can reduce mis-detecting} glass-like rectangular areas \tao{glass surfaces}. \tao{Our complete network} (Base + RCMM + RGAM) performs best in \tao{all} scenes.}
        \label{fig:examples_AB_study}
        \vspace{-4mm}
\end{figure*}

\subsection{Effectiveness of the Loss Terms}
As shown in Tab.~\ref{tab:ablation_study_loss}, we conduct \tao{an ablation study} on the loss \tao{terms of our method}. We observe that \tao{merely supervising the estimated reflections}, but not supervising \tao{the predicted} glass surface mask, significantly reduces \tao{the performance of our model}. \tao{On the other hand, just going to supervise the predicted glass surface mask, but not supervising the estimated reflections}, also weakens \tao{the performance of our method,} which indirectly proves the importance of reflection cues \tao{for} our method.

\renewcommand{\tabcolsep}{3.5pt}
\begin{table}[ht]
\caption{Ablation study of loss functions on the proposed \textit{NFGlassNet}.} 
  \label{tab:ablation_study_loss}
  \vspace{-2mm}
  \centering
   \textcolor{black}{
    \begin{tabular}{ccc|ccccc}
    \toprule
    $\mathcal{L}_{glass}$ &$\mathcal{L}_{refle}^{\text{flash}}$ &$\mathcal{L}_{refle}^{\text{no-flash}}$ &IoU$\uparrow$ &F$_\beta$$\uparrow$ &MAE$\downarrow$ &BER$\downarrow$ &ACC$\uparrow$ \\
    \midrule
      \checkmark & & &82.21 &0.897 &0.116 &0.101 &0.900 \\
      &\checkmark & &73.47 &0.849 &0.127 &0.148 &0.825 \\
      & &\checkmark &73.60 &0.853 &0.127 &0.143 &0.831 \\
      \checkmark &\checkmark &  &86.11 &0.922 &0.081 &0.078 &0.933 \\
      \checkmark & &\checkmark &86.15 &0.921 &\Best{0.080} &0.077 &\Best{0.935} \\
      \midrule
      \checkmark &\checkmark &\checkmark &\Best{86.46} &\Best{0.926} &\Best{0.080} &\Best{0.076} &0.922 \\
    \bottomrule
  \end{tabular}}
  \vspace{-3mm}
\end{table}

\subsection{Failure Cases}
\lywre{Despite the strong overall performance,} our method does have limitations in some challenging scenes. As shown in \tao{the $1st$ scene of} Fig.~\ref{fig:failure_cases}, our method over-detects \tao{the smooth floor tile} as the glass \tao{surface}, as these glass-like \tao{floor tiles} share similar reflections and shapes \tao{as} glass surfaces. 
\tao{In the $2nd$ scene, the glass-like wall tile on the right side of the image \tao{having prominent reflection} is mis-detected as the glass surface by our method.} 
\lywre{Future work could investigate incorporating complementary geometric or material-aware cues to further improve robustness.}

\renewcommand{\newsubwidth}{0.16}
\begin{figure}[ht]
	\renewcommand{\tabcolsep}{0.8pt}
	\renewcommand\arraystretch{0.6}
        \begin{center}
            \begin{tabular}{cccccc}
                \includegraphics[width=\newsubwidth\linewidth]{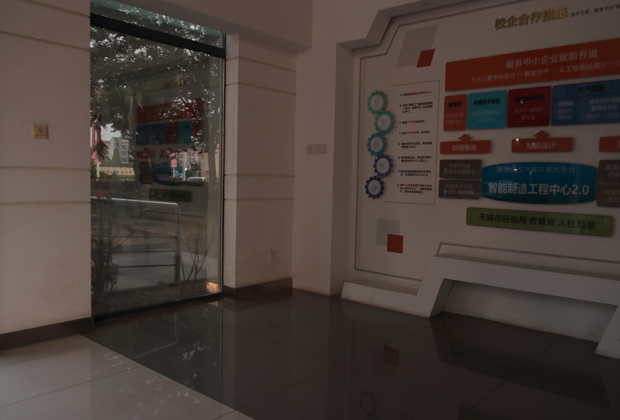}&
                \includegraphics[width=\newsubwidth\linewidth]{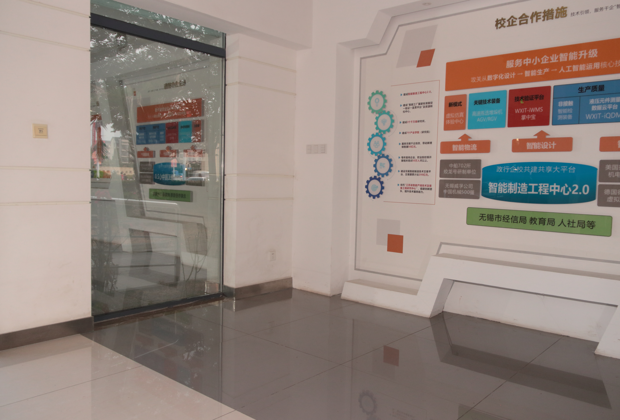}&
                \includegraphics[width=\newsubwidth\linewidth]{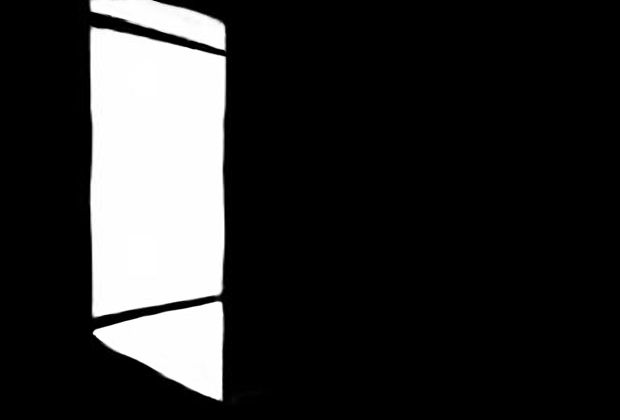}&
                \includegraphics[width=\newsubwidth\linewidth]{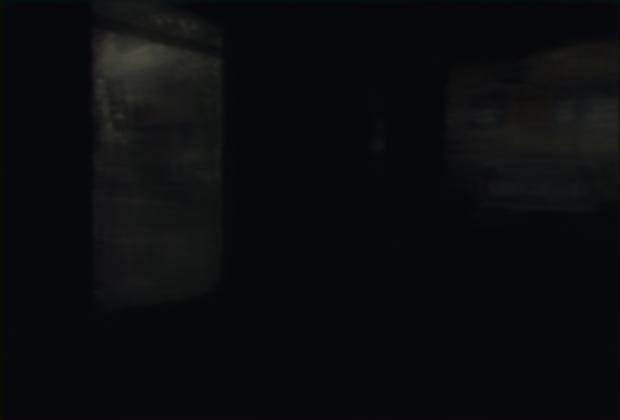}&
                \includegraphics[width=\newsubwidth\linewidth]{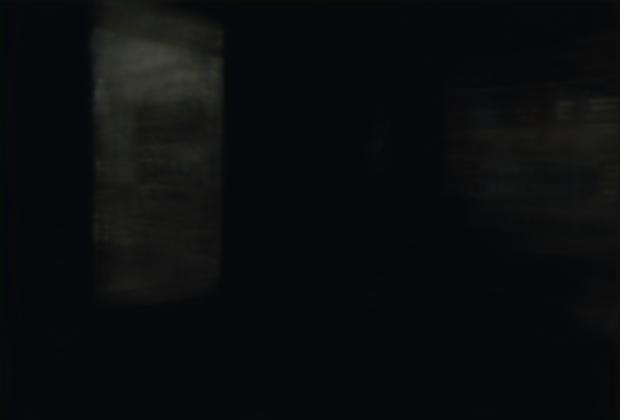}&
                \includegraphics[width=\newsubwidth\linewidth]{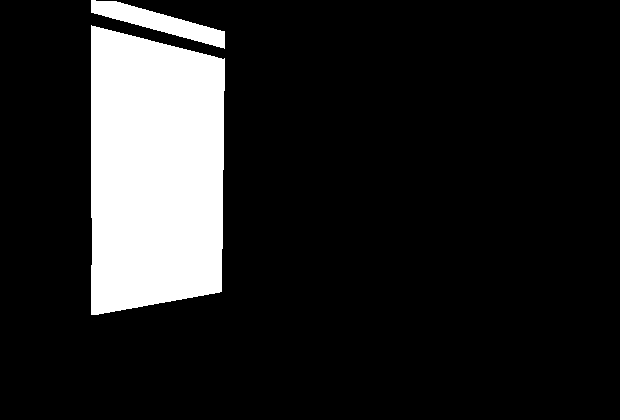}\\

                \includegraphics[width=\newsubwidth\linewidth]{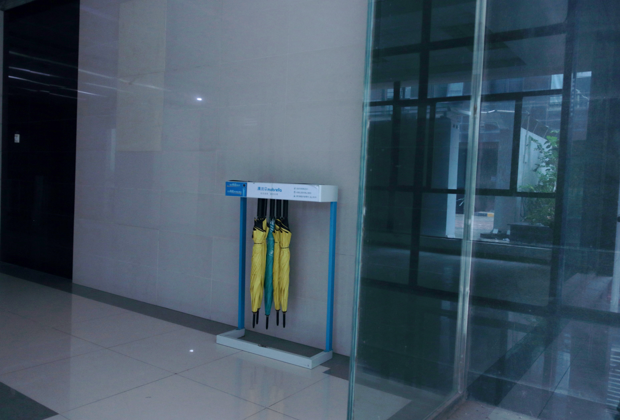}&
                \includegraphics[width=\newsubwidth\linewidth]{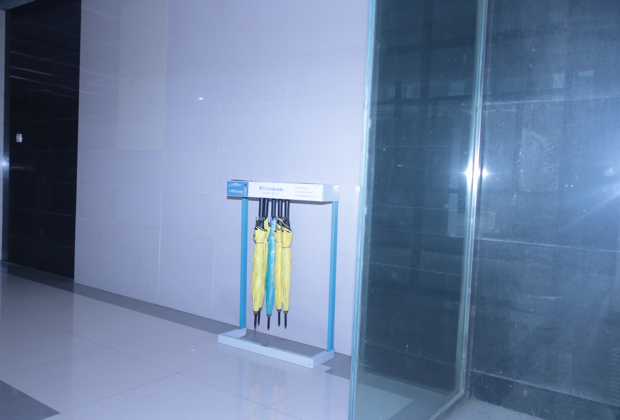}&
                \includegraphics[width=\newsubwidth\linewidth]{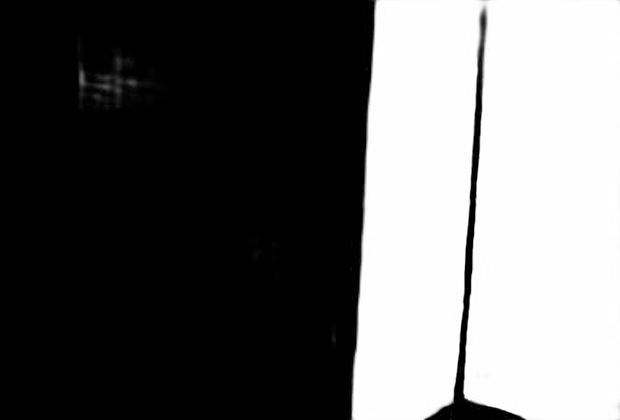}&
                \includegraphics[width=\newsubwidth\linewidth]{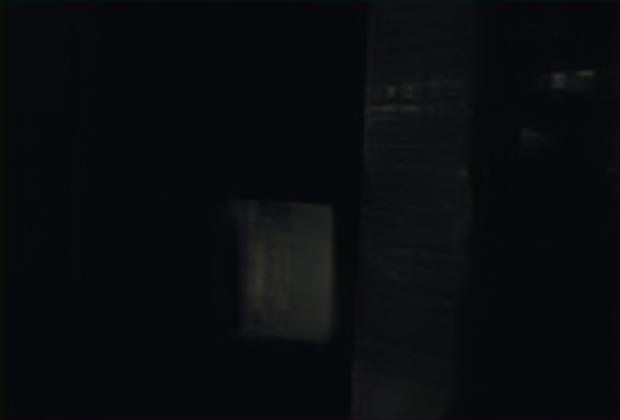}&
                \includegraphics[width=\newsubwidth\linewidth]{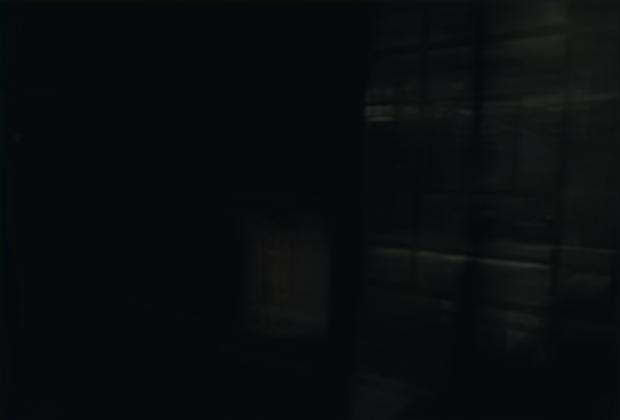}&
                \includegraphics[width=\newsubwidth\linewidth]{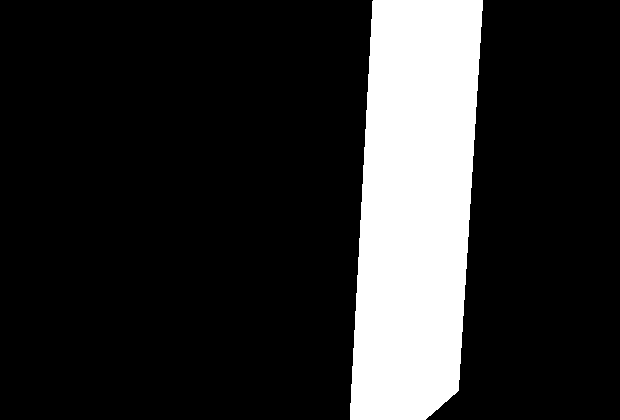}\\ 

                \scriptsize{No-flash Image}&
                \scriptsize{Flash Image}&
                \scriptsize{Ours}&
                \scriptsize{$\mathbf{R}_\text{no-flash}$}&
                \scriptsize{$\mathbf{R}_\text{flash}$}&
                \scriptsize{GT}\\
                
            \end{tabular}
        \end{center}
 	  \vspace{-0.016\textwidth}
        \caption{Failure cases: our method may over-detect glass regions in some challenging scenes. In the two scenes above, our method mis-detected smooth tiles as glass surfaces, as \tao{the prominent} reflections \tao{on their surfaces} may confuse our method.}
        \label{fig:failure_cases}
        \vspace{-4mm}
\end{figure}

\section{Conclusion}

\tao{In this paper, we observed that in most real-world scenes, the illumination intensity in front of the glass surface differs from that behind it, which leads to variations in the reflections visible on the glass.
Based on this phenomenon, we have proposed a novel method, \textit{NFGlassNet}, for glass surface detection, which takes no-flash and flash image pairs as input. Specially, we proposed a Reflection Contrast Mining Module (RCMM) for extracting reflections, and a Reflection Guided Attention Module (RGAM) for fusing features from reflection and glass surface for accurate glass surface detection.
}
For learning our network, we \tao{have also proposed} a no-flash and flash image pair dataset (NFGD), which contains approximately $\sim3.3K$ image pairs along with the corresponding ground truth annotations. 
Extensive experiments demonstrate that our method effectively extracts and \hh{exploits} reflection cues to identify glass surfaces and outperforms existing state-of-the-art methods on our dataset. 

\tao{In future work, we plan to leverage specialized sensors to improve our method, as our method may struggle with smooth non-glass surfaces that have prominent reflections. Moreover, we aim to extend our method to video-based glass surface detection, as motion cues from reflections in video could be further exploited to enhance detection performance.}


\bibliographystyle{IEEEtran}
\bibliography{egbib}
\clearpage
\appendices

\section{Evaluation Metrics}
We adopt IoU, F$_{\beta}$, MAE, BER, and ACC to evaluate our NFGlassNet and competing methods.
Specifically, the intersection over union (IoU) score is widely used in segmentation tasks, which is formulated as:
\begin{equation}
    IoU=\frac{\sum |{G \cap P}|}{\sum |G \cup P|},
\end{equation}
and MAE is formulated as:
\begin{equation}
    MAE=\frac{1}{H W} \sum_{i=1}^{H} \sum_{j=1}^{W}|P(i, j)-G(i, j)|,
\end{equation}
where $P$ denotes the predicted mask, and $G$ denotes the ground truth. $H$ and $W$ represent the width and height of the input image. (i, j) is the location of the pixel in the image.

F-measure is calculated by a weighted combination of Precision and Recall:
\begin{equation}
    F_{\beta}=\frac{1+\beta^{2}(\textit {Precision} \times \textit {Recall})}{\beta^{2} \textit {Precision}+ \textit {Recall}},
\end{equation}
where $\beta^2$ is set to $0.3$ to emphasize more on precision over recall as suggested in ~\cite{achanta2009frequency}.

BER is widely used in shadow detection to measure the binary prediction from a balance-aware perspective and is formulated as:
\begin{equation}
    BER=1-0.5 \times\left(\frac{N_{t p}}{N_{p}}+\frac{N_{t n}}{N_{n}}\right),
\end{equation}
and ACC is formulated as:
\begin{equation}
    ACC = \frac{N_{tp}}{N_{p}},
\end{equation}
where $N_{tp}$, $N_{tn}$, $N_{p}$, $N_{n}$ are the numbers of true positive, true
negative, glass, and non-glass pixels, respectively.

\section{Additional Ablation Studies on the Two Proposed Modules}~\label{sec:abl}
\subsection{Effectiveness of the RCMM}
As shown in Tab.~\ref{tab:ablation_study_RCMM}, we conduct \tao{an ablation study on the structure of RCMM}. 
$\mathit{B}$ shows a larger drop in performance than $\mathit{A}$, which indicates that the contrast branch (Eq.~\ref{eq:f_contrast}) plays a more important role in reflection extraction than the UNet branch (Eq.~\ref{eq:f_unet}). $\mathit{C}$ demonstrates that RCMM without permutation addition (Eq.~\ref{eq:rf}) results in a performance decrease, as more complementary information could interact during this step. \lywre{The same trend can also be seen when replacing the addition in Eq.~\ref{eq:rf} with subtraction ($\mathit{D}$) or concatenation operations ($\mathit{E}$).} $\mathit{F}$ shows that replacing feature addition with feature subtraction significantly reduces the performance of RCMM, which proves the importance of difference extraction through feature subtraction (Eq.~\ref{eq:f_contrast}).

\renewcommand{\tabcolsep}{2.1pt}
\renewcommand\arraystretch{1}
\begin{table}[t]
\caption{
Ablation study of blocks on the proposed RCMM. ``RCMM w/o UNet branch" refers to RCMM without using two UNet branches. ``RCMM w/o contrast branch" does not use contrast branch for difference extraction. ``RCMM w/o permutation addition" does not use rich difference features but instead uses single features from each convolution directly.
``addition$\rightarrow$subtraction" using subtraction to get the richer representations.
``addition$\rightarrow$concatenation" using concatenation to get the richer representations.
``subtraction$\rightarrow$addition" replaces rich difference feature subtraction with feature addition.} 
  \label{tab:ablation_study_RCMM}
  \vspace{-2mm}
  \centering
    \begin{tabular}{lccccc}
    \toprule
    Method  &IoU$\uparrow$ &F$_\beta$$\uparrow$ &MAE$\downarrow$ &BER$\downarrow$ &ACC$\uparrow$ \\
    \midrule
     $\mathit{A}$ \textit{w/o} UNet branch &86.02 &0.918 &0.081 &0.079 &0.920 \\
     $\mathit{B}$ \textit{w/o} contrast branch &84.27 &0.905 &0.089 &0.097 &0.913 \\
     $\mathit{C}$ \textit{w/o} permutation addition &85.53 &0.915 &0.084 &0.091 &0.922 \\
     $\mathit{D}$ addition$\rightarrow$subtraction &85.66 &0.920 &\Best{0.080} &0.081 &0.924 \\
     $\mathit{E}$ addition$\rightarrow$concatenation &86.09&0.924 &0.081 &0.080 &\Best{0.932} \\
     $\mathit{F}$ subtraction$\rightarrow$addition &85.45 &0.914 &0.085 &0.090 &0.922 \\

     \midrule
     $\mathit{G}$ Ours &\Best{86.46} &\Best{0.926} &\Best{0.080} &\Best{0.076} &0.922 \\
    \bottomrule
  \end{tabular}
  \vspace{-2mm}
\end{table}

\subsection{Effectiveness of the RGAM}
Similarly, we conduct ablation experiments on RGAM. As shown in Tab.~\ref{tab:ablation_study_RGAM}, the comparison between $\mathit{A}$ and $\mathit{G}$ demonstrates that $\mathbf{M_\text{shared}}$ (Eq.~\ref{eq:m_shared}) effectively fuses features between glass and reflection. $\mathit{B}$ and $\mathit{C}$ show that alternately using $\mathbf{F}_\text{glass}^{\mathbf{Q}}$ and $\mathbf{F}_\text{refle}^{\mathbf{Q}}$ (Eq.~\ref{eq:f_refle}) as queries significantly improves detection performance, with glass features as the query bringing a greater benefit. $\mathit{D}$ indicates that parallel branches, compared to serial branches, better facilitate feature fusion. The comparison between $\mathit{E}$ and $\mathit{G}$ demonstrates that the shift operator (Eq.~\ref{eq:m_shared}) improves the detection performance, as it effectively enhances the glass surface features. $\mathit{F}$ shows that the negative value features before shifting have a minor improvement in detection.

\renewcommand{\tabcolsep}{1.7pt}
\renewcommand\arraystretch{1}
\begin{table}[t]
\caption{
Ablation study of blocks on the proposed RGAM. ``$\mathbf{M_\text{shared}}$ $\rightarrow$ $\mathbf{M_\text{relfe}}$, $\mathbf{M_\text{glass}}$" refers to RGAM without using a shared attention map but using dual cross attention with two attention maps, respectively. ``RGAM w/o alternate $\mathbf{Q}$ ($\mathbf{F}_\text{glass}^{\mathbf{Q}}$/$\mathbf{F}_\text{refle}^{\mathbf{Q}}$ only)" does not let $\mathbf{F}_\text{glass}$ and $\mathbf{F}_\text{refle}$" take turns as the $\mathbf{Q}$, and instead, $\mathbf{F}_\text{glass}$/$\mathbf{F}_\text{refle}$ serves as the $\mathbf{Q}$ twice. ``parallel branch$\rightarrow$series branch" replaces dual parallel cross attention branches with single series cross attention branch like~\cite{lee2024guided} and $\mathbf{F}_\text{glass}^\mathbf{Q}$ is linked in series before $\mathbf{F}_\text{refle}^\mathbf{Q}$. ``shift$\rightarrow$relu" replaces shift with relu function.} 
  \label{tab:ablation_study_RGAM}
  \vspace{-2mm}
  \centering
    \begin{tabular}{lccccc}
    \toprule
    Method  &IoU$\uparrow$ &F$_\beta$$\uparrow$ &MAE$\downarrow$ &BER$\downarrow$ &ACC$\uparrow$ \\
    \midrule
     $\mathit{A}$ $\mathbf{M_\text{shared}}$ $\rightarrow$ $\mathbf{M_\text{relfe}}$, $\mathbf{M_\text{glass}}$ &85.48 &0.916 &0.087 &0.085 &0.917 \\
     $\mathit{B}$ \textit{w/o} alternate $\mathbf{Q}$ ($\mathbf{F}_\text{glass}^{\mathbf{Q}}$ only) &85.26 &0.915 &0.082 &0.092 &0.913 \\
     $\mathit{C}$ \textit{w/o} alternate $\mathbf{Q}$ ($\mathbf{F}_\text{refle}^{\mathbf{Q}}$ only) &84.89 &0.910 &0.091 &0.099 &0.907 \\
     $\mathit{D}$ parallel branch$\rightarrow$series branch  &84.97 &0.914 &0.088 &0.098 &0.909 \\
     $\mathit{E}$ \textit{w/o} shift &85.35 &0.916 &0.084 &0.083 &0.911 \\
     $\mathit{F}$ shift$\rightarrow$relu  &85.93 &0.919 &0.096 &0.082 &0.919 \\
     \midrule
     $\mathit{G}$ Ours &\Best{86.46} &\Best{0.926} &\Best{0.080} &\Best{0.076} &\Best{0.922} \\
    \bottomrule
  \end{tabular}
  \vspace{-2mm}
\end{table}

\section{Feature Visualizations of the Proposed Modules}
As shown in Fig.~\ref{fig:heat_map}, we also visualize some intermediate results of our modules using heat maps. We can observe that the output features from the backbone ($F_{no-flash}^{3}$) are rough and random. After passing through the RCMM, the output features ($F_{refle}^{3}$) are able to \tao{highlight} \hh{the reflection regions.}
\hh{After passing through RGAM, the regions that contain both reflections and glass frames are further highlighted in the output features ($F_{RGAM}^{3}$), which indicates the location of the glass surface. Finally, after passing through the decoder modules, the highlighted features in ($D_3$) are closer to the glass surfaces through the cascade feature decoding.}

\renewcommand{\newsubwidth}{0.121}
\begin{figure*}[ht]
	\renewcommand{\tabcolsep}{0.8pt}
	\renewcommand\arraystretch{0.6}
        \begin{center}
            \begin{tabular}{cccccccc}
                \includegraphics[width=\newsubwidth\linewidth]{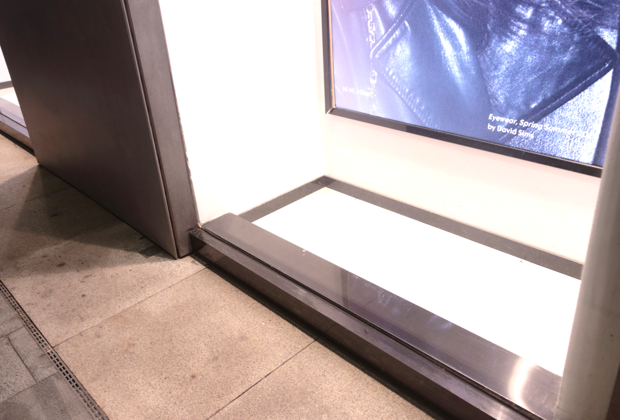}&
                \includegraphics[width=\newsubwidth\linewidth]{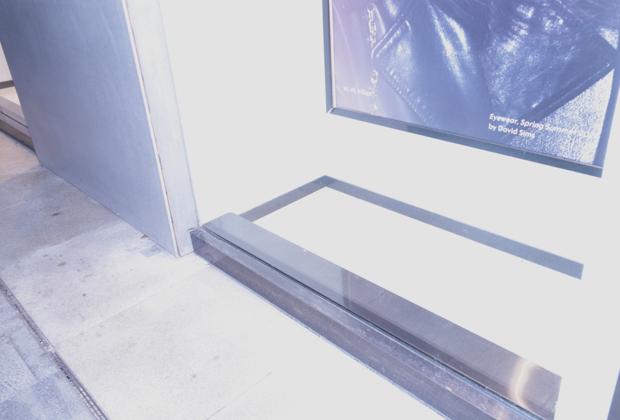}&
                \includegraphics[width=\newsubwidth\linewidth]{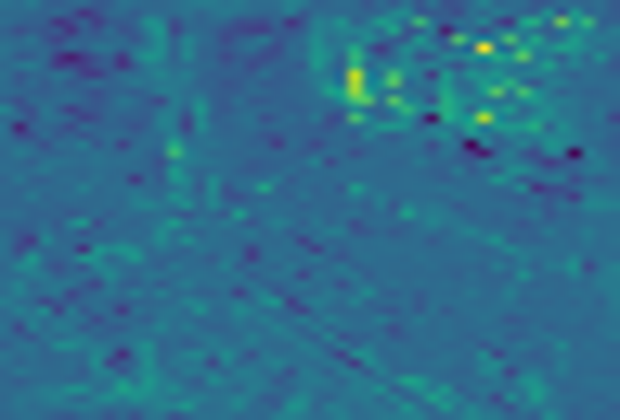}&
                \includegraphics[width=\newsubwidth\linewidth]{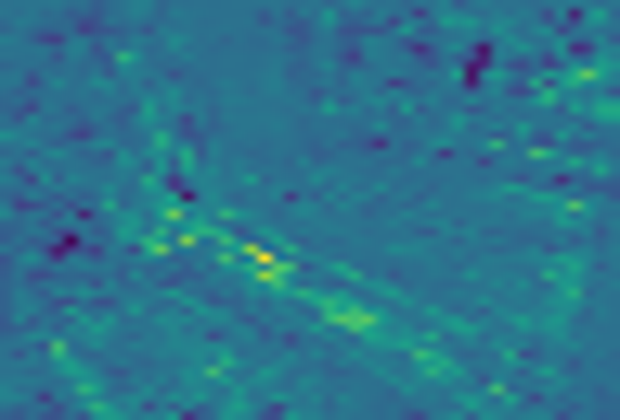}&
                \includegraphics[width=\newsubwidth\linewidth]{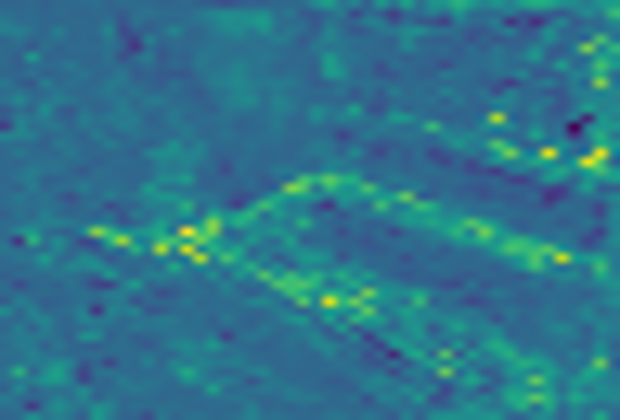}&
                \includegraphics[width=\newsubwidth\linewidth]{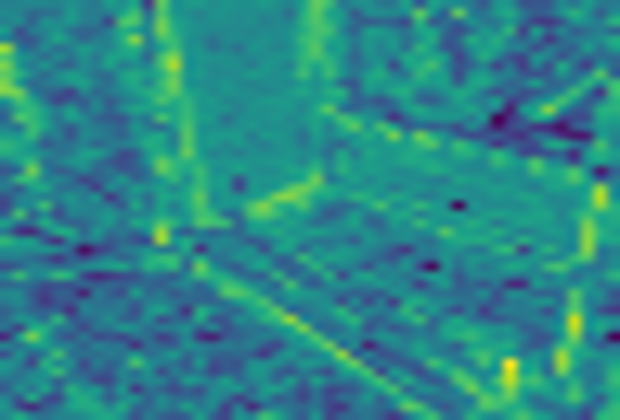}&
                \includegraphics[width=\newsubwidth\linewidth]{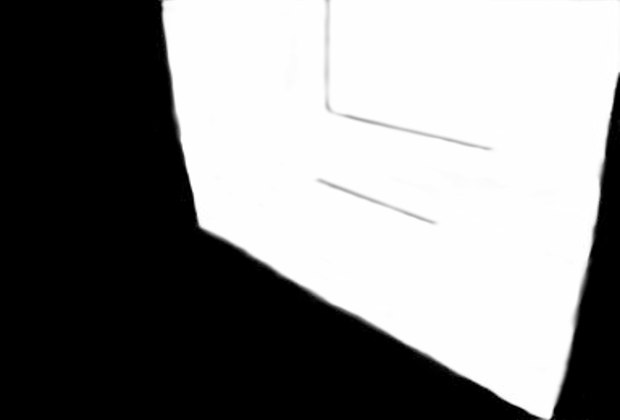}&
                \includegraphics[width=\newsubwidth\linewidth]{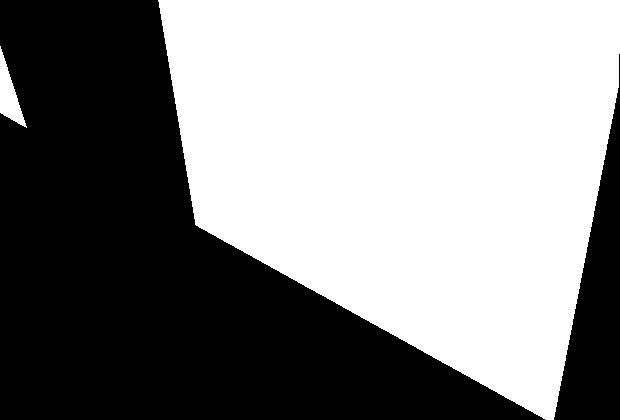}
                \\

                \includegraphics[width=\newsubwidth\linewidth]{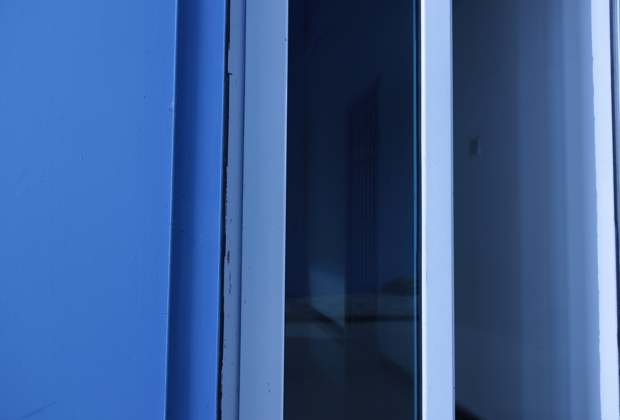}&
                \includegraphics[width=\newsubwidth\linewidth]{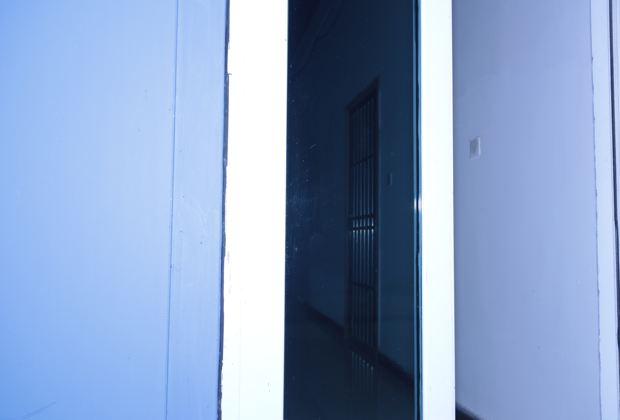}&
                \includegraphics[width=\newsubwidth\linewidth]{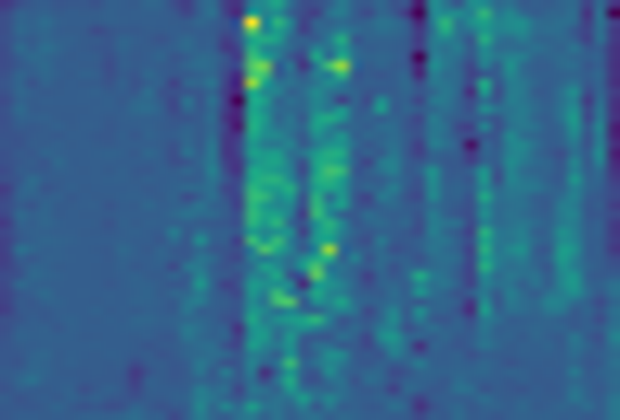}&
                \includegraphics[width=\newsubwidth\linewidth]{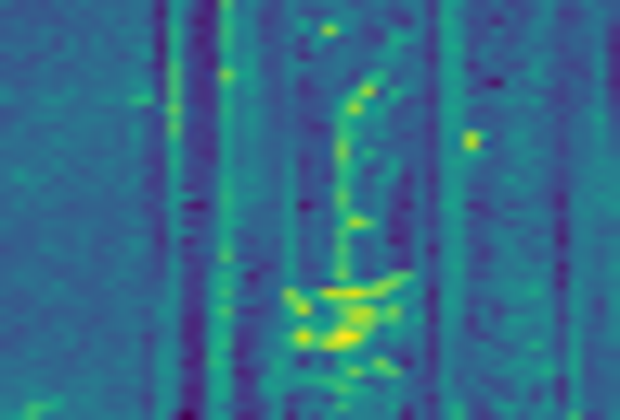}&
                \includegraphics[width=\newsubwidth\linewidth]{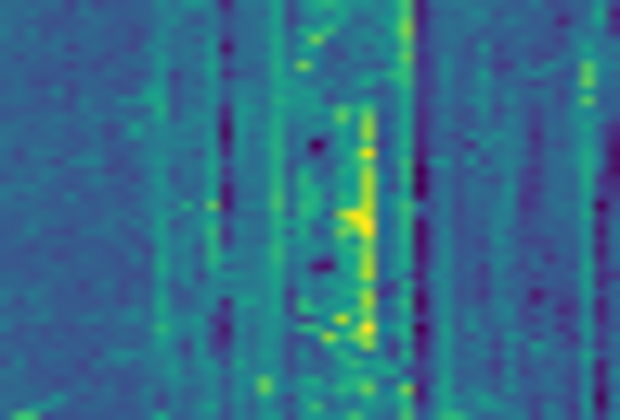}&
                \includegraphics[width=\newsubwidth\linewidth]{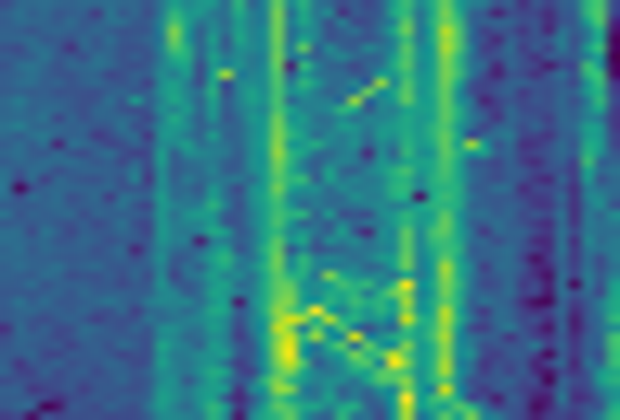}&
                \includegraphics[width=\newsubwidth\linewidth]{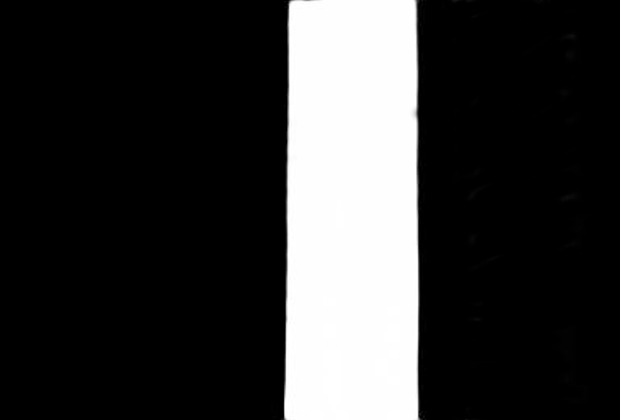}&
                \includegraphics[width=\newsubwidth\linewidth]{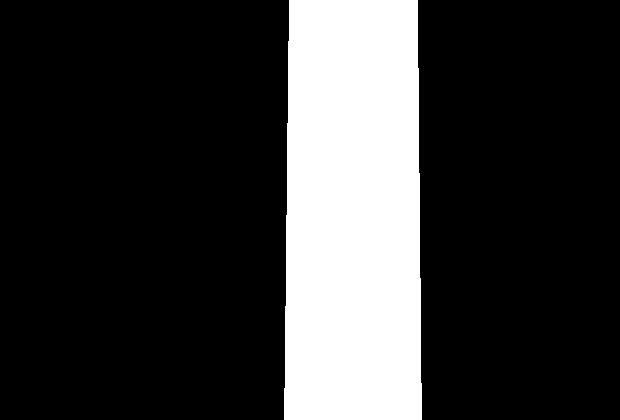}
                \\
               
                \scriptsize{No-flash Image}&
                \scriptsize{Flash Image}&
                \scriptsize{$\mathbf{F_{\text{no-flash}}^\text{3}}$}&
                \scriptsize{$\mathbf{F_{\text{refle}}^\text{3}}$}&
                \scriptsize{$\mathbf{F_{\text{RGAM}}^\text{3}}$}&
                \scriptsize{$\mathbf{D_{\text{3}}}$}&
                \scriptsize{Mask (Ours)}&
                \scriptsize{GT}
                \\
                
            \end{tabular}
        \end{center}
 	\vspace{-0.016\textwidth}
        \caption{Visual heat maps of the output features of key modules in our network.}
        \label{fig:heat_map}
\end{figure*}

\section{Study on misalignment Inputs}
\lywre{In our framework, the RCMM module adopts subtraction of feature maps to capture difference cues. This operation inherently assumes well-aligned features to ensure stable and accurate fusion. However, in real applications, the obtained image pairs do not always align perfectly. To simulate this, we added some random shifts to the $384\times384$ flash images to test the robustness of our model to misaligned images.}

\lywre{Table~\ref{tab:results_misalient} reports the quantitative results under different levels of random misalignment. As expected, the performance gradually degrades as the magnitude of the spatial shift increases. Nevertheless, when the misalignment is limited to 1--5 pixels, our model only suffers a marginal performance drop, with the IoU decreasing by 1.37\%, indicating that the proposed framework is robust to the slight registration errors commonly encountered in real-world image acquisition. More severe misalignments (5--15 and 15--30 pixels) lead to more noticeable performance degradation, because the RCMM module relies on feature-level subtraction to model complementary cues, and large spatial discrepancies inevitably weaken the effectiveness of difference-based feature fusion.
}

\begin{table}[t]
  \centering
  \caption{The impact of misaligned images.}
  \begin{tabular}{@{}lccccc@{}}
    \toprule
    pixel range & IoU↑ & F$_\beta$↑ & MAE↓ & BER↓ &ACC↑\\
    \midrule
    0-0 pixel&\Best{86.46}&\Best{0.926}&\Best{0.080}&\Best{0.076}&\Best{0.922}\\
    1-5 pixel&85.09&0.914&0.089&0.097&\Best{0.922}\\
    5-15 pixel&79.03&0.882&0.124&0.137&0.878\\
    15-30 pixel&71.47&0.836&0.169&0.186&0.808\\
    \bottomrule
  \end{tabular}
  \label{tab:results_misalient}
\end{table}

\section{More Results}

\begin{figure}[ht]
  \centering
    \begin{minipage}[b]{\linewidth}
      \subfloat[Light Side w/o Flash]{
        \label{fig:light_side_no_flash_camera_view}
        \includegraphics[width=0.45\linewidth]{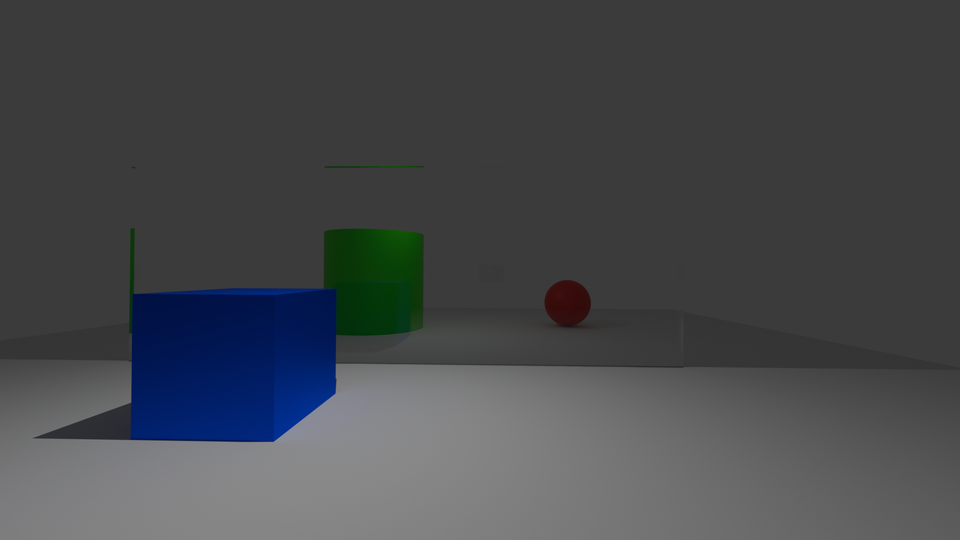}
      }
      \subfloat[Light Side w/ Flash]{
        \label{fig:light_side_flash_camera_view}
        \includegraphics[width=0.45\linewidth]{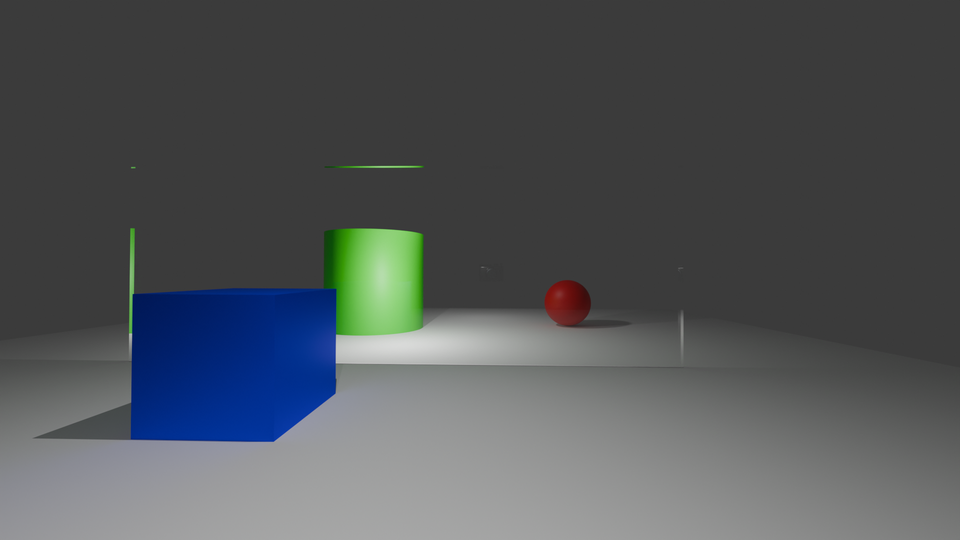}
      }
    \end{minipage}
    \begin{minipage}[b]{\linewidth}
      \subfloat[Dark Side w/o Flash]{
        \label{fig:dark_side_no_flash_camera_view}
        \includegraphics[width=0.45\linewidth]{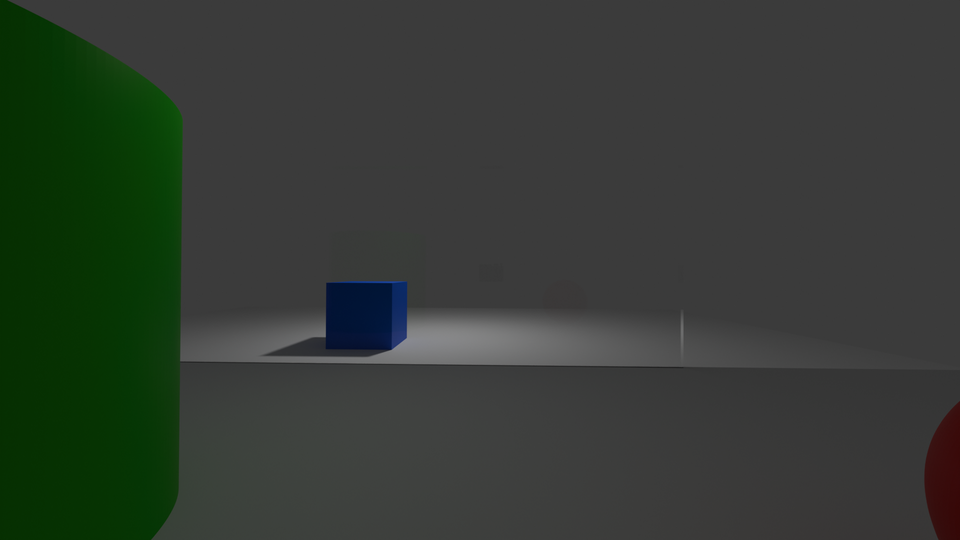}
      }
      \subfloat[Dark Side w/ Flash]{
        \label{fig:dark_side_flash_camera_view}
        \includegraphics[width=0.45\linewidth]{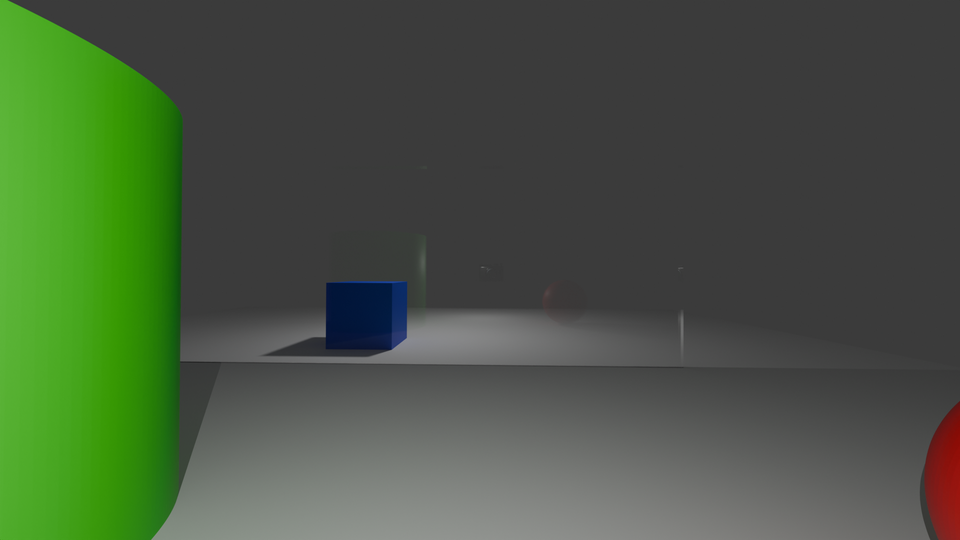}
      }
    \end{minipage}
    \caption{The corresponding camera views of the no-flash and flash scenes in different illumination sides.}
    \label{fig:blender_rendering_camera_view}
\end{figure}

Fig.~\ref{fig:blender_rendering} illustrates the appearance and disappearance of reflections on the glass surface from a bird’s-eye view. Fig.~\ref{fig:blender_rendering_camera_view} presents the corresponding camera views, which more accurately represent the perspective of a real observer.

\begin{figure}[ht]
\centerline{\includegraphics[width=0.45\textwidth]{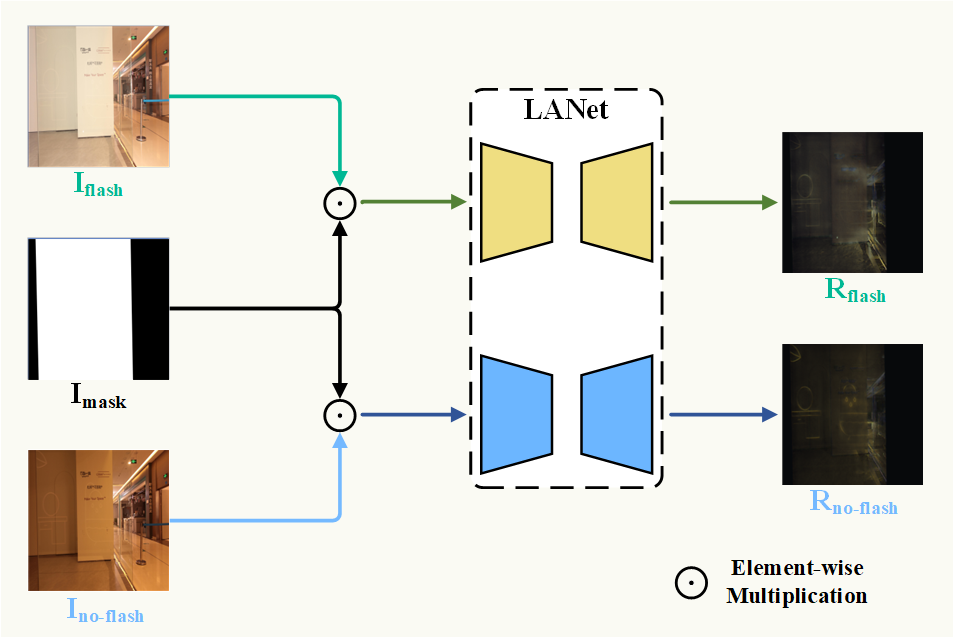}}
\caption{The process of making pseudo ground truth for reflection.}
\label{fig:PseudoGT_Process}
\end{figure}

Fig.~\ref{fig:PseudoGT_Process} shows the process of generating pseudo ground truth for reflections in our NFGD. We inferred the result through the reflection removal method LANet\cite{dong2021location} to obtain the masked reflections, ensuring that the reflections exist only on the glass surface.

Fig.~\ref{fig:glass_comparision_result_on_NFGD} shows a more visual comparison of glass surface detection for our network and the competing methods (GDNet~\cite{mei2020don}, GSDNet~\cite{lin2021rich}, CSFwin~\cite{xie2024csfwinformer}, RGB-S~\cite{lin2022semantic}, RGB-NIR~\cite{yan2024nrglassnet}, and RGB-D~\cite{mei2021depth}) evaluated on our NFGD dataset.

Fig.~\ref{fig:glass_comparision_result_on_RGBT} and ~\ref{fig:glass_comparision_result_on_RGBNir} show visual comparison of competing methods (SPNet~\cite{SPNet}, PDNet~\cite{mei2021depth}, CIRNet~\cite{cong2022cir}, WaveNet~\cite{WaveNet}, RGB-T~\cite{huo2023glass} and RGB-NIR~\cite{yan2024nrglassnet}) which perform well in our NFGD on the RGB-T~\cite{huo2023glass} and RGB-NIR~\cite{yan2024nrglassnet} glass surface dataset, respectively.


\clearpage
\renewcommand{\newsubwidth}{0.086}
\begin{figure*}[ht]
	\renewcommand{\tabcolsep}{0.8pt}
	\renewcommand\arraystretch{0.6}
        \begin{center}
\begin{tabular}{ccccccccccc}
                \includegraphics[width=\newsubwidth\linewidth]{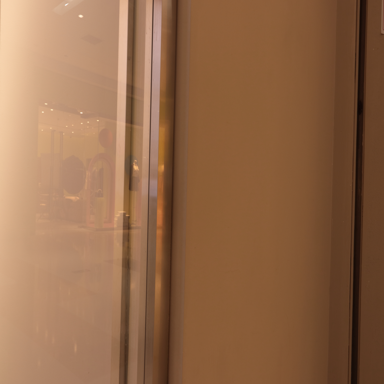}&
                \includegraphics[width=\newsubwidth\linewidth]{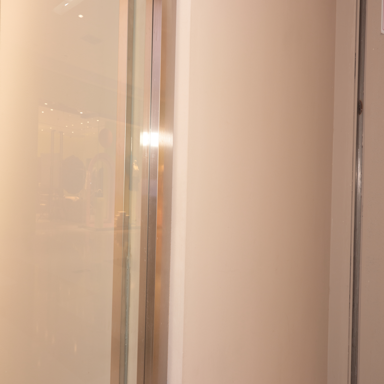}&
                \includegraphics[width=\newsubwidth\linewidth]{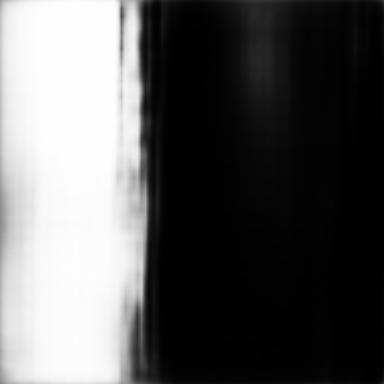}&
                \includegraphics[width=\newsubwidth\linewidth]{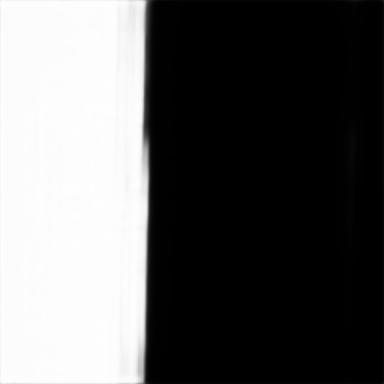}&
                \includegraphics[width=\newsubwidth\linewidth]{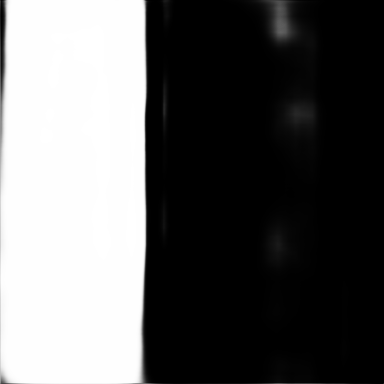}&
                \includegraphics[width=\newsubwidth\linewidth]{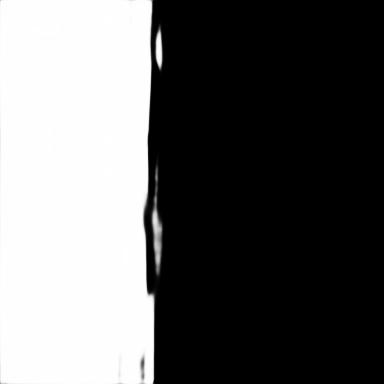}&
                \includegraphics[width=\newsubwidth\linewidth]{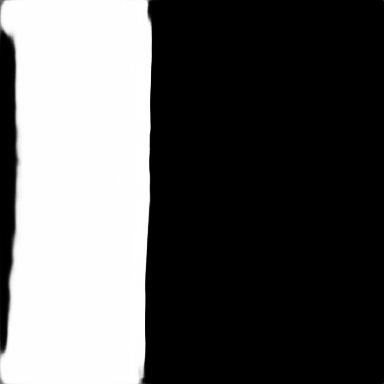}&
                \includegraphics[width=\newsubwidth\linewidth]{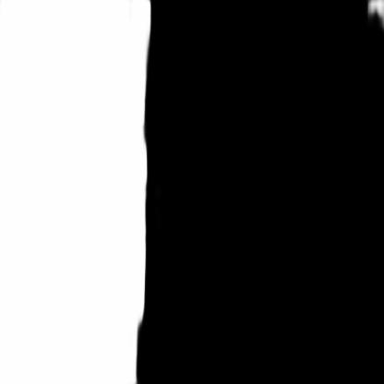}&
                \includegraphics[width=\newsubwidth\linewidth]{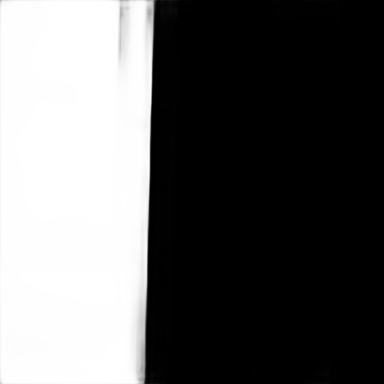}&
                \includegraphics[width=\newsubwidth\linewidth]{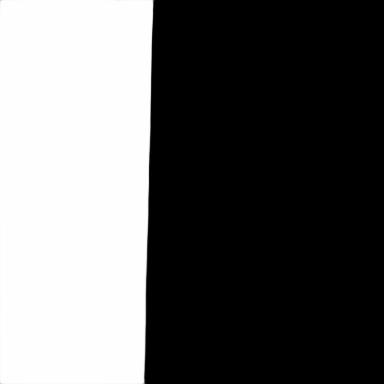}&
                \includegraphics[width=\newsubwidth\linewidth]{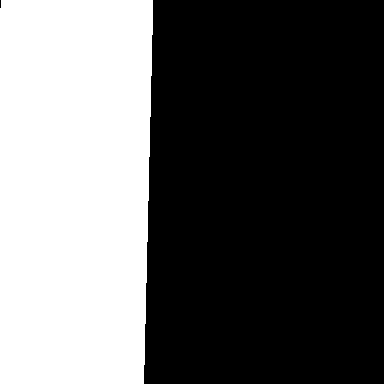} \\
\end{tabular}
        \end{center}
 	\vspace{-0.016\textwidth}
        \caption{More visual comparison of the glass surface detection results against state-of-the-art methods on our NFGD dataset. 
        }
        \label{fig:glass_comparision_result_on_NFGD}
\end{figure*}

\clearpage
\renewcommand{\newsubwidth}{0.094}
\begin{figure*}
	\renewcommand{\tabcolsep}{0.8pt}
	\renewcommand\arraystretch{0.6}
        \begin{center}
            \begin{tabular}{cccccccccc}
                \includegraphics[width=\newsubwidth\linewidth]{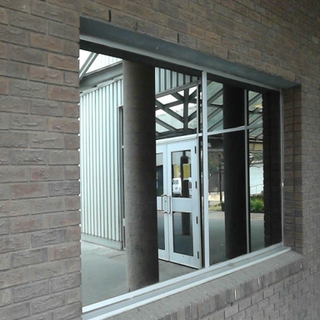}&
                \includegraphics[width=\newsubwidth\linewidth]{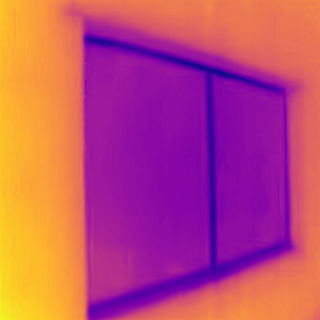}&
                \includegraphics[width=\newsubwidth\linewidth]{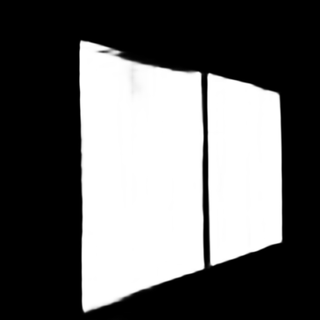}&
                \includegraphics[width=\newsubwidth\linewidth]{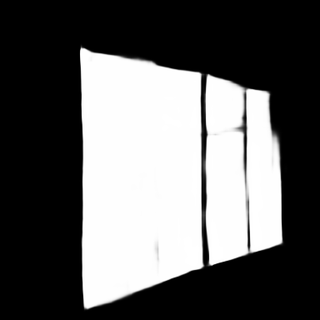}&
                \includegraphics[width=\newsubwidth\linewidth]{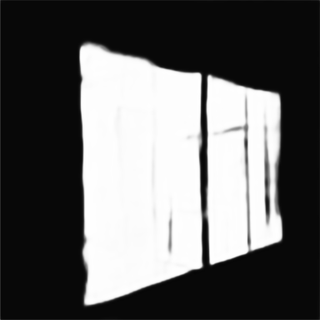}&
                \includegraphics[width=\newsubwidth\linewidth]{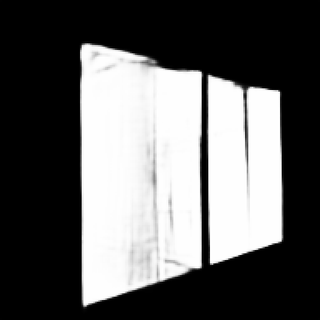}&
                \includegraphics[width=\newsubwidth\linewidth]{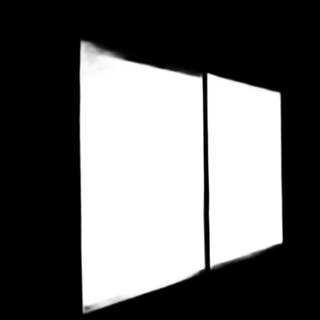}&
                \includegraphics[width=\newsubwidth\linewidth]{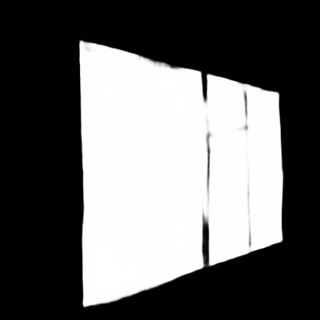}&
                \includegraphics[width=\newsubwidth\linewidth]{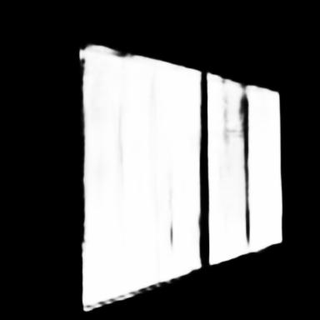}&
                \includegraphics[width=\newsubwidth\linewidth]{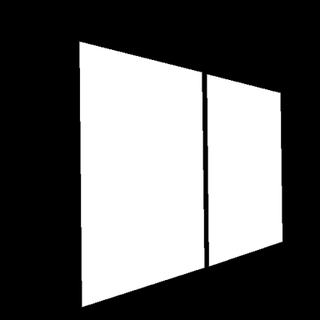} 
                \\
\end{tabular}
        \end{center}
 	\vspace{-0.016\textwidth}
        \caption{Visual comparison of the glass surface detection results against competing methods on the RGB-T dataset~\cite{huo2023glass}.}
        \label{fig:glass_comparision_result_on_RGBT}
\end{figure*}
\renewcommand{\newsubwidth}{0.094}
\begin{figure*}
	\renewcommand{\tabcolsep}{0.8pt}
	\renewcommand\arraystretch{0.6}
        \begin{center}
            \begin{tabular}{cccccccccc}
                \includegraphics[width=\newsubwidth\linewidth]{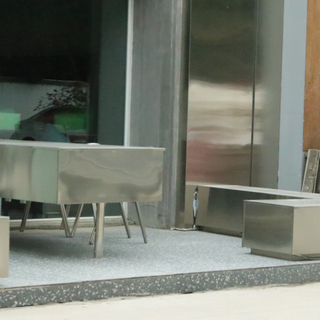}&
                \includegraphics[width=\newsubwidth\linewidth]{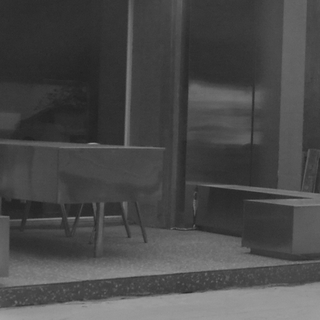}&
                \includegraphics[width=\newsubwidth\linewidth]{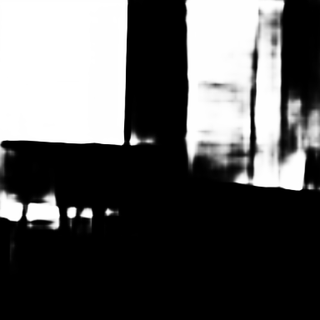}&
                \includegraphics[width=\newsubwidth\linewidth]{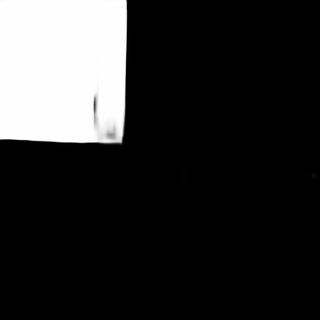}&
                \includegraphics[width=\newsubwidth\linewidth]{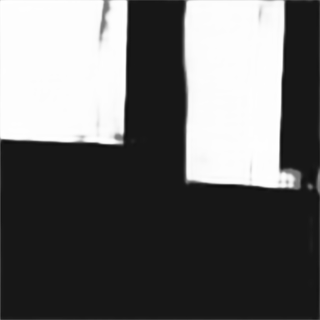}&
                \includegraphics[width=\newsubwidth\linewidth]{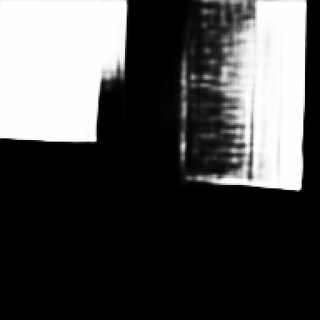}&
                \includegraphics[width=\newsubwidth\linewidth]{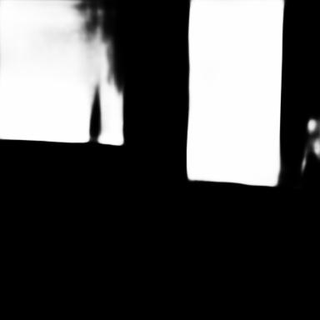}&
                \includegraphics[width=\newsubwidth\linewidth]{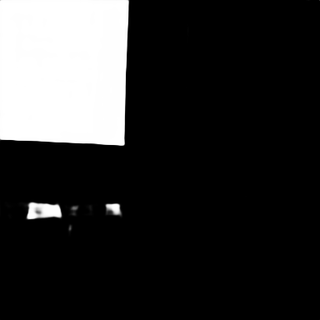}&
                \includegraphics[width=\newsubwidth\linewidth]{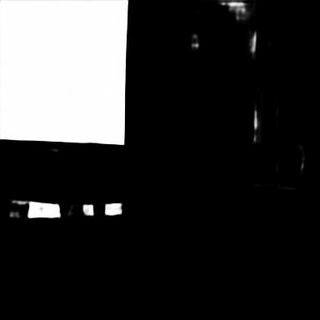}&
                \includegraphics[width=\newsubwidth\linewidth]{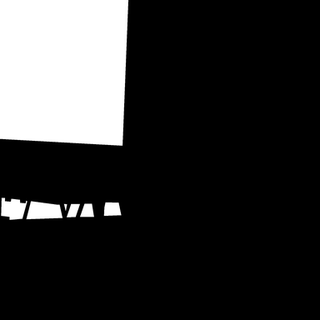} 
                \\
\end{tabular}
        \end{center}
 	\vspace{-0.016\textwidth}
        \caption{Visual comparison of the glass surface detection results against competing methods on the RGB-NIR dataset~\cite{yan2024nrglassnet}.}
        \label{fig:glass_comparision_result_on_RGBNir}
\end{figure*}






\end{document}